\documentclass[a4paper,fleqn]{cas-dc}

\usepackage{xcolor}
\usepackage{enumitem}
\usepackage{flushend}
\usepackage{balance}
\usepackage{adjustbox}
\usepackage{float}
\usepackage{algpseudocode}
\usepackage{todonotes}
\usepackage[ruled,vlined,noresetcount]{algorithm2e} 
\usepackage{pifont}
\usepackage{soul}
\usepackage{amsmath}
\usepackage{amsfonts}
\usepackage{amssymb}
\usepackage{graphicx}
\usepackage{caption}
\usepackage{hyperref}
\usepackage{subfig}
\usepackage{multirow}
\usepackage{booktabs}
\usepackage{geometry}
\usepackage{longtable}
\usepackage{pifont}
\newcommand{\cmark}{\ding{51}}%
\newcommand{\xmark}{\text{\ding{55}}}
\usepackage{textcomp}

\usepackage{tikz}

\usepackage{graphicx}
\graphicspath{{./img/}}
\usepackage{subfig}

\usepackage{multirow}

\def\tsc#1{\csdef{#1}{\textsc{\lowercase{#1}}\xspace}}
\tsc{WGM}
\tsc{QE}
\tsc{EP}
\tsc{PMS}
\tsc{BEC}
\tsc{DE}
\usepackage{float}
 \usepackage[usenames,dvipsnames]{pstricks}
\usepackage{pstricks-add}
 \usepackage{epsfig}
 \usepackage{pst-grad} 
 \usepackage{pst-plot} 
 \usepackage[space]{grffile} 
 \usepackage{etoolbox} 
 \makeatletter 
 \patchcmd\Gread@eps{\@inputcheck#1 }{\@inputcheck"#1"\relax}{}{}
 \makeatother

\usepackage{verbatim}
\usepackage{multicol}
\usepackage{epstopdf}
\usepackage[numbers]{natbib}

\begin{document}
\let\WriteBookmarks\relax
\def\floatpagepagefraction{1}
\def\textpagefraction{.001}

\shorttitle{Application of Knowledge Distillation in Remote Sensing}

\shortauthors{Y. Himeur, et al.}

\title [mode = title]{Applications of Knowledge Distillation in Remote Sensing: A Survey }

\vskip2mm

\author[1]{ Yassine Himeur }
\ead{yhimeur@ud.ac.ae}

\author[2]{Nour Aburaed}

\author[3]{Omar Elharrouss}

\author[3]{Iraklis Varlamis}

\author[1]{Shadi Atalla}

\author[1]{Wathiq Mansoor}

\author[2]{Hussain Al Ahmad}

\address[1]{College of Engineering and Information Technology, University of Dubai, Dubai, UAE}

\address[2]{MBRSC Lab, University of Dubai, Dubai 2713, UAE}

\address[3]{Department of Computer Science and Software Engineering, United Arab Emirates University, UAE}

\address[3]{Department of Informatics and Telematics, Harokopio University of Athens, GR-17778 Athens, Greece}

\begin{abstract}
With the ever-growing complexity of models in the field of remote sensing (RS), there is an increasing demand for solutions that balance model accuracy with computational efficiency. Knowledge distillation (KD) has emerged as a powerful tool to meet this need, enabling the transfer of knowledge from large, complex models to smaller, more efficient ones without significant loss in performance. 
This review article provides an extensive examination of KD and its innovative applications in RS. KD, a technique developed to transfer knowledge from a complex, often cumbersome model (teacher) to a more compact and efficient model (student), has seen significant evolution and application across various domains. Initially, we introduce the fundamental concepts and historical progression of KD methods. The advantages of employing KD are highlighted, particularly in terms of model compression, enhanced computational efficiency, and improved performance, which are pivotal for practical deployments in RS scenarios. The article provides a comprehensive taxonomy of KD techniques, where each category is critically analyzed to demonstrate the breadth and depth of the alternative options, and illustrates specific case studies that showcase the practical implementation of KD methods in RS tasks, such as instance segmentation and object detection. Further, the review discusses the challenges and limitations of KD in RS, including practical constraints and prospective future directions, providing a comprehensive overview for researchers and practitioners in the field of RS. Through this organization, the paper not only elucidates the current state of research in KD but also sets the stage for future research opportunities, thereby contributing significantly to both academic research and real-world applications.

\end{abstract}




\maketitle

\section{Introduction}
\subsection{Preliminary}
Remote sensing (RS) image analysis plays a pivotal role in interpreting and managing Earth's natural and human-made environments \cite{wu2024takd}. This technology harnesses data captured by satellites or high-altitude aircraft, providing crucial insights across a broad spectrum of applications—from agricultural monitoring and disaster management to urban planning and climate science \cite{xie2024decoupled}. By enabling timely and efficient observation of vast, inaccessible, or dangerous areas, RS becomes indispensable for tracking environmental changes, predicting weather patterns, and managing natural resources \cite{chen2024discretization}. Consequently, the ability to quickly process and analyze RS images leads to more informed decision-making, enhancing our global capability to respond to challenges such as food security, natural disasters, and climate change \cite{pang2024exploring,paranata2023catastrophe}.

However, the complexity of RS tasks varies significantly depending on the specific application, and many of these tasks are inherently challenging and computationally intensive \cite{wu2024beyond,atalla2023iot}. Key tasks such as image classification, object detection, change detection, and segmentation involve processing high-dimensional data, often characterized by large spatial and spectral resolutions \cite{ma2024transfer,liu2024spectral}. For instance, distinguishing between different land cover types or detecting minute changes over time in vast geographical areas necessitates sophisticated algorithms capable of handling enormous datasets \cite{salem2023deep}. Moreover, the presence of noise, variability in lighting conditions, atmospheric distortions, and the need for high precision further compound the complexity of these tasks \cite{liu2024multimodal,ouamane2024enhancing}. These challenges lead to extensive training times, particularly for deep learning models, which require large datasets to achieve high accuracy and generalization. Therefore, optimizing these models to balance accuracy and computational efficiency remains an ongoing challenge in RS \cite{lu2024goa,zhang2024ffca}.

Artificial Intelligence (AI), particularly machine learning (ML) and deep learning (DL), has revolutionized RS image analysis by introducing levels of precision and efficiency previously unattainable with traditional methods \cite{himeur2023video,lopes2024sensor,sohail2024advancing}. DL models, especially those based on Convolutional Neural Networks (CNNs), are highly adept at handling high-dimensional data from RS imagery \cite{yu2024distillation}. These models excel in tasks such as pattern recognition, object detection, and semantic segmentation, where they can automatically identify features like roads, buildings, or vegetation changes \cite{xu2024double}. Furthermore, the deployment of AI enables the processing of large datasets in real-time, significantly improving the accuracy of predictions and analyses. Moreover, DL's ability to learn feature representations without manual intervention reduces reliance on expert-driven feature design, thus scaling up the analytical capabilities of RS technologies \cite{rs15153859}.

Despite these advances, the integration of AI and DL into RS presents several significant challenges. One of the foremost issues is the requirement for substantial computational resources, particularly for training large neural network models \cite{kerdjidj2024uncovering,sayed2023time,limits20}. This becomes a critical barrier for organizations with limited access to high-performance computing infrastructure \cite{himeur2022using}. Additionally, DL models often require vast labeled datasets for training, which can be difficult and costly to acquire in the context of RS. Furthermore, these models are prone to overfitting, especially when trained on limited datasets, reducing their ability to generalize well to new, unseen data \cite{JMLR:v15:srivastava14a}. Another pressing concern is the "black box" nature of DL models, which often leads to difficulties in interpreting their decision-making processes—a critical requirement in applications where transparency and understanding are paramount, such as in environmental compliance and strategic planning \cite{KONYA2024167705,Biloslavo2024}.

To address some of these challenges, knowledge distillation (KD) emerges as a promising technique. KD involves training a smaller, more efficient student model to replicate the performance of a larger, more complex teacher model \cite{ji2023coarse}. By transferring knowledge from a high-performing neural network to a compact model, KD reduces the computational resources required for deployment, making advanced AI-driven RS technologies more accessible \cite{yu2022data}. Moreover, the student model can often achieve comparable accuracy with less data, mitigating the issues of extensive data requirements and overfitting \cite{kheddar2023deep}. In resource-constrained environments, KD proves particularly advantageous, as it enables energy-efficient deployment, thereby reducing the carbon footprint of AI systems \cite{zhang2024knowledge}. Additionally, KD facilitates the transfer of pre-trained models to other domains through fine-tuning, extending the versatility of AI applications even in scenarios with scarce data. Furthermore, KD techniques can assist in generating synthetic training data when annotated data is limited, thus addressing one of the critical bottlenecks in RS \cite{kerdjidj2023exploring,kheddar2023deep}. 
The resulting simpler models from the distillation process also offer easier interpretability, providing clearer insights into their decision-making mechanisms \cite{sun2024logit,BECHAR20241903}. This interpretability is essential for applications requiring transparency, such as environmental monitoring and regulatory compliance. As a result, KD not only democratizes AI capabilities within RS but also enhances the practical utility of these technologies in critical applications, ensuring a balance between performance, energy efficiency, and scalability across diverse domains \cite{zhang2024knowledge}.

\begin{table*}[h!]

\scriptsize
\begin{tabular}{|p{1.5cm}|p{6.5cm}|p{1.5cm}|p{6.5cm}|}
\hline
\textbf{Abbreviation} & \textbf{Full Form} & \textbf{Abbreviation} & \textbf{Full Form} \\ 
KD   & Knowledge Distillation      & YOLOv8 & You Only Look Once version 8            \\ 
RS   & Remote Sensing              & MS2RGB & Multispectral to RGB Knowledge Distillation \\ 
CNN  & Convolutional Neural Network & PseKD  & Phase-shift Encoded Knowledge Distillation \\ 
S-T  & Student-Teacher             & GSGNet & Graph Semantic Guided Network           \\ 
ARSD & Adaptive Reinforcement Supervision Distillation & LPIS   & Land Parcel Identification System     \\ 
RGB  & Red, Green, Blue            & DOTA   & Dataset for Object Detection in Aerial Images \\ 
R-CNN & Region-based Convolutional Neural Network & DIOR  & Dataset for Object Detection in Remote Sensing \\ 
FPN  & Feature Pyramid Network     & AID    & Aerial Image Dataset                    \\ 
MCFI & Multiscale Core Features Imitation & SSKDNet & Self-supervised Knowledge Distillation Network \\ 
SSRD & Strict Supervision Regression Distillation & MSKA   & Multi-level Semantic Knowledge Alignment \\ 
CFKD & Cross-layer Fusion for Knowledge Distillation & ViTs   & Vision Transformers                    \\ 
YOLO & You Only Look Once          & FPN    & Feature Pyramid Network                 \\ 
HSI  & Hyperspectral Image         & ERKT   & Efficient and Robust Knowledge Transfer \\ 
CKD  & Collaborative Consistent Knowledge Distillation & TWA    & Two-way Adaptive Distillation           \\ 
GKD  & Generalized Knowledge Distillation & NLD    & Noisy Label Distillation               \\ 
DKD  & Decoupled Knowledge Distillation & CAMs   & Class Activation Maps                  \\ 
SSFD & Spatial Feature Blurring for Distillation & RS-SSKD & Remote Sensing Self-supervised Knowledge Distillation \\ 
LEVIR & Large-scale Earth Vision Image Recognition & SAR SSDD & Synthetic Aperture Radar Ship Detection Dataset \\ 
UCMerced & University of California Merced Land-use Dataset & NWPU-RESISC & Northwestern Polytechnical University Remote Sensing Image Scene Classification \\ 
CMD  & Class Mean Distillation     & MSW    & Maximum Sustained Wind                 \\ \hline
\end{tabular}
\end{table*}

\subsection{Comparison with Existing Reviews}
Several recent reviews and surveys have provided comprehensive analyses of various aspects of Knowledge Distillation (KD) and its applications across different domains. These works highlight the evolution, challenges, and future directions of KD, focusing on areas such as computer vision, medical applications, and large language models. For instance, \cite{wang2021knowledge} offers an in-depth examination of KD within the framework of the Student-Teacher (S-T) learning model, providing a thorough overview of KD's core concepts, methods, and applications, particularly in vision tasks. The study also identifies key challenges and potential future research directions. Similarly, \cite{chen2021distilling} explores the significance of cross-stage connection paths between teacher and student networks, introducing a novel approach that enhances the effectiveness of KD while maintaining low computational overhead. This framework is shown to improve performance across various tasks such as classification and object detection. Additionally, \cite{gou2021knowledge} presents a survey focusing on KD as a model compression and acceleration technique, categorizing KD methods by knowledge types, training schemes, and architectures. The paper discusses challenges like the trade-off between model size and performance and suggests potential research avenues to advance the field further.

In another study, Yadikar et al. \cite{yadikar2023review} examine the application of KD in target detection within computer vision, focusing on the challenges of balancing detection speed and accuracy. The study highlights how knowledge compression techniques, particularly knowledge refinement, can enhance the performance of target detection algorithms on edge devices with limited computational power. The authors also propose potential improvements and future trends in integrating distillation learning with target detection \cite{yadikar2023review}. Similarly, Alkhulaifi et al. \cite{alkhulaifi2021knowledge} explore KD as a solution for deploying deep learning models on resource-constrained devices. They introduce a "distillation metric" to compare different KD methods based on model size and accuracy, providing a detailed survey of techniques such as soft label distillation and logit and feature map distillation, both offline and online. The study also discusses real-world KD applications in domains such as autonomous vehicles, healthcare, and IoT, outlining current challenges and future research directions. Furthermore, Yu et al. \cite{yu2023dataset} review dataset distillation (DD), a technique related to KD that focuses on creating smaller, synthetic datasets that retain the performance of models trained on larger datasets. The study presents an algorithmic framework for DD methods, categorizes existing approaches, and identifies challenges such as privacy, copyright, and data storage, offering insights into future research directions for this emerging field.

Additionally, Meng et al. \cite{meng2021knowledge} explore the use of KD in the medical field, addressing challenges such as deploying large models on lightweight devices and the difficulty of sharing medical datasets. The study reviews various KD applications in healthcare, demonstrating how KD can compress complex models while improving their performance in medical tasks. It highlights the potential of KD to alleviate issues related to medical resource shortages by optimizing model deployment effectively. Similarly, Li et al. \cite{li2023object} present a survey on KD in object detection (OD), discussing the evolution of KD-based OD models and their advantages in performance and resource efficiency. The study analyzes different distillation techniques and explores their applications in domains like remote sensing (RS) and the management of 3D point cloud datasets, offering a comprehensive comparison of model performance across various datasets.

Furthering the exploration of KD, Luo et al. \cite{luo2023comprehensive} provide an overview of modern approaches to distilling Diffusion Models (DMs), focusing on distilling DMs into neural vector fields and reviewing stochastic and deterministic implicit generators. The authors also examine accelerated diffusion sampling algorithms as a training-free method for distillation, offering valuable insights for researchers interested in DM distillation. Additionally, Acharya et al. \cite{acharya2024survey} address the emerging field of symbolic KD in large language models (LLMs), emphasizing the transformation of implicit knowledge within these models into a more explicit, symbolic form. This survey categorizes existing research, highlights the importance of symbolic KD in enhancing interpretability and efficiency, and proposes future research directions to advance this growing field.

In the context of computer vision, Kaleem et al. \cite{kaleem2024comprehensive} provide a comprehensive review of KD techniques, covering major methods such as response-based, feature-based, and relation-based knowledge transfer. The study discusses the benefits and challenges of using KD to compress and optimize deep learning models, especially in resource-constrained environments. It explores the application of KD in tasks such as image classification, object detection, and video captioning, and highlights recent developments in multimodal models with KD. Similarly, Habib et al. \cite{habib2023knowledge} focus on KD in Vision Transformers (ViTs), addressing the challenges of deploying these models in environments with limited computational resources. The study reviews various KD approaches for compressing ViTs, emphasizing KD’s role in reducing computational and memory requirements while maintaining model performance. It also provides a comparative analysis of different KD techniques for ViTs and identifies unresolved challenges that warrant further research. Table \ref{tab:comraison} presents a comparison of several KD surveys and reviews across various aspects such as focus on vision tasks, the use of teacher-student frameworks, real-world and medical applications, distillation techniques, and future research directions. It highlights which aspects are covered by each reference, along with the proposed study, indicating areas of focus and gaps in the existing literature.

The proposed review offers a comprehensive and structured analysis of KD, significantly expanding upon previous works by integrating a wide-ranging taxonomy and exploring its diverse applications across various domains, particularly in RS. Unlike existing reviews, which tend to focus on specific aspects of KD such as its role in model compression or its application in computer vision, this review provides a holistic overview, categorizing KD models based on architecture, distillation techniques, and application areas. Furthermore, it delves into advanced topics such as dynamic distillation, layer-wise distillation, and the integration of KD with real-time processing and edge AI—areas that remain relatively underexplored in prior literature. Additionally, the review addresses practical challenges such as data heterogeneity, scalability, and the balance between efficiency and accuracy, offering insights into emerging trends and future directions. This approach not only contextualizes KD within the broader landscape of machine learning but also highlights its potential for innovation in areas like precision agriculture, urban planning, and oceanographic monitoring. Thus, this review serves as a valuable resource for researchers and practitioners aiming to leverage KD in diverse and complex environments. Overall, this review makes several key contributions to the field of knowledge distillation (KD) in RS, which can be briefly summarized into the following:
\begin{itemize}
    \item Provides a comprehensive and structured analysis of KD, significantly expanding on previous works by integrating a wide-ranging taxonomy.
    \item Explores the diverse applications of KD across various domains, with a particular focus on RS.
    \item Categorizes KD models based on architecture, distillation techniques, and application areas, offering a holistic overview.
    \item Delves into advanced topics such as dynamic distillation, layer-wise distillation, and the integration of KD with real-time processing and edge-AI, which are underexplored in prior literature.
    \item Addresses practical challenges, including data heterogeneity, scalability, and the balance between efficiency and accuracy, providing insights into emerging trends and future directions.
    \item Contextualizes KD within the broader landscape of machine learning over RS data, highlighting its potential for innovation in areas like precision agriculture, urban planning, and oceanographic monitoring.
\end{itemize}

\begin{table*}[ht]
    \centering
    \caption{Comparison of KD Surveys and Reviews}
    \label{tab:comraison}
    \begin{tabular}{|lccccccccccc|}
        \toprule
        \textbf{Aspect} & \cite{wang2021knowledge} & \cite{chen2021distilling} & \cite{gou2021knowledge} & \cite{yadikar2023review} & \cite{alkhulaifi2021knowledge} & \cite{yu2023dataset} & \cite{meng2021knowledge} & \cite{li2023object} & \cite{kaleem2024comprehensive} & \cite{habib2023knowledge} & Proposed\\
        \midrule
        \textbf{Focus on Vision Tasks} & \cmark & \xmark & \cmark & \cmark & \xmark & \xmark & \xmark & \cmark & \cmark & \cmark  & \cmark\\
        \textbf{Teacher-Student Framework} & \cmark & \cmark & \cmark & \xmark & \cmark & \xmark & \cmark & \cmark & \cmark & \cmark & \cmark\\
        \textbf{Real-world Applications} & \cmark & \cmark & \cmark & \cmark & \cmark & \xmark & \cmark & \cmark & \cmark & \cmark & \cmark\\
        \textbf{Medical Applications} & \xmark & \xmark & \xmark & \xmark & \xmark & \xmark & \cmark & \xmark & \cmark & \xmark & \cmark\\
        \textbf{RS Applications} & \xmark & \xmark & \xmark & \xmark & \xmark & \xmark & \xmark & \cmark & \xmark & \xmark & \cmark\\
        \textbf{Distillation Techniques} & \cmark & \cmark & \cmark & \cmark & \cmark & \xmark & \cmark & \cmark & \cmark & \cmark & \cmark\\
        \textbf{Discussion on Challenges} & \cmark & \xmark & \cmark & \cmark & \cmark & \cmark & \cmark & \cmark & \cmark & \cmark & \cmark\\
        \textbf{Future Research Directions} & \cmark & \cmark & \cmark & \cmark & \cmark & \cmark & \cmark & \cmark & \cmark & \cmark & \cmark\\
        \textbf{Model Compression Techniques} & \cmark & \cmark & \cmark & \cmark & \cmark & \xmark & \cmark & \cmark & \cmark & \cmark & \cmark\\
        \textbf{Introduction of New Metrics} & \xmark & \xmark & \xmark & \xmark & \cmark & \xmark & \xmark & \xmark & \xmark & \xmark & \cmark\\
        \textbf{Multimodal Model Applications} & \xmark & \xmark & \xmark & \xmark & \xmark & \xmark & \xmark & \xmark & \cmark & \xmark & \cmark\\
        \textbf{Discussion of Existing Datasets} & \xmark & \xmark & \cmark & \xmark & \xmark & \cmark & \cmark & \cmark & \cmark & \cmark & \cmark\\
        \bottomrule
    \end{tabular}
    \label{tab:kd_comparison}
\end{table*}


\color{black}

\subsection{Literature Screening Approach}

\subsubsection{Inclusion and Exclusion Criteria}
The inclusion and exclusion criteria have been identified by including studies that are directly relevant to KD, particularly in the context of RS, with a focus on research published within the last 5 years to capture the latest advancements. Peer-reviewed articles, conference papers, preprints from reputable platforms and book chapters published in English are prioritized, in addition to to empirical studies, reviews, case studies, and theoretical papers. Studies that do not specifically address KD in RS or focus on unrelated technologies, as well as outdated research published more than 5 years ago unless it is seminal are excluded, as well as and non-peer-reviewed sources such as blog posts, opinion pieces, and non-academic publications.

\subsubsection{Search Strategy}
\begin{figure}[t!]
\centering
\includegraphics[width=0.5\textwidth]{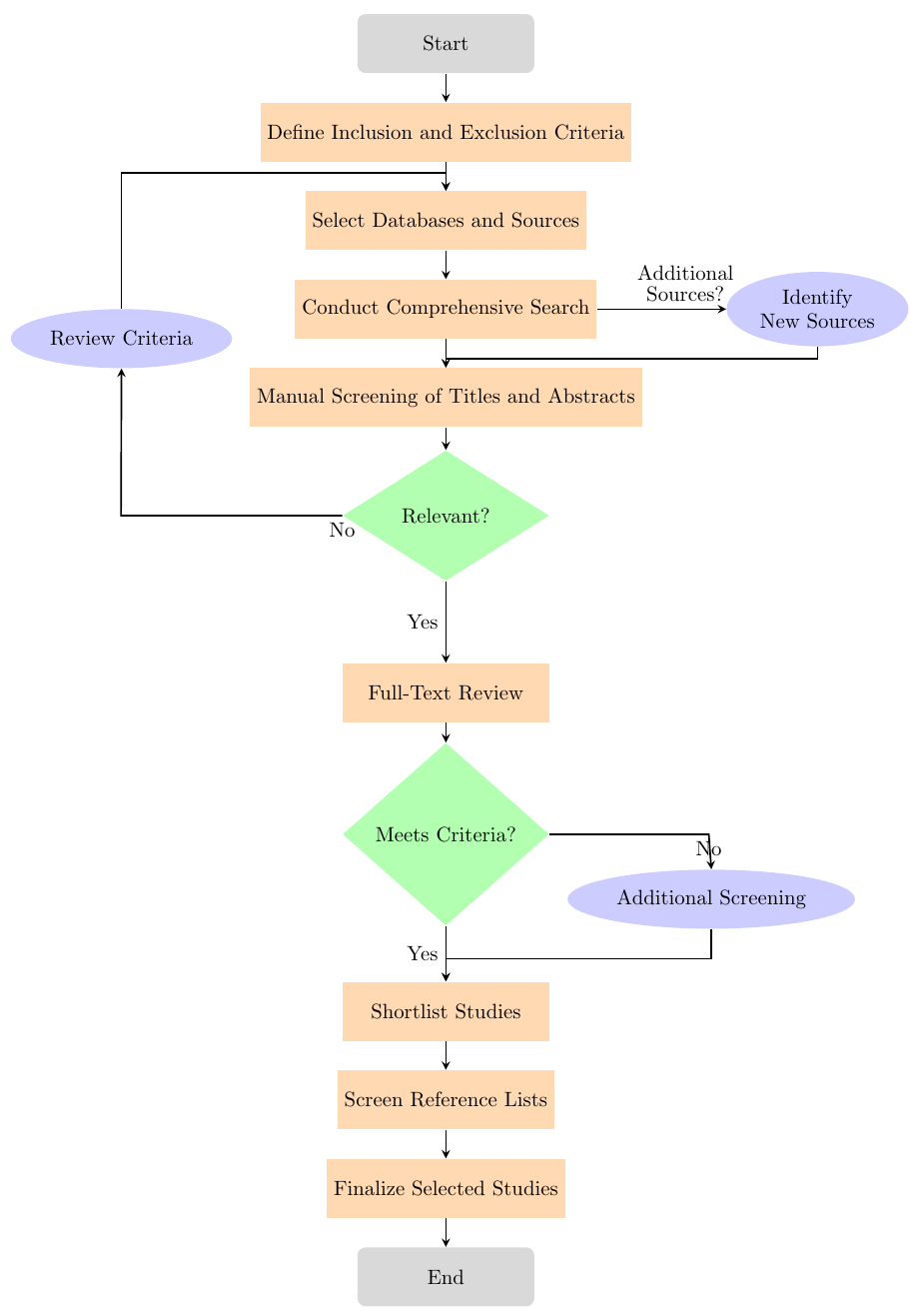} 
\caption{Summary of literature screening approach used in this review.}
\label{search_strategy}
\end{figure}

The search strategy involves using multiple academic databases such as IEEE Xplore, Scopus, Web of Science, Elsevier, Springer Nature, Wiley, Taylor \& Francis, MDPI, Google Scholar, etc. to conduct a comprehensive search using relevant keywords like "Knowledge Distillation", "Model Compression", "Model Distillation", "Feature Distillation", "Data Distillation", "Remote Sensing"  "Urban Planning", "Precision Agriculture", "Land Cover Classification", etc., refined with Boolean operators (AND, OR, NOT). Initial screening of titles, abstracts, and keywords is performed manually to identify potentially relevant studies. Studies that meet the inclusion criteria are shortlisted for a full-text review, where a detailed evaluation confirms their relevance and quality. Additionally, reference lists of selected studies are screened to identify any further relevant studies that may have been overlooked. Fig. \ref{search_strategy} explains the literature screening approach adopted in this study.

\color{black}
\subsection{Organization of the Paper}
The organization of the paper is meticulously structured to provide a thorough exploration of KD and its applications in remote sensing. Section 2 lays the groundwork by covering the fundamentals of KD, starting with a brief overview, defining essential concepts, and discussing the historical evolution of KD techniques. This section also delves into the basic principles and mechanisms of KD, including the objective function and overall loss that guide the distillation process. Additionally, the benefits of KD are highlighted, such as model compression, improved efficiency, enhanced performance on smaller models, and the broader implications for various applications. 
Following this, Section 3 focuses on RS tasks and the public datasets that are pivotal for applying KD in this domain. Section 4 introduces a comprehensive taxonomy of KD models, categorizing them based on variations in the model or input data, the type of transferred knowledge (including response-based, feature-based, and relation-based distillation), distillation targets (data, model, and feature distillation), and structural relationships within network layers (layer-to-layer and cross-layer distillation). 
In Section 5, the paper transitions to discussing the applications of KD in remote sensing, with a detailed examination of its use in image/scene classification, object detection, land cover classification, semantic segmentation, precision agriculture, urban planning, and oceanographic monitoring. Section 6 then addresses the challenges and limitations associated with KD, including model complexity, data heterogeneity, overfitting, scalability, real-time applicability, dependency on high-quality data, balancing efficiency and accuracy, and integration complexity. 
Looking ahead, Section 7 outlines future directions for KD research. It suggests advancements such as dynamic distillation, layer-wise distillation, efficient training and inference techniques, low-cost training algorithms, hardware-aware distillation, and improvements in data quality and robustness. The section also discusses scalability solutions like distributed and incremental distillation, real-time processing enhancements, and the integration of cross-modal and multi-modal distillation. Additionally, it explores the potential for seamless integration with existing workflows through plug-and-play distillation modules, toolkits, and frameworks, as well as enhancing model interpretability through explainable distillation and feature importance preservation. The potential of hybrid approaches, combining KD with other techniques and developing adaptive distillation frameworks, is also considered. 
Finally, Section 8 offers a comprehensive conclusion, synthesizing the insights gained throughout the paper and highlighting the potential for future advancements in the field of KD in remote sensing.

\section{Fundamentals of KD}

\subsection{A Brief Overview}

\subsubsection{Definition and Basic Concepts of KD}

\begin{figure*}[b!]
\begin{center}
\includegraphics[width=0.78\textwidth]{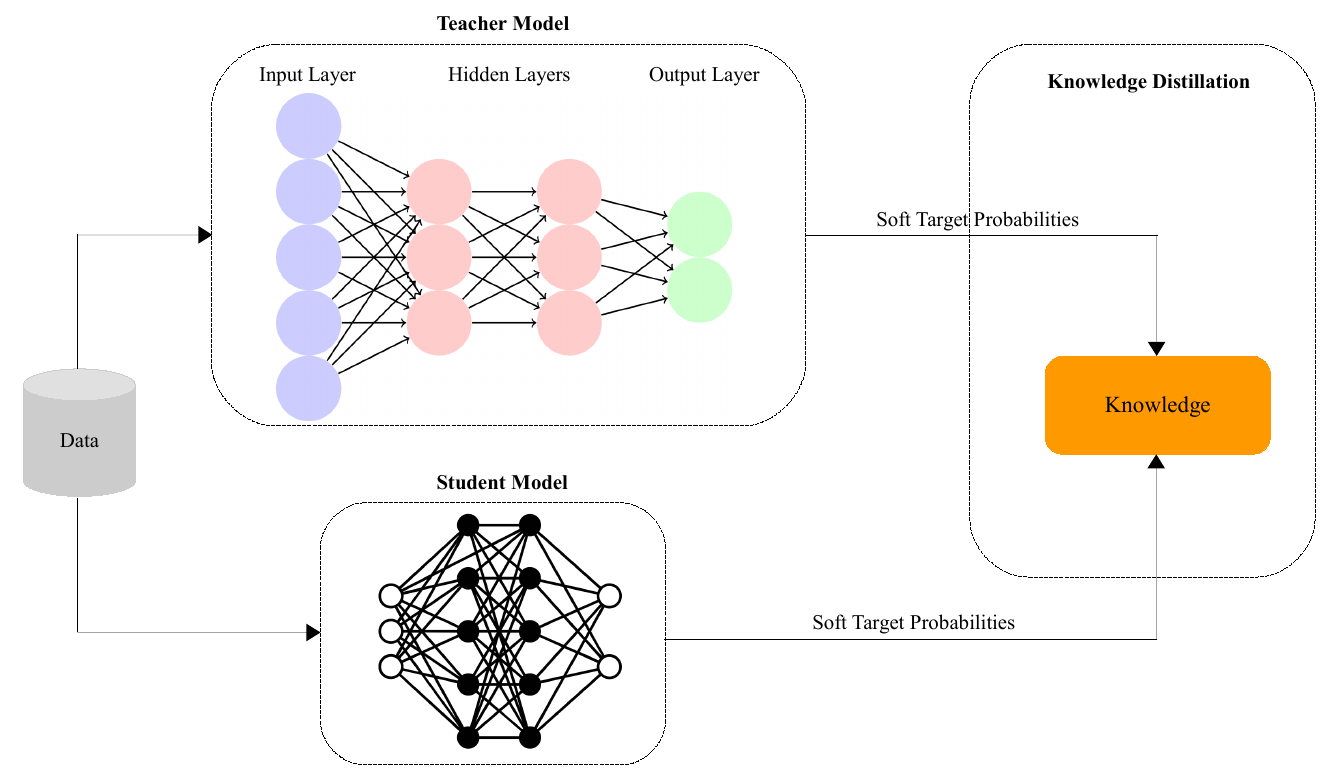}\\
\end{center}
\caption{An overview of the knowledge distillation principle.}
\label{fig:KD_highlevel}
\end{figure*}

KD is a ML technique where a smaller, simpler model (known as the student) is trained to emulate the behavior of a larger, more complex model (known as the teacher) \cite{xu2024double}. \textcolor{black}{As shown in Fig. \ref{fig:KD_highlevel}, KD relies on two deep neural network models, a more complex one that is called the Teacher and a simpler one that is called the Student}. The core idea is to transfer the knowledge from the teacher model, which typically performs better due to its greater capacity, to the student model, which is less resource-intensive \cite{liu2024text} \textcolor{black}{and tries to mimic the teacher's behavior. This can be achieved by aligning the student's outputs with those of the teacher using a Distillation Loss function that compares the two outputs. Usually, the teacher's soft target probabilities (the outputs from the softmax layer before applying the final decision function, \textcolor{black}{as depicted in Fig. \ref{fig:KD_highlevel}}) are used for this purpose.} These soft targets provide richer information than hard labels, as they contain insights about the relative probabilities of incorrect answers, giving the student model clues about the underlying data structure and feature relationships that the teacher model has learned \cite{ma2024knowledge}. \textcolor{black}{However, apart from this response-based knowledge distillation tactic, the student also can learn the output of intermediate teacher layers, or other representations, making the KD approach very flexible and powerful.}

\subsubsection{Brief History and Evolution of the KD Technique}

The concept of KD can be traced back to earlier works in model compression and hints training, where simpler models were trained to mimic more complex ones using additional information from those models. However, the term ``knowledge distillation'' was popularized by Hinton et al. \cite{hinton2015distilling} in a seminal 2015 paper, where they demonstrated the effectiveness of using soft targets to train neural networks. Since then, the field has seen rapid development and broad applications across various domains of Artificial Intelligence. Originally, KD was primarily used to reduce the size and computational demands of large neural networks so that they could be deployed on devices with limited hardware capabilities, such as mobile phones and embedded systems. This was particularly valuable for applications that require real-time processing, such as speech recognition and mobile vision \cite{xue2024feature}.

As research progressed, the scope of KD expanded beyond model compression. Researchers began exploring its potential to improve model generalization by smoothing the decision boundaries, making them more stable and improving generalization. Using ensembles of teacher networks to train a student network further stabilizes training by distilling the collective knowledge of multiple models into a single model, and facilitates transfer learning across different domains or tasks \cite{liu2024sar}. The technique has been adapted and refined to include not just direct output mimicry, but also feature-based and relation-based distillation, where intermediate representations and relationships between data points are also transferred from the teacher to the student \cite{han2024improving}. Today, KD is an active area of research with ongoing innovations that aim to further enhance its effectiveness and expand its applicability. This includes cross-modal distillation for transferring knowledge between different types of data, such as video-to-text, and self-distillation, where a model is iteratively trained on its own softened outputs to refine its capabilities \cite{zhang2024object}.

\subsection{Basic Principle and Mechanism}
This section provides the mathematical background of KD \cite{lu2024weakly}. Let us denote the output logits (pre-softmax activations) of the teacher model as $z^T$ and those of the student as $z^S$. The softmax function applied to these logits is given by:

\begin{equation}
\sigma(z_i, T) = \frac{e^{z_i/T}}{\sum_j e^{z_j/T}},
\end{equation}

\noindent where $i$ indexes the output classes, and $T$ is the temperature parameter that controls the softness of the probability distribution. A higher value of $T$ produces a softer probability distribution \cite{du2024object}.

\subsubsection{Objective Function}
The training of the student network involves minimizing a loss function that typically comprises two terms: the distillation loss and the traditional hard target loss \cite{zhao2024center}.

\begin{itemize}
\item \textbf{Distillation Loss:} This loss measures the difference between the softened outputs of the teacher and the student, encouraging the student to mimic the teacher's generalized behavior. It is often computed using the Kullback-Leibler ($KL$) divergence \cite{miles2024understanding}:

    \begin{equation}
    L_{\text{KD}} = T^2 \cdot KL(\sigma(z^T, T) \| \sigma(z^S, T))
    \end{equation}

The factor of $T^2$ is used to scale the gradients appropriately, as the gradients produced by the softmax function are scaled by $\frac{1}{T}$ \cite{miles2024understanding}.

\item \textbf{Hard Target Loss:} This is a standard loss, such as Cross-entropy (CE), used in training neural networks, calculated between the student's output (at $T=1$) and the true labels \cite{oki2020triplet}:

    \begin{equation}
    L_{\text{CE}} = CE(y, \sigma(z^S, 1))
    \end{equation}

    where $y$ are the true labels.
\end{itemize}

\subsubsection{Overall Loss}
The total loss function used to train the student model is a weighted sum of the distillation and hard target losses:

\begin{equation}
L = \alpha L_{\text{CE}} + (1 - \alpha) L_{\text{KD}}
\end{equation}

\begin{figure*}[t!]
\begin{center}
\includegraphics[width=0.7\textwidth]{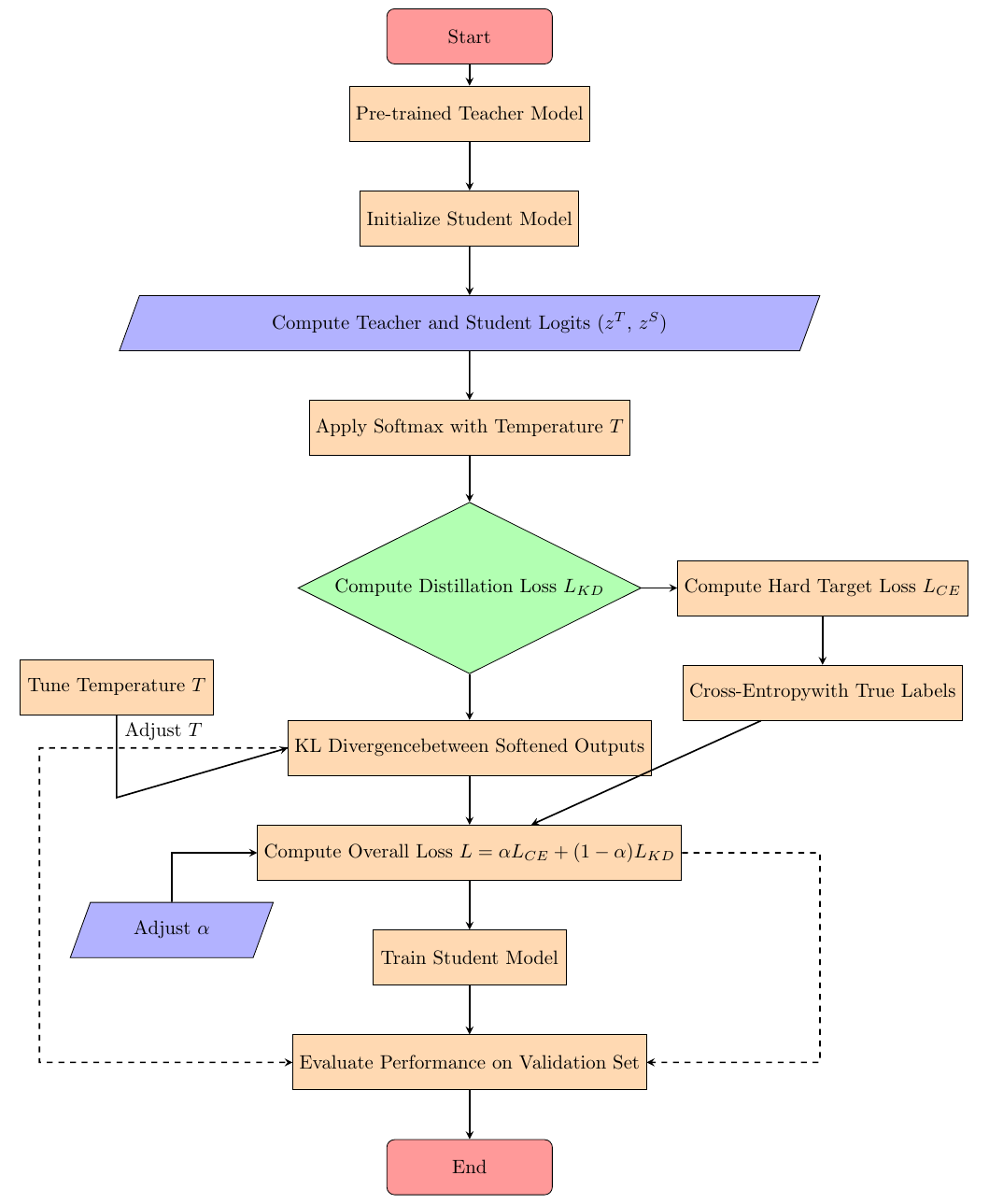}\\
\end{center}
\caption{Principal steps of applying KD in RS applications.}
\label{fig:flowchart_KD}
\end{figure*}
\begin{figure*}[t!]
\begin{center}
\includegraphics[width=1\textwidth]{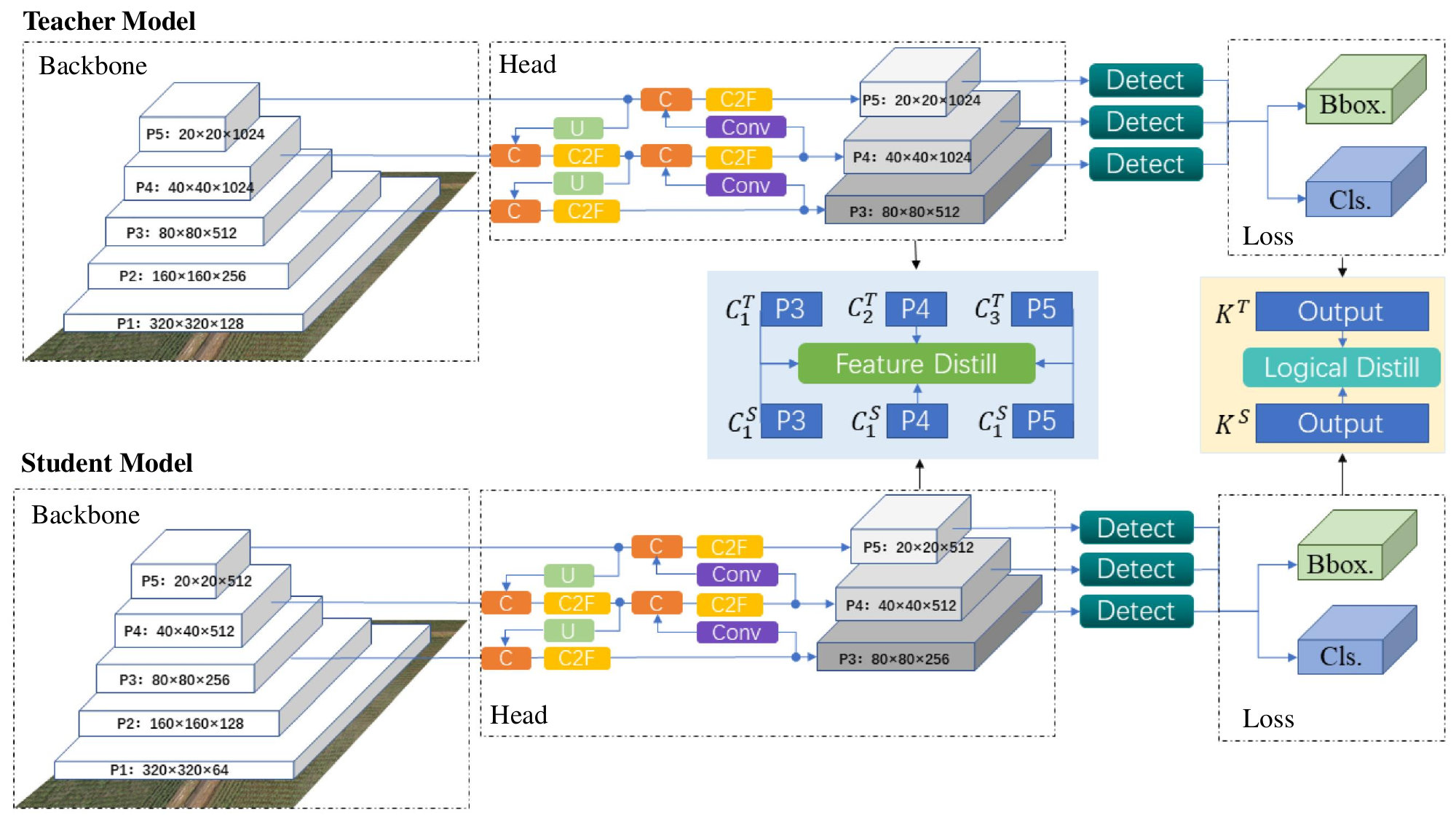}\\
\end{center}
\caption{The YOLOv8n DT network architecture is structured into three primary components: the teacher network, the student network, and the distillation loss function module. This architecture incorporates both feature loss and logit loss within the distillation process to effectively transfer knowledge from the teacher to the student network, thereby enhancing the student's performance while maintaining efficiency.}
\label{fig:guo2024intelligent}
\end{figure*}


\noindent where $\alpha$ is a hyperparameter that balances the importance of the two loss components. By optimizing this loss, the student learns not only the explicit knowledge represented by the class labels but also the implicit, richer information embedded in the teacher's output distribution, thus achieving better generalization from a more compact model \cite{yang2023knowledge_2}. Fig. \ref{fig:flowchart_KD} summarizes the main steps of applying KD in RS applications.
Fig. \ref{fig:guo2024intelligent} \textcolor{black}{illustrates the architecture of a knowledge distillation (KD) framework based on YOLOv8, designed for precision agriculture applications such as weed recognition and variable rate spraying. In this framework, a YOLOv8l model, which has the highest recognition accuracy, was chosen as the teacher network, while a YOLOv8n model, which has the lowest recognition accuracy and the smallest model size, was selected as the student network. The resulting KD model, named YOLOv8n-DT, is specifically tailored for rice field weed recognition and comprises three main components: the teacher network, the student network, and the distillation loss function module, that performs both target and feature distillation.}

\subsection{Benefits of KD}
KD offers several compelling advantages that make it an attractive technique in the field of ML, particularly when deploying models in resource-constrained environments.

\subsubsection{Model Compression}

One of the primary benefits of KD is model compression. Traditional DL models often require substantial computational resources due to their depth and complexity, which limits their deployment on devices with restricted hardware capabilities such as mobile phones, IoT devices, and embedded systems. KD addresses this challenge by enabling the training of smaller, lighter models (students) that mimic the behavior of larger, more complex models (teachers). This process involves transferring the intricate knowledge and insights learned by the teacher model into a more compact form within the student model. The student thereby learns to approximate the function of the teacher but with fewer parameters and lower computational demands. This compression not only reduces the size of the model but also lessens the energy consumption and heat production, which are critical factors for battery-powered devices.

\subsubsection{Improved Efficiency}

Efficiency in model training and inference is another significant advantage of KD. By distilling a cumbersome model into a smaller one, KD effectively reduces the time and computational power needed for training and deploying AI systems. This improved efficiency is particularly beneficial for applications requiring real-time data processing, such as autonomous driving and real-time surveillance. Smaller models also allow for more frequent updates and easier maintenance, which is crucial for systems that need to adapt to changing conditions or data streams.

\subsubsection{Enhanced Performance on Smaller Models}

KD not only compresses the size of the models but often also enhances their performance, especially in smaller models. Typically, smaller neural networks are prone to underfitting and may not capture the complex patterns in large datasets as effectively as their larger counterparts. However, when trained through the KD process, these smaller models inherit refined insights from the teacher models, which include soft probabilities and inter-class relationships that are not visible through traditional hard-label training. This enriched training set helps the student models to perform better than if they were trained independently from scratch. Moreover, the nuanced knowledge transferred includes the handling of edge cases and anomalies, which significantly improves the robustness and generalizability of the student models.

\subsubsection{Broader Implications}

The advantages of KD extend beyond individual model improvements. In educational settings, distillation techniques can democratize access to advanced AI capabilities by enabling more institutions to deploy high-performing AI solutions without the need for expensive infrastructure. Furthermore, in a research context, KD facilitates greater experimental flexibility and faster iteration speeds, accelerating the pace of innovation in AI.

\section{RS Tasks and} Public datasets
RS has become a pivotal tool for monitoring and understanding changes in both urban and agricultural environments. 
\textcolor{black}{RS involves the acquisition and analysis of data from satellite or airborne sensors to observe and interpret features on the Earth's surface. The main RS tasks encompass a variety of applications that leverage spectral, spatial, and temporal information. These tasks include image classification, object detection, change detection, segmentation, and data fusion.}
\textcolor{black}{The primary RS tasks are centered around image classification and analysis. Image classification categorizes pixels in an image into distinct classes, such as different land cover types, using methods like convolutional neural networks (CNNs) for high accuracy. Object detection identifies and locates specific objects within an image, such as vehicles or buildings, making it crucial for applications in urban planning and agriculture. Change detection focuses on identifying differences in images taken at different times, which is essential for monitoring environmental changes like deforestation. Segmentation further refines this process by partitioning an image into meaningful regions, helping extract detailed information about specific features like roads or rivers. Collectively, these tasks highlight that RS primarily involves sophisticated image classification and analysis.}

\textcolor{black}{Data fusion tasks arise from the need to integrate and analyze data from various sources, such as multispectral, hyperspectral, or LiDAR data, to enhance the comprehensiveness and accuracy of RS applications. This integration is vital when dealing with the complex nature of environmental features that cannot be fully captured by a single sensor type. Consequently, data fusion is an essential approach to addressing the limitations of individual datasets, providing a more holistic view of the Earth's surface.}
\textcolor{black}{All the aforementioned tasks are fundamental to various environmental, agricultural, and urban studies, providing essential insights for decision-making and resource management.} During the years, diverse datasets have been developed to support the advancement of instance segmentation techniques in this field, each tailored to specific challenges and applications.

SpaceNet 7 \cite{SpaceNet7} and SpaceNet 4 \cite{van2018spacenet} represent significant contributions to urban development analysis. SpaceNet 7 offers insights into the evolution of building footprints across 100 global locations over two years, using Planet imagery. This dataset is crucial for tracking urban expansion and infrastructure development. Conversely, SpaceNet 4 focuses on the technical challenge of detecting buildings from steep observation angles—up to 54 degrees off-nadir. This is particularly valuable in emergency response situations where quick, accurate assessments are necessary.
Similarly, the Microsoft BuildingFootprints dataset \cite{MicrosoftBuildingFootprints} provides detailed building footprints across several countries, extracted from Bing imagery. This resource supports urban planning and management by offering extensive building delineations. Additionally, the xView 2 Building Damage Assessment Challenge \cite{Gupta2019xView2} leverages high-resolution Worldview-3 imagery to assess building damage from natural disasters, a critical component of effective disaster response.
In the agricultural sector, datasets like PASTIS \cite{garnot2022multi} and the Agriculture-Vision Database \cite{chiu20201st, chiu2020agriculture} are invaluable. PASTIS provides panoptic labels for over 124,000 agricultural parcels in France, captured across Sentinel-2 timeseries images. This dataset aids in the precise monitoring and management of agricultural lands. The Agriculture-Vision challenge, on the other hand, focuses on identifying field anomalies from aerial imagery across the United States, promoting enhanced agricultural practices through detailed monitoring.

For more specialized applications, datasets like RarePlanes \cite{shermeyer2021rareplanes}, which includes both synthetic and real data for plane detection, and iSAID \cite{waqas2019isaid}, which covers a wide range of categories from planes to bridges, are particularly noteworthy. RarePlanes is essential for developing models that differentiate between aircraft types, useful in both civilian and defense sectors. iSAID facilitates broad applications in aerial image analysis by providing extensive annotations for diverse objects.
Furthermore, the introduction of SpaceNet 6: Multi-Sensor All-Weather Mapping \cite{SpaceNetAWM} combines SAR data and optical imagery to enhance building footprint detection in challenging weather conditions, illustrating the value of multi-sensor data integration in RS.
The technological advancements in datasets like Airbus Ship Detection Challenge \cite{AirbusShipDetection}, which focuses on ship detection using satellite imagery, and novel methodologies in the LPIS agricultural field boundaries dataset highlight the industry's shift towards more sophisticated and fine-grained analysis capabilities.
The PASTIS dataset \cite{pastis2022} provides detailed panoptic labels for over 124,000 agricultural parcels across France, captured in 2,433 Sentinel-2 image timeseries. This dataset is instrumental for applications in agricultural monitoring, allowing for the differentiation of crops at the parcel level through both instance and semantic segmentation. It is particularly useful for tracking changes in agricultural land over time.


\section{Taxonomy of KD Models}
\begin{figure*}[ht!]
\begin{center}
\includegraphics[width=0.9\textwidth]{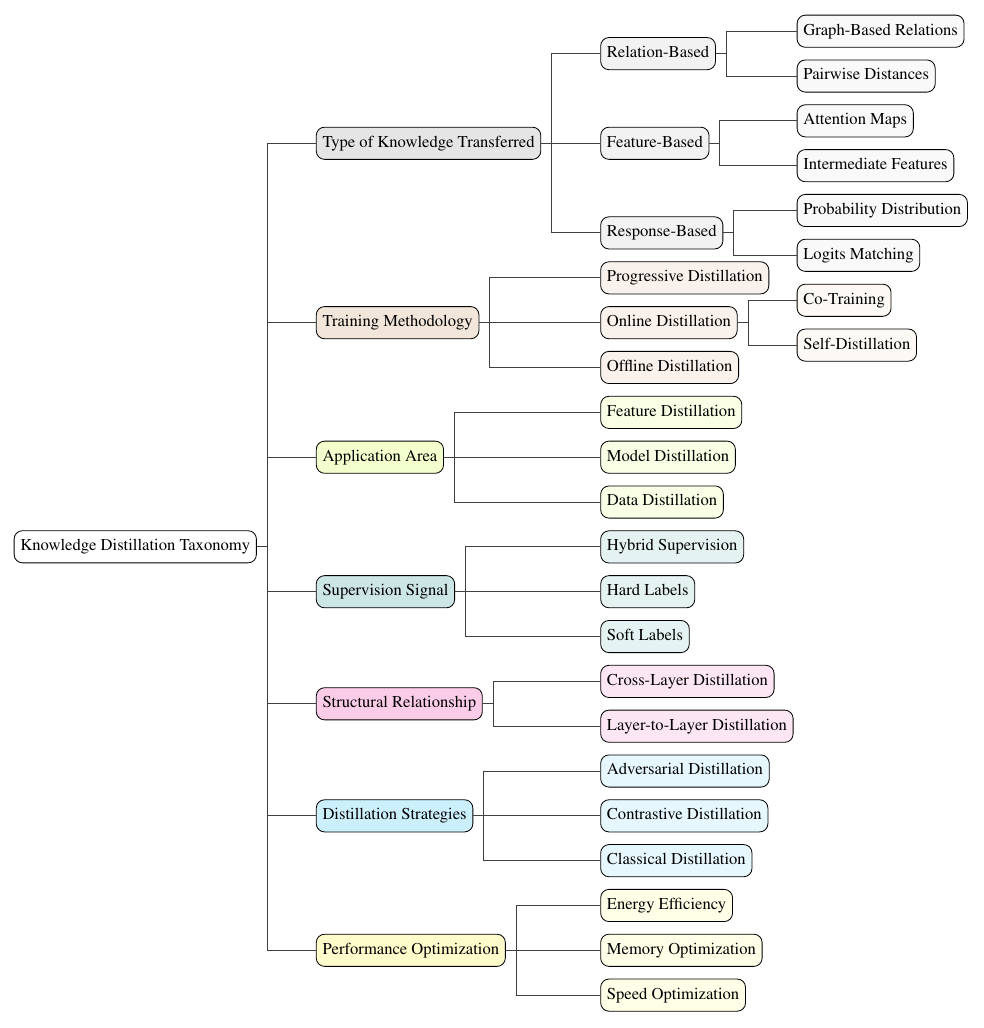}\\
\end{center}
\caption{A Comprehensive Taxonomy of Existing KD Techniques.}
\label{fig:taxonomy_of_KD_techniques}
\end{figure*}

KD methods in RS (RS) can be categorized into several key approaches, each with unique attributes and applications. As depicted in Fig. \ref{fig:taxonomy_of_KD_techniques}, the \textcolor{black}{variations may come from the differences in the data or architecture used by the teacher and student networks resulting to Heterogeneous and Cross-modal KD approaches that are based on the Teacher-Student Architecture, or from the different types of knowledge that are distilled between the teacher and the student, resulting to Response-based, Feature-based, and Relation-based approaches}. These approaches are tailored to optimize RS models by transferring knowledge from complex teacher models to more efficient student models, \textcolor{black}{using all the available data per case}, thereby enhancing performance in tasks such as object detection, scene classification, and image segmentation. \textcolor{black}{Of course, there are several more variations that depend on the training methodology, the application area, the structural representation, the distillation strategy, etc., as shown in Fig. \ref{fig:taxonomy_of_KD_techniques} and explained in the following.}

\subsection{Varying the Model or Input Data}

\subsubsection{Heterogeneous KD}
Heterogeneous KD (HKD) is a method of transferring knowledge from a teacher model to a student model where the teacher and student models have significantly different architectures \cite{9157683}. Traditional KD methods typically assume that the teacher and student models have similar architectures, which allows for straightforward layer-by-layer transfer of knowledge. However, in HKD, the architectures may vary greatly, posing a challenge for direct knowledge transfer.

The study in \cite{ienco2020generalized} presents a Generalized KD (GKD) framework for multi-source Earth Observation analysis, specifically for land cover mapping using radar and optical satellite image time series data. This approach tackles data misalignment due to atmospheric conditions or acquisition costs, using radar data consistently and treating optical data as privileged information. This makes it a case of heterogeneous distillation, where different modalities (radar and optical) are involved, requiring the student model to adapt to a less data-rich environment at test time compared to training. The authors in \cite{tian2021knowledge} propose using information from deep convolutional networks to guide the training of shallow Grassmannian manifold networks, addressing the need for high-performance yet small-sized networks in resource-limited scenarios. The approach bridges DL with manifold learning, fitting well within the heterogeneous category, as it involves transferring knowledge between fundamentally different architectures. Moving forward, Yang et al. \cite{yang2024two} introduce a two-way assistant distillation method for lightweight object detection in RS. This method incorporates compression and multiscale adaptive modules to address feature disparities and background noise, utilizing a heterogeneous distillation approach by applying complex operations from larger models to enhance smaller, simpler ones.

Besides, Nabi et al. \cite{nabi2022cnn} propose a compound loss computed on a Transformer-based student and a CNN teacher for single-label scene classification in RS. The use of heterogeneous architectures, where a CNN and a Transformer are involved, leverages the long-range visual capabilities of the Transformer and the inductive biases of the CNN, aiming to enhance classification accuracy in complex scenes.
Similarly, the research in \cite{ma2021cross} involves a teacher-ensemble learning approach using KD in cross-source content-based image retrieval for high-resolution RS images. The method combines source-shared and source-specific classifiers, constructing an effective heterogeneous ensemble of teacher models to transfer useful information to the student model.

\subsubsection{Cross-Modal KD}
Cross-modal KD (CMKD) refers to the process of transferring knowledge from a model trained with superior modalities (e.g., depth maps or point clouds) to another model trained with weaker modalities (e.g., RGB images) \cite{zhao2020knowledge}. The goal is to improve the performance of the student model trained on the weaker modality by leveraging the knowledge from the teacher model trained on the superior modality. This transfer is achieved by aligning the intermediate feature representations and activation maps between the teacher and student models \cite{zhao2020knowledge}. In CMKD, the knowledge from the teacher model is used as an additional supervision signal to guide the training of the student model, enhancing its learning process and performance.

Expanding on this concept, Geng et al. \cite{geng2020topological} propose a topological space network for road extraction, where a denser teacher network focused on topological feature extraction guides a lighter student network. This distillation process transfers knowledge about complex road topology from a heavy network, illustrating a clear case of cross-modal architecture distillation by integrating high-dimensional topological features into a simplified network. Similarly, Xiong et al. \cite{xiong2020discriminative} introduce a discriminative distillation network for cross-source Content-Based RS Image Retrieval (CBRSIR), addressing the challenge of harmonizing features between multispectral and panchromatic images, thereby further exemplifying cross-modal architecture by handling variations between different types of RS data sources. Additionally, Liu et al. \cite{liu2022multispectral} propose a cross-modal KD framework designed to improve multispectral scene classification by transferring knowledge from teacher models pre-trained on RGB images to a student model processing multispectral images. This approach highlights the adaptability of CMKD by addressing the differences between modalities and enhancing the student's performance, particularly in scenarios with limited samples.

Furthermore, Pande et al. \cite{pande2019adversarial} contribute to the field with an adversarial training-driven hallucination architecture for modality distillation in RS image classification, focusing on learning discriminative feature representations from multiple sensor modalities, even in the presence of missing data during the model inference phase. This work aligns closely with cross-modal architectures as it effectively distills features across varying sensor modalities, enhancing model robustness. Lastly, Liu et al. \cite{liu2023distilling} present a universal Super-Resolution-Assisted Learning (SRAL) framework aimed at improving the performance and efficiency of salient object detection in RS images. By incorporating super-resolution techniques into a multitask learning framework, this approach distills domain knowledge from the super-resolution task to significantly boost object detection performance, further showcasing the potential of cross-modal knowledge transfer in enhancing model accuracy and efficiency.

\subsection{Varying the Type of Transferred Knowledge}

\subsubsection{Response-Based (Soft Targets Distillation)}


Response-based KD focuses on the knowledge extracted from the final layer of the teacher model. It aims to align the final predictions between the teacher and the student models. The primary goal is outcome-driven learning, which involves distilling the class probability distribution via a softened softmax function, known as 'soft labels.' This method guides the student model by matching the output distributions of the teacher and student models using various distance functions such as Kullback-Leibler divergence, mean squared error, or Pearson correlation coefficient \cite{yang2023categories}.

The study in \cite{chen2018training} introduces a KD framework applied to RS scene classification. By using the high-temperature softmax outputs from a large, deep teacher model to train a smaller, shallow student model, the study showcases how KD can improve the performance of less complex models on multiple public datasets, increasing accuracy significantly even on smaller and unbalanced datasets. This approach directly employs the response-based distillation technique by leveraging the teacher's softened output probabilities to enhance the student's learning process. Moving on, Zhao et al. \cite{zhao2022pair} introduces a novel pairwise similarity KD method for reducing the complexity of CNN models in RS image scene classification, maintaining accuracy while using less computational resources. This method focuses on distilling discriminative information between sample pairs.

\subsubsection{Feature-Based (Intermediate Representations)}
Feature-based KD addresses the limitation of response-based KD by providing supervision at intermediate layers of the network. This method focuses on transferring intermediate feature representations, such as feature maps, attention mechanisms, activation boundaries, and probability distributions, from the teacher to the student model \cite{yang2023categories}. The goal is to ensure that the student model learns more meaningful semantic information throughout its hidden layers. The distillation loss in feature-based KD measures the similarity between the transformed intermediate features of the teacher and student models using various distance functions. For example, Chen et al. \cite{chen2020incremental} introduce a new architecture for incremental object detection in RS, which utilizes a Feature Pyramid Network (FPN) to handle objects of various sizes and orientations. Importantly, the study incorporates KD to maintain previously learned information during incremental learning, applying it to outputs from different layers of the FPN. This approach is particularly aligned with feature-based KD, as it focuses on preserving and transferring detailed feature representations across various scales and model iterations, which is essential for detection tasks in dynamically changing datasets. Similarly, to address the efficiency challenge in lightweight object detectors, Yang et al. \cite{yang2022adaptive} propose a training method called adaptive reinforcement supervision distillation (ARSD). This method enhances lightweight models by using a multiscale core features imitation (MCFI) module and a strict supervision regression distillation (SSRD) module, improving feature selection and regression accuracy during training. Furthermore, Li et al. \cite{li2022remote} introduce a dual KD model that incorporates dual attention and spatial structure modules to enhance the local feature extraction and high-level semantic representation abilities of lightweight CNN models for RS image scene classification. This approach, where knowledge in the teacher network about dual attention and spatial structure is transferred to the student network, directly targets the transfer of intermediate representations. In the same vein, Wang et al. \cite{wang2023efficient} propose a fine-grained object recognition method for high-resolution RS images, which utilizes two stages of KD, emphasizing efficient fine-grained object recognition through feature learning and category correction. 

Moreover, Shin et al. \cite{shin2023multispectral} focus on transferring detailed spectral feature representations from a teacher model to a student model, allowing the latter to perform complex scene classification tasks traditionally dependent on multispectral data inputs, using simpler RGB inputs. The approach involves intricate feature imitation and retention of critical spectral information, spanning across different data modalities, making it a clear example of feature-based KD. Similarly, Chi et al. \cite{chi2022novel} propose a self-supervised learning method with KD for hyperspectral image classification, focusing on generating soft labels for unlabeled samples by considering spatial and spectral distances, fitting well under feature-based distillation as it involves creating and transferring complex spectral features. Additionally, Jiang et al. \cite{jiang2018deep} introduce the Deep Distillation Recursive Network (DDRN) for satellite image super-resolution, utilizing ultra-dense residual blocks and a multi-scale purification unit to enhance feature sharing and compensation, particularly focusing on high-frequency components in image super-resolution. Yuan et al. \cite{yuan2022buildings} also contribute by proposing a CNN framework for building change detection that uses self-attention KD strategies to refine features for detecting changes in building regions. This approach integrates globally changed information, emphasizing intermediate feature enhancement and integration for improved accuracy in change detection. Finally, Liu et al. \cite{liu2021zoominnet} introduce ZoomInNet, a cross-scale KD method designed to improve the detection of small objects in drone-based imagery, which often presents complex and dynamic backgrounds. Utilizing a feature pyramid network, the method trains teacher and student networks with differently scaled images to enhance feature harmonization across scales, incorporating layer adaptation, feature level alignment, and an adaptive key distillation point algorithm to refine and distill essential features, showcasing significant advancements in the precision of object detection.

\subsubsection{Relation-Based (Learning Relationships Between Different Data Layers)}

Relation-based KD explores the relationships between different data samples or across different layers within the neural network. Unlike response-based and feature-based KD, which typically handle individual samples, relation-based KD captures cross-sample or cross-layer relationships as meaningful knowledge \cite{yang2023categories}. This method constructs relational graphs to model dependencies and similarities between instances or layers and uses similarity metrics and distance functions to measure these relationships. The goal is to transfer structured knowledge that encapsulates higher-order dependencies and interactions within the dataset.

Chen et al. \cite{chen2023consistency} develop consistency- and dependence-guided KD methods for object detection in RS images. They introduce modules that focus on extracting and transferring discriminative spatial locations and channels, as well as establishing the consistency and dependence of features between the teacher and student models. This approach utilizes relation-based distillation by focusing on the inter-layer and inter-feature relationships to guide the student model's learning process. moving on, Li et al. \cite{li2023instance} introduce an instance-aware distillation method, which combines feature-based and relation-based distillation techniques. The method enhances the student model's performance by focusing on instance-related foreground information and constructing relationships between different instances to improve detection accuracy in complex remote-sensing images. Zhao et al. \cite{zhao2022remote} propose a self-supervised KD network (SSKDNet) that uses feature maps of the backbone as supervision signals and transfers the "dark knowledge" through KD. This method focuses on enhancing the discriminative feature extraction capabilities by learning the relationships between different data layers in a self-supervised setting.

Dong et al. \cite{dong2023distilling} present a cross-model KD framework, distilling segmenters from CNNs and transformers, which uses a channel-weighted attention-guided feature distillation and a target–nontarget KD module to guide the student model in learning complex representations and decision boundaries. This study distinctly focuses on relation-based distillation by leveraging the interdependencies of features and classification decisions between different network architectures.
On the other hand, Zhou et al. \cite{zhou2024mstnet} introduce the Multi-level Semantic Transfer Network (MSTNet), a KD framework designed for dense prediction of RS images. This network utilizes a Multi-level Semantic Knowledge Alignment (MSKA) framework to distill semantic information from a complex teacher model to a more compact student model. The MSKA framework emphasizes cross-layer semantic alignment, dynamic semantic aggregation, and softening learning to adaptively transfer knowledge and optimize the learning of semantic information, thus addressing the complexities of deploying models in practical scenarios.

\subsection{Varying Distillation Target}

\subsubsection{Data Distillation}
Data distillation refers to techniques that aim to synthesize small, high-fidelity data summaries which capture the most important knowledge from a given dataset \cite{sachdeva2023data}. These distilled summaries are optimized to serve as effective substitutes for the original dataset in various data-usage applications such as model training, inference, and architecture search. The goal is to create a concise representation of the data that maintains its critical characteristics, allowing for faster and more efficient model training and evaluation \cite{sachdeva2023data}.

Building on this concept, Zhang et al. \cite{zhang2020remote} introduce a novel noisy label distillation method within an end-to-end teacher-student framework, which distills knowledge from labels across various noise levels. This approach exemplifies data distillation by effectively utilizing knowledge from noisy data to improve classification performance in RS image scene classification. Extending the application of data distillation, Zhao et al. \cite{zhao2022pair} propose a pair-wise similarity KD method for RS image scene classification. By distilling discriminative information from a cumbersome model to a compact model, this study aims to maintain high accuracy while reducing model complexity, demonstrating another facet of data distillation. Furthermore, Yue et al. \cite{yue2021self} contribute to this field with a self-supervised learning method that incorporates adaptive distillation for hyperspectral image classification. Their approach, which focuses on generating adaptive soft labels based on spatial-spectral similarity, underscores the importance of utilizing extensive unlabeled data in the data distillation process.

\subsubsection{Model Distillation}
Model distillation refers to the process of replacing a complex ML model with a simpler model that approximates the original model's performance \cite{boix2024towards}. This technique is used to improve computational efficiency by distilling large or ensemble models into smaller, more manageable models that maintain similar accuracy. The primary goal is to reduce the computational cost associated with deploying large models while preserving their predictive capabilities \cite{boix2024towards}. Model distillation also aids in model interpretability by converting ``black-box'' models, such as neural networks, into more transparent forms.

In the context of model distillation for RS applications, a variety of approaches have been developed to enhance the performance and efficiency of lightweight models. Zhang et al. \cite{zhang2021learning} introduce a dynamic knowledge distillation (KD) framework that enables CNN models to be lightweight while maintaining high detection accuracy, with an emphasis on selective learning through a dynamic instance selection distillation module. Building on the concept of model distillation, Yang et al. \cite{yang2023knowledge} develop a lightweight semantic segmentation network that combines KD with a multiscale pyramidal pooling module and attention mechanisms, resulting in a pruned model that retains high accuracy. Similarly, Wang et al. \cite{wang2024knowledge} propose a change detection method that integrates prototypical contrastive distillation and channel-spatial-normalized distillation, allowing the student model to learn complex feature distributions from the teacher, thereby fitting into the model distillation framework.

Further advancing the field, Chen et al. \cite{chen2023building} propose a multi-teacher collaborative distillation approach that uses adaptive weight and feature knowledge exchange to enhance the robustness of student models, while Gu et al. \cite{gu2023learning} introduce a Context-aware Dense Feature Distillation (CDFD) strategy for CubeSat-based RS object detection, integrating multiple teacher networks to optimize a lightweight detector. Chai et al. \cite{chai2020compact} contribute to the model distillation category with their Bidirectional Self-Attention Distillation (Bi-SAD) approach, aimed at enhancing cloud detection models by enabling compact models to learn detailed textural and semantic information.

Addressing the challenge of few-shot learning, Liu et al. \cite{liu2021integrating} present a ranking-preserving KD method that improves the generalization capabilities of student models in RS scene classification. Similarly, Wang et al. \cite{wang2024psekd} explore the enhancement of lightweight models through a Phase-shift encoded KD method (PseKD) that improves object orientation prediction. In a broader application, Chen et al. \cite{chen2023semi} propose a semi-supervised KD framework for global-scale urban object mapping, emphasizing the handling of urban diversity and large-scale sample growth.

Complementing these efforts, Zhao et al. \cite{zhao2023weakly} propose a weakly correlated distillation learning framework for RS object recognition with limited samples, leveraging large-scale natural image datasets to enhance small-scale RS datasets. Lin et al. \cite{lin2022lightweight} address the issue of denoising by presenting a lightweight model that uses KD to efficiently extract spatial and spectral features while maintaining computational efficiency. Yu et al. \cite{yu2024incremental} focus on incremental learning, introducing a dual KD method to mitigate catastrophic forgetting, which aligns with the incremental learning approach proposed by Xu et al. \cite{xu2023robust} and Xu et al. \cite{xu2022hyperspectral}, who use KD to enhance multimodal learning and hyperspectral image classification, respectively.

Lastly, Zhou et al. \cite{zhou2023gsgnet} introduce a graph semantic guided network (GSGNet) for optical RS scene analysis, utilizing knowledge refinement to maintain high inference speed and contextual inference capability. Zhao et al. \cite{zhao2022target} propose a target detection model distillation framework that uses feature transition and label registration to improve the learning ability of lightweight networks in RS imagery, further contributing to the body of work on model distillation.

\subsubsection{Feature Distillation}
Feature distillation refers to a method in which the student network learns to mimic the hidden feature values of a teacher network \cite{9009585}. This process involves transferring the intermediate representations (features) learned by the teacher network to the student network. Unlike traditional KD that focuses on the output probabilities (logits), feature distillation emphasizes the transfer of internal activations or feature maps. The primary goal is to improve the student network's performance by leveraging the knowledge encapsulated in the teacher's feature representations \cite{9009585}.

Building upon this concept, Zhou et al. \cite{zhou2023graph} propose a lightweight student network framework for semantic segmentation of high-resolution RS images. By employing a graph attention guidance network, they distill knowledge from a large teacher network to optimize image features, thereby enhancing segmentation accuracy. This method aligns with feature distillation, where the objective is to boost the student's feature representation capabilities to closely match those of the teacher. Similarly, Zhang et al. \cite{zhang2021rs} introduce a few-shot classification method for RS scene classification, which also falls under the feature distillation category. This approach utilizes a novel two-branch network and incorporates self-KD during training to generate powerful representations, prevent overfitting, and enhance overall performance.

In parallel, Hu et al. \cite{hu2022variational} contribute to the field with a variational self-distillation network designed for RS scene classification. This method hierarchically distills class entanglement information from deep to shallow layers, further illustrating the application of feature distillation by refining and transferring feature information across different network layers. Expanding on these ideas, Xing et al. \cite{xing2022collaborative} present a collaborative consistent KD method aimed at improving classification accuracy for RS image scenes on embedded devices. Their approach emphasizes feature distillation across multiple network branches, focusing on reducing parameter redundancy and enhancing model efficiency, thus reinforcing the relevance of feature distillation in RS applications.


\begin{table*}[t!]
\centering
\caption{Summary of Studies on KD in RS}
\scriptsize
\begin{tabular}{|p{0.6cm}|p{2.5cm}|p{3.5cm}|p{2.5cm}|p{1.5cm}|p{2cm}|p{2.5cm}|}
\hline
\textbf{Ref.} & \textbf{Model Used} & \textbf{Main Contribution} & \textbf{Database} & \textbf{Task/Appl.} & \textbf{Best Performance Value} & \textbf{Limitation} \\ \hline
\cite{chen2018training} & Small and shallow student models & Introduced a KD framework for scene classification. & AID, UCMerced, NWPU-RESISC, EuroSAT & Scene Classification & Increased accuracy by 1\% to 5\% & Performance on small and unbalanced datasets \\ \hline
\cite{yang2022adaptive} & Lightweight object detector & Developed ARSD to enhance detection capability through feature and regression distillation. & DOTA, DIOR, NWPU VHR & Object Detection & Outperforms SOTA methods & Noise in training due to complicated backgrounds \\ \hline
\cite{chen2023consistency} & Consistency and dependence-guided model (CDKD) & Improved object detection with structured discriminative modules and consistency techniques. & RSOD & Object Detection & 92\% mean average precision & High model volume and computation in RS images \\ \hline
\cite{chen2020incremental} & Incremental learning model with FPN & Employed feature pyramid and KD for incremental learning in object detection. & Various RS datasets & Object Detection & Comparative performance to SOTA & Challenges with object size diversity and directions \\ \hline
\cite{zhang2021learning} & Dynamic KD (DKD) & Developed a dynamic KD framework to improve model performance on edge devices. & DOTA, NWPU VHR-10 & Object Detection & SOTA performance & Complex model deployment on low-computation devices \\ \hline
\cite{zhou2023graph} & GAGNet with KD & Utilized graph attention and dense fusion for semantic segmentation. & Potsdam, Vaihingen & Semantic Segmentation & Excellent performance on datasets & Resource-intensive model deployment \\ \hline
\cite{zhang2021rs} & RS-SSKD for few-shot classification & Introduced a two-branch network with self-KD for few-shot classification. & NWPU-RESISC45, RSD46-WHU & Scene Classification & Surpasses current SOTA & Requires high model adaptability to new data \\ \hline
\cite{li2023instance} & Instance-aware distillation (InsDist) & Combined feature-based and relation-based KD for object detection. & DIOR, DOTA & Object Detection & Noticeable gains over other methods & Integration complexity with existing detectors \\ \hline
\cite{hu2022variational} & Variational self-distillation network (VSDNet) & Implemented a VKT module for robust and end-to-end optimization. & Multiple RS datasets & Scene Classification & Significant improvement over backbones & Managing uncertainty and perturbation in images \\ \hline
\cite{xing2022collaborative} & Collaborative consistent KD (CKD) & Designed a KD method for high classification accuracy on embedded devices. & SIRI-WHU, NWPU-RESISC45 & Scene Classification & 0.943 and 0.916 average accuracy & Redundancy and parameter management on devices \\ \hline

\cite{li2022remote} & DKD Model with DA and SS & Dual KD with dual attention and spatial structure modules & AID, NWPU-45 & RSI Scene Classification & Improved accuracy by 7.57\% and 7.28\% & Model complexity and computational cost \\ \hline
\cite{liu2023distilling} & SRAL Framework & Super-resolution-assisted learning for salient object detection & Multiple datasets & Object Detection in RSIs & Superior to 20+ algorithms & High computational cost of high-resolution processing \\ \hline
\cite{wang2023efficient} & Oriented R-CNN, CF-ORNet & Two-stage fine-grained object recognition with KD & VEDAI, HRSC2016 & Object Recognition in HR-RSIs & Competitive performance & Limited by size of geospatial objects \\ \hline
\cite{zhao2022remote} & SSKDNet & Self-supervised KD network for feature learning & Multiple datasets & Scene Classification & Effective feature extraction & Difficulty in training self-supervised networks \\ \hline
\cite{yang2023knowledge} & KD-MSANet & Lightweight semantic segmentation with multiscale pooling and attention & Vaihingen, Potsdam & Semantic Segmentation & Accuracy near 99.30\% of teacher model & Reduced model size might impact some complex scene parsing \\ \hline
\cite{wang2024knowledge} & CDKD Method & Change detection with prototypical contrastive and channel-spatial-normalized distillation & Public CD datasets & Change Detection & Comparable to large models & Requires careful tuning of distillation parameters \\ \hline
\cite{zhang2020remote} & NLD Method & Noisy label distillation for robust training on noisy datasets & UC Merced Land-use, NWPU-RESISC45, AID & Scene Classification & Outperforms directly fine-tuning methods & Performance variability with noise levels \\ \hline
\cite{dong2023distilling} & DSCT Framework & Cross-model KD from CNNs and transformers for semantic segmentation & ISPRS Potsdam, Vaihingen, GID, LoveDA & Semantic Segmentation & Outperforms state-of-the-art KD methods & Complexity of integrating CNNs and transformers \\ \hline
\cite{shin2023multispectral} & MS2RGB-KD & MS-to-RGB KD for scene classification using RGB images & EuroSAT & Scene Classification & Effective compared to KD baselines & Dependent on quality of MS teacher model \\ \hline
\cite{zhou2024mstnet} & MSTNet with MSKA & Dense prediction using multilevel semantic transfer and KD & Vaihingen, Potsdam & Dense Prediction in RSIs & Excellent performance with reduced parameters & Balancing between model complexity and performance \\ \hline

\end{tabular}
\label{tab:study_summary}

\end{table*}


\subsection{Varying the Structural Relationship of Network Layers}

\subsubsection{Layer-to-Layer Distillation}
Layer-to-layer distillation refers to the process where the teacher model's intermediate layers directly guide the corresponding layers of the student model. This method ensures that the student model learns similar feature representations as the teacher model at different stages of its depth \cite{wu2020skip,chang2024colld}.

\textbf{Direct Mapping:} In this approach, each layer of the teacher model is aligned with the corresponding layer in the student model. The outputs of each intermediate layer in the teacher model are used as targets for the corresponding layer in the student model. This direct mapping can help the student model learn hierarchical features similar to those learned by the teacher model \cite{deepa2023knowledge}.

\textbf{Feature Representation:} By mimicking the intermediate representations of the teacher, the student model can capture complex features and patterns, which might be difficult to learn solely from the final output. This method is particularly useful when the student model has a similar or reduced architecture compared to the teacher.

\textbf{Loss Function:} Often, additional loss terms are introduced to minimize the difference between the teacher's and student's intermediate layer outputs. This can include mean squared error (MSE) or other similarity measures.

Suppose a deep CNN is used as the teacher model with layers: $T_1, T_2, T_3, \ldots, T_n$. The student model has corresponding layers: $S_1, S_2, S_3, \ldots, S_n$. During training, the output of $T_1$ will guide $S_1$, $T_2$ will guide $S_2$, and so on, ensuring that each student layer learns to mimic the feature maps of the corresponding teacher layer.

\subsubsection{Cross-Layer Distillation}
Cross-layer distillation refers to the process where the teacher and student models do not have a direct correspondence between layers. Instead, the knowledge transfer happens between non-matching layers, for example, higher layers of the teacher model guiding lower layers of the student model or vice versa.

\textbf{Non-Matching Layers:} In this approach, there is no strict one-to-one correspondence between the layers of the teacher and the student. The knowledge from higher (more abstract) layers of the teacher model can be distilled into lower (more detailed) layers of the student model, allowing for flexible guidance. Chen et al. \cite{chen2021cross} propose Semantic Calibration for Cross-layer Knowledge Distillation (SemCKD), which automatically assigns target layers from a teacher model to each student layer using an attention mechanism. This method allows student layers to distill knowledge from multiple teacher layers rather than following a fixed, one-to-one correspondence. Building on this concept, Wang et al. \cite{wang2022semckd} further refine the idea of non-matching layers by using a learned attention distribution to assign appropriate teacher layers to student layers, thereby enhancing cross-layer supervision and subsequently improving student model performance. In addition, Nath et al. \cite{nath2024rnas} introduce Robust Neural Architecture Search by Cross-Layer Knowledge Distillation (RNAS-CL), which searches for the best teacher layer to supervise each student layer, thus allowing for non-matching layer associations that enhance robustness in neural architectures. Furthermore, Zhao et al. \cite{zhao2023cross} develop Cross-Architecture Knowledge Distillation (CAKD), where non-matching layers are utilized to transfer knowledge from a Transformer-based teacher model to a CNN-based student model, involving the alignment of pixel-wise spatial information across different architectures and expanding the applicability of non-matching layers in cross-architecture scenarios.

\textbf{Layer Interaction:} This method leverages the hierarchical nature of neural networks, where different layers capture different levels of abstraction. By using high-level features from the teacher to guide the student's learning process, the student can gain a richer understanding of the data. Yao et al. \cite{yao2020knowledge} propose Dense Cross-layer Mutual-distillation (DCM), which involves layer interaction by integrating auxiliary classifiers and bidirectional knowledge distillation operations across different layers of the teacher and student models, thereby enhancing knowledge representation and performance. Building on this concept, Su et al. \cite{su2022deep} present Deep Cross-layer Collaborative Learning (DCCL), focusing on layer interaction through intermediate cross-layer supervision among peer student models, which integrates features from different layers to enhance representation and learning outcomes. Similarly, Zhu et al. \cite{zhu2022cross} introduce Cross-layer Fusion for Knowledge Distillation (CFKD), which aggregates features from both teacher and student models, allowing for rich layer interactions that further enhance the student model's learning process. In a related effort, Hu et al. \cite{hu2023layer} propose an online knowledge distillation method with layer-level feature fusion modules that connect sub-networks, thereby facilitating mutual learning through enhanced layer interaction among student networks. Expanding on the concept, Nguyen et al. \cite{nguyen2023cross} develop CLAFusion, a framework that employs cross-layer alignment for fusing neural networks with different numbers of layers, leveraging layer interaction to improve model accuracy and efficiency. Finally, Zhang et al. \cite{zhang2023cross} propose Patch Aware Knowledge Distillation (PAKD), which emphasizes cross-layer patch alignment and interaction within and across instances, guiding the student's learning of multi-level information and further reinforcing the importance of layer interaction in knowledge distillation.

\textbf{Hierarchical Guidance:} Cross-layer distillation can help in scenarios where the student model is significantly smaller or has a different architecture compared to the teacher. It allows the student to learn abstract representations earlier in its layers. Imagine a teacher model with layers: $T_1, T_2, T_3, \ldots, T_n$, and a student model with layers: $S_1, S_2, S_3, \ldots, S_m$. In cross-layer distillation, $T_n$ (the final layer of the teacher) might guide $S_3$ (a middle layer of the student), $T_3$ might guide $S_1$, and so on, depending on the distillation strategy and the specific architecture of the models. In this regard, Zou et al. \cite{zou2021coco} develop CoCo DistillNet, which utilizes cross-layer correlations to guide the student model in learning abstract representations from a teacher model in the context of pathological image segmentation, thereby enhancing the student model's performance in resource-constrained environments. Building on this concept, Zou et al. \cite{zou2022graph} propose Graph Flow Distillation, a method that transfers cross-layer variations from a large teacher network to a compact student network in medical image segmentation, enabling the student model to learn from both high-level and low-level abstractions of the teacher. In a similar vein, Zhai et al. \cite{zhai2024strengthening} introduce a method that uses the deepest feature maps from the teacher to guide the shallow layers of the student model, providing hierarchical guidance that effectively balances performance and efficiency. Furthermore, Guo et al. \cite{guo2023online} propose Alignahead++, an online knowledge distillation framework for GNNs that transfers structure and feature information across layers, facilitating hierarchical guidance and significantly improving performance on edge devices. Together, these studies underscore the importance of hierarchical guidance in enhancing the efficiency and effectiveness of knowledge distillation across various architectures.




\section{Tasks and Applications of KD in RS}
\subsection{Tasks}
\begin{figure*}[ht!]
\begin{center}
\includegraphics[width=0.65\textwidth]{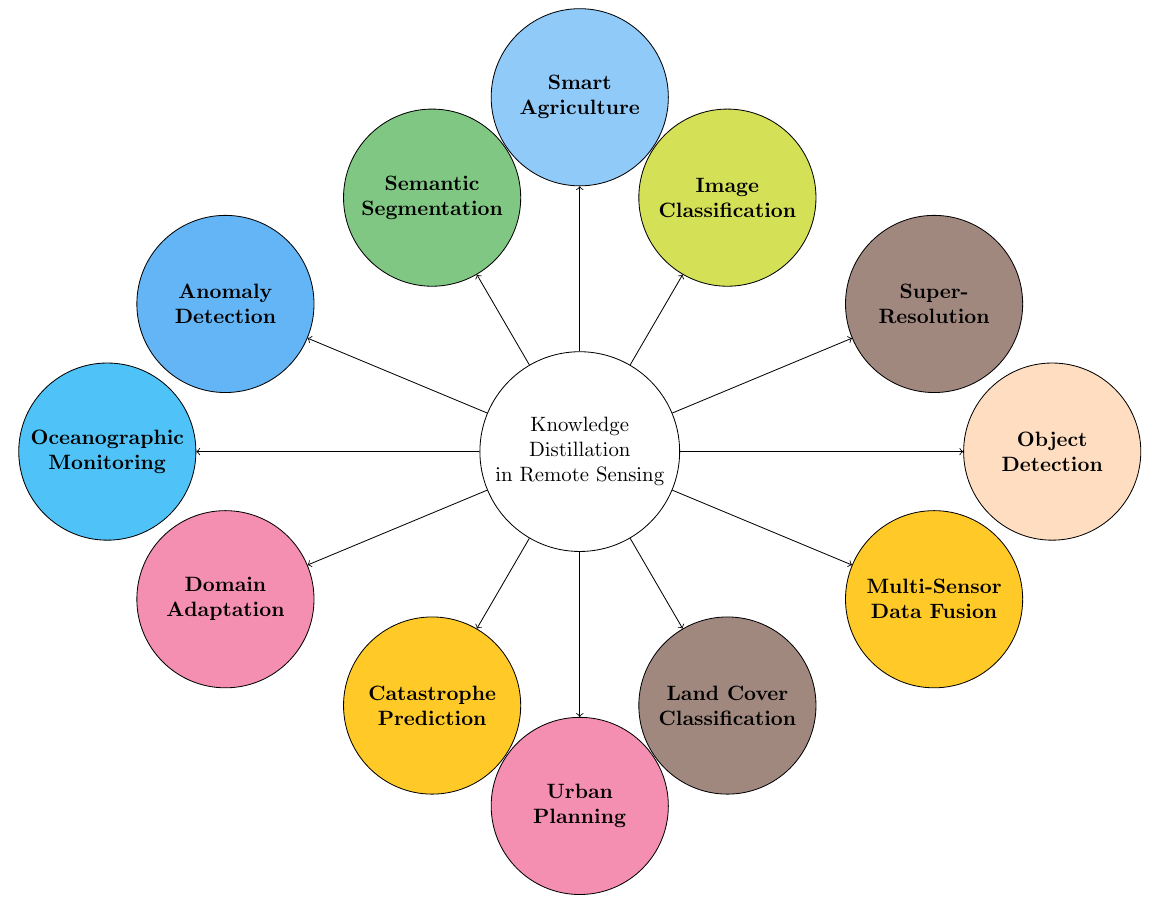}\\
\end{center}
\caption{Applications of KD in RS.}
\label{fig:applications}
\end{figure*}

As previously described, KD has emerged as a transformative approach in RS, enabling the development of more efficient models \textcolor{black}{that handle RS tasks with the same or even better performance across various applications. The main applications of KD is RS are depicted in Fig. \ref{fig:applications}.}


\subsubsection{Image/Scene Classification}
In the context of the classification of RS images/scenes, KD can be particularly beneficial. High-resolution satellite or hyperspectral images, which are rich in spatial and spectral information, can be computationally intensive to process \textcolor{black}{online} using large models. By employing KD, a large, powerful model (teacher) that has been trained on such images can pass on its learned representations and decision-making capabilities to a smaller, more efficient model (student). This allows the student model to achieve high classification accuracy while significantly reducing computational and storage requirements. Techniques such as spatial feature blurring can be incorporated to enhance the student's learning by making the training data more challenging, which helps in better generalization and improved classification performance. Various studies have been proposed in the literature to enhance RS image classification, focusing on KD, model efficiency, feature extraction, and handling noisy or incomplete data. \textcolor{black}{Table \ref{tab:comparison_rs_image} provides the main features of these works.} 
Building on this, Xu et al. \cite{xu2022hyperspectral} propose a hyperspectral image classification method based on class-incremental learning to learn new land-cover types without forgetting the old ones. This method uses a KD strategy to recall information of old classes and a channel attention mechanism to effectively utilize spatial-spectral information, demonstrating high accuracy on three hyperspectral image datasets. Similarly, Chi et al. \cite{chi2022novel} introduce a self-supervised learning method with KD for HSI classification, termed SSKD, which generates soft labels for unlabeled samples by considering spatial and spectral distances. This method significantly improves classification accuracy on three public HSI datasets. 
In addition, Xing et al. \cite{xing2022collaborative} address the challenge of using large deep neural networks on embedded devices by proposing a collaborative consistent KD (CKD) method. This method reduces the number of redundant parameters and improves the classification accuracy when tested on the SIRI-WHU and NWPU-RESISC45 datasets. Furthermore, Chen et al. \cite{chen2018training} focus on scene classification using a KD framework to improve the performance of smaller and shallower network models. Their method increases the overall accuracy when tested on AID, UCMerced, NWPU-RESISC, and EuroSAT datasets. Along similar lines, Song et al. \cite{song2024erkt} present ERKT-Net, an efficient and robust knowledge transfer network designed for lightweight yet accurate CNN classifiers, demonstrating superior accuracy and compactness on three RSI datasets. Likewise, Wu et al. \cite{wu2024takd} propose the TAKD method, which reduces background disturbance and improves the accuracy of student models for RS scene classification on three benchmark datasets. 
Moreover, Ienco et al. \cite{ienco2020generalized} propose a Generalized KD (GKD) framework to manage information misalignment between training and test data, demonstrating improved classification results using radar and optical satellite image time series data. Similarly, Zhang et al. \cite{zhang2020remote} address the challenge of noisy labels in RS image scene classification by proposing a noisy label distillation (NLD) method, which effectively distills knowledge from labels across a range of noise levels, achieving high accuracy on UC Merced Land-use, NWPU-RESISC45, and AID datasets.

In another approach, Zhao et al. \cite{zhao2022remote} propose a self-supervised KD network (SSKDNet) that uses feature maps as supervision signals and dynamically fuses feature maps to extract discriminating features, showing excellent performance on three datasets. Furthermore, Yang et al. \cite{yang2024two} introduce the TWA distillation method for RS object detection, reducing background information and addressing feature disparities, achieving superior performance on the LEVIR and SAR SSDD datasets. 
Additionally, Pande et al. \cite{pande2019adversarial} tackle the problem of missing modalities in RS image classification by proposing an adversarial training-driven hallucination architecture. This method shows that the student model can surpass the teacher model's performance on HSI datasets. 
In a similar vein, Yu et al. \cite{yu2024incremental} propose a two-stage training method for incremental learning that includes dual KD to prevent catastrophic forgetting, improving accuracy on CIFAR100 and RESISC45 datasets. Finally, Xie et al. \cite{xie2024decoupled} introduce an improved decoupled KD (DKD) strategy for HSI classification using a spatial feature blurring (SFB) module, achieving high overall accuracy on the Salinas dataset.

\begin{table*}[t]
\centering
\caption{Comparison of KD-based RS Image/Scene Classification Studies}
\scriptsize
\begin{tabular}{|m{0.4cm}|m{2cm}|m{3cm}|m{4cm}|m{3cm}|m{3cm}|}
\hline
\textbf{Ref.} & \textbf{Model(s) Used} & \textbf{Dataset/Data Type} & \textbf{Main Contribution} & \textbf{Best Performance Value Achieved} & \textbf{Limitation} \\ \hline

\cite{xu2022hyperspectral} & Class-incremental learning & PaviaU & KD with channel attention mechanism & 99.91\% OA & Bias towards new classes \\ \hline

\cite{xing2022collaborative} & Collaborative consistent KD & SIRI-WHU, NWPU-RESISC45 & Multi-branch fused redundant feature mapping & 0.943 accuracy (SIRI-WHU) & Parameter redundancy \\ \hline

\cite{chi2022novel} & Self-supervised learning with KD & Three HSI datasets & Adaptive generation of soft labels & 7.09\% improvement & Limited labeled samples \\ \hline

\cite{chen2018training} & KD framework & AID, UCMerced, NWPU-RESISC, EuroSAT & KD training method for small and shallow models & 5\% accuracy improvement (UCMerced) & Computationally expensive \\ \hline

\cite{song2024erkt} & ERKT-Net & Three RSI datasets & Efficient and robust KD network & 22.4\% OA (NWPU45) & Slight accuracy sacrifice \\ \hline

\cite{zhao2022remote} & SSKDNet & AID & Self-supervised KD network & 95.98\% accuracy & Complex training \\ \hline

\cite{pande2019adversarial} & Adversarial training & HSI datasets & Handling missing modalities with hallucination architecture & 98.17\% accuracy (Houston) & Modality dependency \\ \hline

\cite{zhang2021rs} & RS-SSKD & NWPU-RESISC45, RSD46-WHU & Few-shot classification with CAMs and KD & 86.26\% accuracy (NWPU-RESISC45) & Overfitting risk \\ \hline

\cite{zhang2020remote} & NLD & UC Merced, NWPU-RESISC45, AID & Handling noisy labels with end-to-end KD & 99.08\% accuracy (UC Merced) & Noisy data handling \\ \hline

\cite{ienco2020generalized} & GKD framework & Dordogne study site & Handling data misalignment with privileged information & 64.27\% F-Measure & Incomplete coverage \\ \hline

\cite{yang2024two} & TWA distillation & LEVIR, SAR SSDD & Reducing background noise and feature disparities & 95.4\% AP50 (SAR SSDD) & Background interference \\ \hline

\cite{yu2024incremental} & Incremental learning & CIFAR100, RESISC45 & Dual KD to prevent catastrophic forgetting & 6.9\% accuracy improvement & Stability-plasticity dilemma \\ \hline

\cite{xie2024decoupled} & DKD with SFB module & Four HSI datasets & Spatial feature blurring for better KD & 97.55\% OA (Salinas) & Fixed receptive fields \\ \hline

\end{tabular}
\label{tab:comparison_rs_image}
\end{table*}

\begin{figure*}[ht!]
\begin{center}
\includegraphics[width=0.9\textwidth]{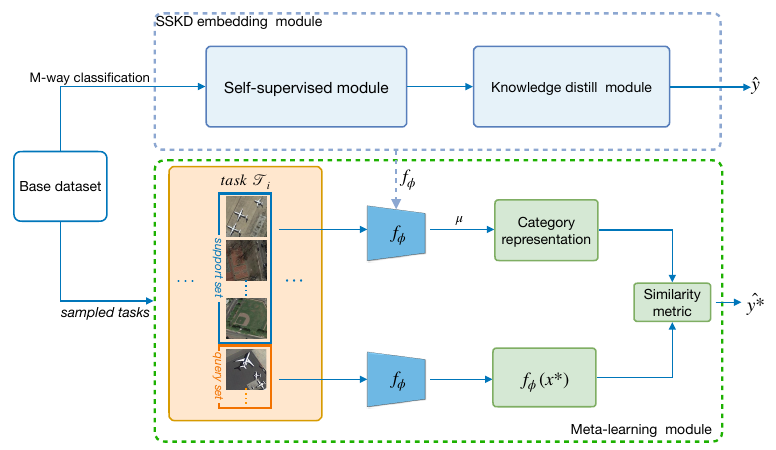}\\
\end{center}
\caption{The overall framework includes the SSKD module for embedding learning and the meta-learning module based on ProtoNets. The parameter $\gamma$ is used to adjust cosine similarity in the meta-learning process.}
\label{fig:zhang2021r}
\end{figure*}

Moving forward, Zhang et al. \cite{zhang2021rs} present RS-SSKD for few-shot RS scene classification, which uses Class Activation Maps (CAMs) and self-KD to generate powerful representations, achieving high accuracy on NWPU-RESISC45 and RSD46-WHU datasets. 
As the availability of airborne and satellite imagery increases, the challenge in RS (RS) scene classification has shifted from data scarcity to the lack of ground truth samples. Addressing these challenges, especially in unfamiliar environments with limited training data, few-shot classification offers a promising solution within meta-learning by extracting rich knowledge from minimal data. In \cite{zhang2021rs}, the authors introduce RS-SSKD, a method designed for few-shot RS scene classification that focuses on generating robust representations for downstream meta-learners. This approach features a two-branch network that uses three pairs of original-transformed images and incorporates Class Activation Maps (CAMs) to focus on the most relevant category-specific regions, ensuring the creation of discriminative embeddings. Additionally, a self-KD is applied to prevent overfitting and enhance performance \textcolor{black}{(see Fig. \ref{fig:zhang2021r})}. 

\subsubsection{Object Detection}
In RS, object detection is crucial for identifying specific features such as buildings, vehicles, and vegetation. KD helps in creating lightweight models that maintain high accuracy, making it feasible to run these models on devices with limited computational power.
Several studies focus on the use of KD for improving object detection in RS images, each introducing innovative strategies to address specific challenges. Algorithm \ref{algo1} outlines a process for KD in RS object detection. It starts by training a teacher model on a dataset, and then defines a student model with a simpler architecture. The teacher model generates soft targets, which are probability distributions over classes, using a softened softmax function. The student model is trained using a combined loss function that includes the cross-entropy loss and the Kullback-Leibler divergence between the teacher's and student's outputs. The process iterates over several epochs, optimizing the student model to mimic the teacher while also learning from the original labels. Finally, the trained student model is deployed.
\textcolor{black}{The main works on the use of KD in the object detection task in RS images and their main features are summarized in Table \ref{tab:kd_object_detection_apps}.}

\begin{algorithm}
\SetAlgoLined
\caption{KD for RS Object Detection}
\KwIn{Training data $D$, Teacher model $T$, Student model architecture $S$, Temperature $T_{emp}$, Loss weights $\alpha, \beta$, Number of epochs $N$}
\label{algo1}
\KwOut{Trained Student Model $S$}

\textbf{Step 1: Train the Teacher Model} \\
$T \gets \text{TrainTeacherModel}(D, T)$

\textbf{Step 2: Define the Student Model} \\
$S \gets \text{DefineStudentModel}(S)$

\textbf{Step 3: Compute the Soft Targets from the Teacher Model} \\
\For{each batch $(x, y) \in D$}{
    $z_T \gets T(x)$ \\
    $p_T \gets \text{Softmax}(z_T / T_{emp})$
}

\textbf{Step 4: Define the Loss Functions} \\
$\mathcal{L}_{CE} \gets \text{CrossEntropy}(S(x), y)$ \\
$\mathcal{L}_{KD} \gets \text{KLDiv}(\text{LogSoftmax}(S(x) / T_{emp}), p_T) \times T_{emp}^2$ \\
$\mathcal{L}_{total} \gets \alpha \mathcal{L}_{CE} + \beta \mathcal{L}_{KD}$

\textbf{Step 5: Train the Student Model} \\
\For{epoch = 1 to $N$}{
    \For{each batch $(x, y) \in D$}{
        $p_T \gets \text{ComputeSoftTargets}(T, x, T_{emp})$ \\
        $z_S \gets S(x)$ \\
        $\mathcal{L} \gets \text{ComputeLoss}(z_S, y, p_T, \alpha, \beta, T_{emp})$ \\
        Update $S$ by minimizing $\mathcal{L}$
    }
}

\textbf{Step 6: Deploy the Student Model} \\
$S \gets \text{Trained Student Model}$
\end{algorithm}

For instance, Yang et al. \cite{yang2022adaptive} propose an adaptive reinforcement supervision distillation (ARSD) framework to enhance lightweight object detectors. This method focuses on multiscale core features imitation and strict supervision regression distillation to improve performance, especially for small objects in complex backgrounds. Zhang et al. \cite{zhang2021learning} introduce a dynamic KD (DKD) framework, leveraging dynamic global distillation and instance selection distillation to enhance object detection in cluttered scenes. Another study by Zhang et al. \cite{zhang2024empowering} presents Orientation Distillation (OD) to address issues with boundary discontinuity and spatial feature ossification for detecting arbitrary-oriented objects in RS images, the authors further propose an adaptive composite feature generation (ACFG) strategy to improve feature mapping and handling of foreground and background loss in object detection \cite{zhang2024adaptive}.

Feng et al. \cite{feng2024enhancing} introduce an Instance-aware Distillation approach for Class-incremental Object Detection (IDCOD), which helps in preserving old class knowledge while learning new classes, thus mitigating catastrophic forgetting. Chen et al. \cite{chen2024discretization} propose Discretized Position KD (DPKD), which focuses on transferring high-quality bounding box position and pose information to improve object detection performance. Pang et al. \cite{pang2024exploring} present a pyramid KD (PKD) framework to handle the limitations of model compression, utilizing a hybrid online–offline smooth distillation strategy to enhance recognition accuracy while avoiding knowledge explosion and offset.

Du et al. \cite{du2024object} add a detection head specifically for small targets in the YOLOv5 model, proposing a network KD framework for improved small-scale target detection in RS images. Gao et al. \cite{gao2024feature} design a feature super-resolution fusion framework using cross-scale distillation to improve the detection accuracy of small objects by enhancing feature expression capability. Yang et al. \cite{yang2024weakly} propose a weakly supervised object detection method using self-attention distillation and instance-aware mining to handle varying scales and dense object proximity in RS images.

Other studies address different aspects of RS. Zhang et al. \cite{zhang2024object} combine detection and tracking in a joint framework (OKD-JDT), using KD to improve tracking efficiency and accuracy. Sun et al. \cite{sun2024lightweight} develop the Efficient Multidimensional Global Feature Adaptive Fusion Network (MGFAFNET) for UAV platforms, incorporating a dual-branch multidimensional aggregation backbone and a localized compensation dual-mask distillation strategy to balance detection speed and accuracy. Yang et al. \cite{yang2024dc} introduce DC-KD, a distillation scheme for object detection in satellite images, addressing data distribution differences between aerial and satellite images. Song et al. \cite{song2024efficient} present HMKD-Net, a hybrid-model KD approach combining CNNs and vision transformers to enhance classification performance in RS images. Zhang et al. \cite{zhang2024visual} propose a visual knowledge-oriented approach using pseudo labels to improve object detection in complex and dense RS images.

Lian et al. \cite{lian2024multitask} introduce a multitask learning framework combining image translation and saliency detection networks with KD to enhance feature expressiveness and reduce model complexity. Zeng et al. \cite{zeng2024novel} propose TDKD-Net, a tensor decomposition and KD-based network for UAV detection, focusing on small object detection and handling imbalanced issues. Wan et al. \cite{wan2024small} develop a coarse-to-fine detection method integrating density-aware scale adaptation and KD to improve small object detection in UAV images. Jia et al. \cite{jia2024mssd} suggest a multi-scale self-distillation approach to improve small target detection accuracy without using a teacher model. Lin et al. \cite{lin2024dtcnet} propose DTCNet, a distillation Transform-CNN network for super-resolution reconstruction in RS images, enhancing reconstruction quality while maintaining a smaller parameter count. Tang et al. \cite{tang2024text} introduce a text-guided tail-class generation network (TGN) to address long-tailed data distribution in RS datasets, improving tail-class accuracy by generating diverse and consistent tail-class images.


\begin{table*}[ht]
\caption{Comparison of Studies on KD-based Object Detection in RS Imagery.}
\centering
\scriptsize
\begin{tabular}{|p{0.4cm}|p{2cm}|p{2.4cm}|p{4.8cm}|p{3cm}|p{3cm}|}
\hline
\textbf{Ref.} & \textbf{Model(s) Used} & \textbf{Dataset/Data Type} & \textbf{Main Contribution} & \textbf{Best Performance} & \textbf{Limitation} \\ \hline
\cite{yang2022adaptive} & ARSD framework & DOTA, DIOR, NWPU VHR-10 & Adaptive reinforcement supervision distillation for lightweight object detection & Outperforms SOTA methods & High complexity due to adaptive modules \\ \hline
\cite{zhang2021learning} & DKD framework & DOTA, NWPU VHR-10 & Dynamic KD for multi-scale feature imitation & Suitable for various detectors & Potential overfitting to specific datasets \\ \hline
\cite{zhang2024empowering} & Orientation Distillation (OD) & Multiple datasets & Anti-ambiguous location prediction and feature calibration & Improved performance on non-axially arranged objects & Limited accuracy in complex scenes \\ \hline
\cite{zhang2024adaptive} & ACFG strategy & DIOR, DOTA & Adaptive composite feature generation for KD & Better performance than SOTA KD algorithms & Complexity in composite mask generation \\ \hline
\cite{feng2024enhancing} & IDCOD & DOTA, DIOR, RTDOD, PASCAL VOC & Instance-aware distillation for class-incremental detection & mAP@0.5 of 74.0\% on DIOR & Challenge in handling new classes post-deployment \\ \hline
\cite{chen2024discretization} & DPKD & DOTA, HRSID & Discretized position KD for object detection & mAP of 79.82\% on DOTA & Overlooks certain localization knowledge \\ \hline
\cite{pang2024exploring} & PKD framework & Aircraft, FGSC-23 & Pyramid KD to avoid knowledge explosion and offset & Effective with ResNet and VGG networks & Complexity in finding optimal configuration \\ \hline
\cite{du2024object} & Enhanced YOLOv5 & NWPU VHR-10 & KD framework for small-scale target detection & Detection accuracy of 43.9\% & High computational cost \\ \hline
\cite{gao2024feature} & SSRFPN with CSD & NWPU VHR-10, DIOR & Feature super-resolution fusion for small-object detection & AP0.5 of 95.0\% on NWPU VHR-10 & Difficulty in feature extraction for very small objects \\ \hline
\cite{yang2024weakly} & WSOD with SAD and IAM & NWPU VHR-10, DIOR & Weakly supervised learning for object detection & Accurate bounding boxes & Struggles with varying scales and dense objects \\ \hline
\cite{zhang2024object} & OKD-JDT & JiLin-1 & Joint detection and tracking framework & State-of-the-art performance & Limited to certain types of satellite videos \\ \hline
\cite{sun2024lightweight} & MGFAFNET & SyluDrone & Efficient detection method for UAV platforms & AP of 52.7\%, AP50 of 93.6\% on SyluDrone & Balancing detection speed and accuracy \\ \hline
\cite{yang2024dc} & DC-KD & xView & Distillation scheme for object detection in satellite images & 3.88\% mAP50 improvement on xView & Data distribution differences \\ \hline
\cite{song2024efficient} & HMKD-Net & Multiple datasets & Hybrid-model KD with CNN-ViT ensemble & Max accuracy improvement of 22.8\% & Handling variances during KD \\ \hline
\cite{zhang2024visual} & Visual knowledge-oriented WSOD & NWPU VHR-10, DIOR & Leveraging visual cues as pseudo labels & mAP of 84.25\% on NWPU VHR-10 & Handling noise in object proposals \\ \hline
\cite{lian2024multitask} & WSA-GAN, BGNet & Various RS datasets & Multitask learning for image translation and saliency detection & Outperforms other approaches & Complexity in multimodal context learning \\ \hline
\cite{zeng2024novel} & TDKD-Net & Various RS datasets & Tensor decomposition and KD for UAV detection & High generalization and robustness & Handling imbalanced issues \\ \hline
\cite{wan2024small} & Coarse-to-fine network & VisDrone, UAVDT & Density-aware scale adaptation for small object detection & Superior detection in UAV images & Issues with scale variation \\ \hline
\cite{jia2024mssd} & Self-distillation YOLO & KITTI & Multi-scale self-distillation for object detection & 2.8\% accuracy improvement & Inefficiencies in knowledge transfer \\ \hline
\cite{lin2024dtcnet} & DTCNet & AID & Distillation Transform-CNN for super-resolution & PSNR of 28.73 dB, SSIM of 0.7904 & High model complexity \\ \hline
\cite{tang2024text} & TGN with KMDN and CDTG & DIOR, FGSC-23, DOTA & Text-guided tail-class generation for long-tailed distribution & Superior performance on tail classes & Data distribution imbalance \\ \hline
\end{tabular}
\label{tab:kd_object_detection_apps}
\end{table*}

\begin{figure*}[ht!]
\begin{center}
\includegraphics[width=1\textwidth]{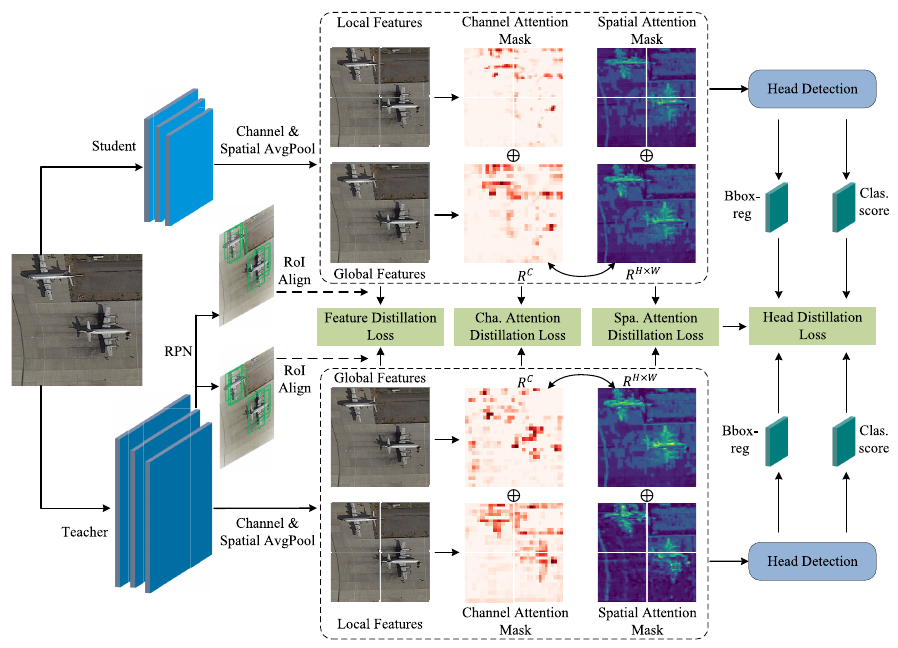}
\end{center}
\caption{The KD architecture enhances AFD through three key advancements \cite{shamsolmoali2023efficient}. Firstly, the method extracts both local and global information from the teacher network. For local distillation, a multi-instance attention mechanism is introduced to effectively distinguish foreground elements from the background. Secondly, the approach reconstructs the relationships between different pixels, facilitating a more comprehensive transfer of knowledge from the teacher to the student detector through both attention-focused local and global distillation strategies.}
\label{fig-shamsolmoali2023efficient}
\end{figure*}

On the other hand, traditional KD-based object detection methods have limitations, such as ignoring crucial background information and focusing solely on global context. To overcome these issues, Attention-based Feature Distillation (AFD) is proposed in \cite{shamsolmoali2023efficient}, which distills both local and global information. AFD enhances local distillation with a multi-instance attention mechanism and reconstructs pixel relationships, resulting in state-of-the-art performance in object detection while remaining efficient. 
Fig. \ref{fig-shamsolmoali2023efficient} illustrates the architecture of the proposed Attention-based Feature Distillation (AFD) method. This framework improves upon traditional KD-based object detection by incorporating both local and global information from the teacher network. The multi-instance attention mechanism within AFD allows the model to distinguish between background and foreground elements effectively. Additionally, the method reconstructs pixel relationships, ensuring that both local details and broader context are accurately transferred from the teacher to the student detector, resulting in enhanced detection performance.

\subsubsection{Semantic Segmentation}
KD is beneficial for semantic segmentation in RS applications, which involves classifying each pixel in an image into predefined categories. The teacher model is first trained on the segmentation task using high-resolution RS data \cite{sun2024cs}. Due to its complexity and larger capacity, it can learn intricate patterns and detailed features from the data. Once trained, the teacher model's predictions, along with its internal representations, are used to guide the training of the student model. The student model, being smaller and more efficient, aims to mimic the performance of the teacher model while maintaining lower computational costs and faster inference times \cite{yuan2024fakd}. Moreover, KD is particularly advantageous because it allows for the deployment of effective semantic segmentation models on edge devices or in scenarios with limited computational resources \cite{naushad2021deep}. By leveraging the distilled knowledge from the teacher model, the student model can achieve high segmentation accuracy despite its reduced size. This is crucial for applications such as real-time environmental monitoring, disaster response, and agricultural analysis, where timely and accurate segmentation of satellite or aerial imagery is needed. The distillation process also helps the student model generalize better to new and unseen data, enhancing its robustness and reliability in diverse RS tasks \cite{wang2023ssd}.

The studies on semantic segmentation in RS show significant advancements but also face several limitations. For instance, Gao et al. \cite{gao2024enrich} introduced the FoMA framework, which significantly improves segmentation performance by leveraging foundation models, but it struggles with data scarcity in novel classes and balancing segmentation performance across classes. Similarly, Zhou et al. \cite{zhou2023graph} proposed a lightweight student network (GAGNet-S*) with KD that achieves excellent segmentation performance but faces challenges related to scalability and complexity in deployment on resource-limited equipment. Dong et al. \cite{dong2023distilling} addressed the limitations of CNNs and transformers by proposing the DSCT framework, which enhances segmentation performance through cross-model KD. However, this approach requires high computational complexity and massive data resources.

Studies focusing on KD methods, such as the MGSAD by Zhang et al. \cite{zhang2022multi}, and MTKD by Li et al. \cite{li2022weather} with MTKD, contribute innovative techniques but encounter challenges such as the need for extensive computation and handling of domain shifts. Liu et al. \cite{liu2022unsupervised} proposed a three-stage UDA method that shows better performance but relies heavily on large-scale annotated data and struggles with domain shift handling. Similarly, Shi et al. \cite{shi2022dsanet} introduced DSANet, which effectively handles spatial and semantic feature enhancement but reduces the model's characterization ability for these features.

Incremental learning and domain adaptation are other areas where significant contributions have been made but also face limitations. Rong et al. \cite{rong2022historical} proposed a generalized framework for CSS but struggled with the challenge of old classes collapsing into the background. Rui et al. \cite{rui2023dilrs} and Le et al. \cite{le2024leveraging} focused on incremental learning methods but faced high computational costs and the complexity of adapting to incremental domains and partial multi-task learning, respectively. Shan et al. \cite{shan2021class, shan2022class} developed class-incremental segmentation methods that address catastrophic forgetting but require balancing old and new class learning and managing feature generation complexity. Li \cite{li2021learning} proposed DSSN with weakly-supervised constraints to handle cross-domain segmentation, but the method heavily depends on labeled data and struggles with geographic variation.

Lastly, Guo et al.  \cite{guo2024contrastive} and Cao et al. \cite{cao2021c3net} proposed methods to balance effectiveness and compactness in segmentation models, but they face high computational demands and challenges in handling noise and redundant features. Zhou et al. \cite{zhou2023gsgnet} introduced GSGNet with high inference speed but had to balance this with contextual reasoning capabilities. Bai et al. \cite{bai2022domain} and Wang et al. \cite{wang2022avoiding} focused on domain adaptation, but they faced difficulties in aligning high-dimensional image representations and managing intermediate domain learning. Michieli et al. \cite{michieli2021knowledge} addressed incremental learning with various KD techniques but struggled with catastrophic forgetting and internal feature representation complexity. Lastly, Pena et al. \cite{pena2024deepaqua} introduced DeepAqua for water detection, which improves segmentation accuracy but lacks specific details on datasets and segmentation scenarios.
\textcolor{black}{Table \ref{tab:comparison_segmentation} provides a summary of the works that use KD for semantic segmentation of RS images.}

\begin{table*}[h!]
\centering
\caption{Comparison of Various Studies on Semantic Segmentation in RS}
\scriptsize
\begin{tabular}{|p{0.4cm}|p{2cm}|p{2.4cm}|p{4.8cm}|p{3cm}|p{3cm}|}
\hline
\textbf{Ref.} & \textbf{Model(s) Used} & \textbf{Dataset/Data Type} & \textbf{Main Contribution} & \textbf{Best Performance Value Achieved} & \textbf{Limitation(s)} \\
\hline
\cite{gao2024enrich} & FoMA Framework & OpenEarthMap & Introduces GFSS with three strategies: SLE, DGK, and VFE for improved segmentation & Improvement of 28.94\% in segmentation performance, with 31.79\% for novel classes and 24.64\% for base classes & Data scarcity in novel classes and complex balancing in segmentation performance \\
\hline
\cite{zhou2023graph} & GAGNet-S* (with KD) & Potsdam, Vaihingen & Proposes a lightweight student network framework with KD & Achieved excellent segmentation performance on Potsdam and Vaihingen datasets & Scalability and complexity in deployment on resource-limited equipment \\
\hline
\cite{dong2023distilling} & DSCT Framework & ISPRS Potsdam, Vaihingen, GID, LoveDA & Cross-model KD using CNNs and transformers & Outperforms state-of-the-art KD methods on four datasets & High computational complexity and massive data resource requirements \\
\hline
\cite{zhang2022multi} & MGSAD & Not specified & Proposes a multi-granularity semantic alignment distillation method for semantic segmentation & Not specified & Details on datasets and specific performance metrics are not provided \\
\hline
\cite{li2022weather} & MTKD & Not specified & Multi-task KD for weather-degraded image segmentation & Achieves 0.038 s in semantic segmentation for a 2048 × 1024 image & Specific performance values not provided, computation-intensive \\
\hline
\cite{liu2022unsupervised} & Covariance-based Channel Attention Module & ISPRS 2-D Semantic Labeling, Urban Drone Dataset (UDD) & Proposes three-stage UDA method with KD for RS images & Shows better performance compared with state-of-the-art methods & Domain shift handling and reliance on large-scale annotated data \\
\hline
\cite{shi2022dsanet} & DSANet & ISPRS Potsdam, Vaihingen & Effective deep supervision-based attention network for RSIs & 79.19\% mIoU on Potsdam, 72.26\% mIoU on Vaihingen with 470.07 FPS on 512 × 512 images & Reduces model characterization ability for spatial and semantic features \\
\hline
\cite{nair2024let} & Cross-modal KD & Not specified & Uses optical images to train a student model for SAR images through cross-modal KD & Increase of 5-20\% IoU score compared to training from scratch & Small training datasets and complexity in cross-modal learning \\
\hline
\cite{rong2022historical} & Generalized Framework for CSS & iSAID, GCSS & Proposes historical information-guided modules for CSS in RS images & Outperforms state-of-the-art methods in most incremental settings & Challenge of old classes collapsing into the background \\
\hline
\cite{rui2023dilrs} & Domain-Incremental Learning & LoveDA-rural & Proposes domain-incremental learning for multi-source RS data & Achieves mIoU of 0.6233 on LoveDA-rural at step 5 & High computational cost and complexity in incremental domain learning \\
\hline
\cite{le2024leveraging} & Partial Multi-Task Learning with KD & ISPRS 2D Semantic Labeling Contest & Enhances partial multi-task learning performance using KD & mIoU of 68.97\% on Vaihingen dataset & Lack of all-task annotations and reliance on soft labels \\
\hline
\cite{shan2021class} & DFD and LM Modules & Aerial images dataset & Proposes class-incremental segmentation method without old data storage & 6.2\% and 15\% mIoU gains from DFD and LM modules respectively & Catastrophic forgetting and balancing old and new class learning \\
\hline
\cite{shan2022class} & PFG and TKD Modules & Not specified & Effective class-incremental segmentation framework without storing old data & More than 4.5\% gains compared with state-of-the-art methods & Limited detail on dataset performance and complexity in feature generation \\
\hline
\cite{li2021learning} & DSSN with Weakly-Supervised Constraints & Not specified & Proposes DSSN for cross-domain RS image segmentation & Mean F1Score: 60.76\%, Mean IoU: 44.53\% & High dependency on labeled data and difficulty in geographic variation handling \\
\hline
\cite{guo2024contrastive} & CLNet-T and CLNet-S (with KD) & MFNet, PST900 & Proposes a balance between effectiveness and compactness using KD & MFNet: mAcc 76.6\%, mIoU 58.2\%; PST900: mAcc 95.59\%, mIoU 80.77\% & High computational demands and complexity in terminal device deployment \\
\hline
\cite{cao2021c3net} & C3Net with Multi-Level KD & ISPRS Vaihingen & Proposes efficient C3Net for multi-modal data semantic segmentation & Overall Accuracy: 91.3\%, High mean F1 score for car class & Noise and redundant feature handling and high running time \\
\hline
\cite{zhou2023gsgnet} & GSGNet with KD & Vaihingen, Potsdam & Proposes GSGNet for ORSI scenario analysis with high inference speed & Outperforms most advanced methods with 19.61 M parameters & Balancing high inference speed and contextual reasoning capability \\
\hline
\cite{bai2022domain} & Contrastive and Adversarial Learning & Not specified & Proposes a model for domain adaptation in representation space and spatial layout & Not specified & Specific performance values and dataset details not provided \\
\hline
\cite{wang2022avoiding} & TDARS & Three domain adaptation datasets & Proposes transitive domain adaptation for RS images & Effectively handles domain shift problem compared to other methods & High complexity in intermediate domain learning and transfer \\
\hline
\cite{michieli2021knowledge} & Various KD Techniques & Pascal VOC2012, MSRC-v2 & Proposes incremental learning for semantic segmentation with KD & Highest Accuracy: 97.5\% (Abisoye et al. 2024), Lowest Error: 0.032 MAE (De 2024) & Catastrophic forgetting and complexity in internal feature representation handling \\
\hline
\cite{pena2024deepaqua} & DeepAqua & Not specified & Proposes an unsupervised method for water detection in RS & Improves accuracy by 3\%, IoU by 11\%, F1-score by 6\% & Specific details on datasets and segmentation scenarios not provided \\
\hline
\end{tabular}
\label{tab:comparison_segmentation}
\end{table*}

\begin{figure*}[ht!]
\begin{center}
\includegraphics[width=1\textwidth]{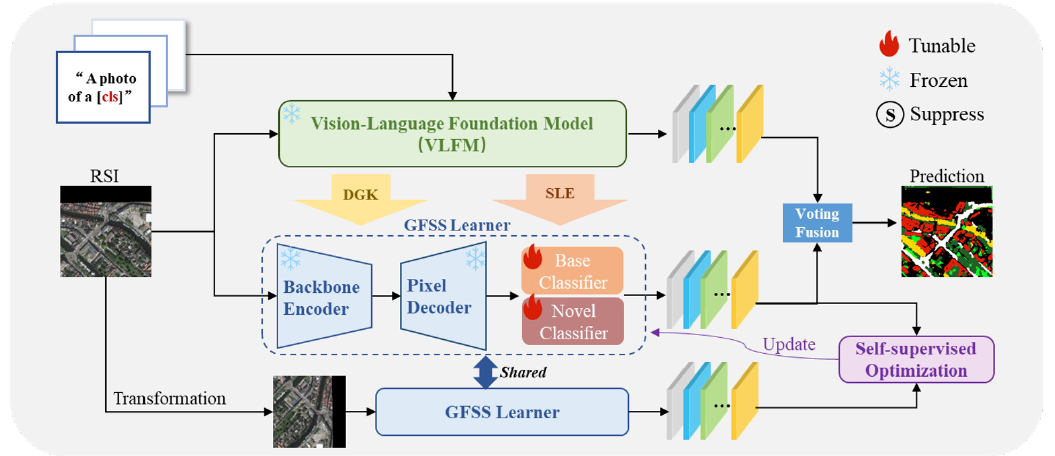}\\
(a) The FoMA GFSS framework\\
\includegraphics[width=1\textwidth]{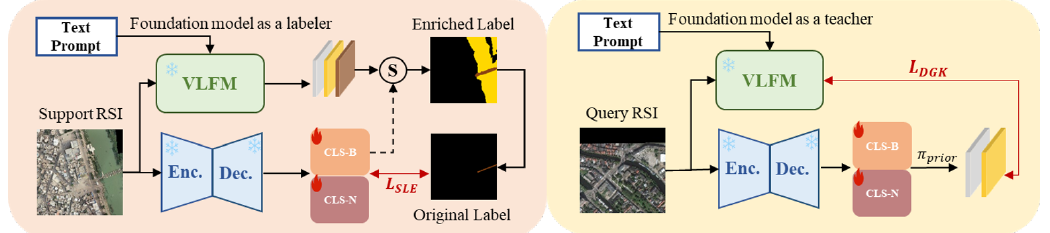}\\
(b) SLE~~~~~~~~~~~~~~~~~~~~~~~~~~~~~~~~~~~~~~~~~~~~~~~~~~~~~~~~~~~~~~~~~~~~~~~~~(c) DGK\\
\end{center}
\caption{The FoMA's architecture proposed in \cite{gao2024enrich}, incorporates the general knowledge from a vision-language foundation model, initially trained on natural images, into the GFSS task for RS images. This is achieved through two key modules: SLE, which integrates the foundation model's predictions as pseudo-labels into the GFSS learner's training on support images, and DGK, which transfers the model's superior performance on novel classes from query images into the learner. Additionally, a voting fusion strategy effectively combines results from both the foundation model and the GFSS learner for enhanced accuracy..}
\label{fig-gao2024enrich}
\end{figure*}

Besides, \cite{gao2024enrich} produces a Foundation Model Assisted (FoMA) for Generalized Few-Shot Semantic Segmentation (GFSS) in RS images, aimed at improving segmentation performance under data scarcity conditions. FoMA leverages foundation models through three strategies: Support Label Enrichment (SLE) to enhance support labels, Distillation of General Knowledge (DGK) to transfer generalizable knowledge, and Voting Fusion of Experts (VFE) to combine zero-shot and few-shot predictions. The method demonstrates state-of-the-art performance on the OpenEarthMap few-shot challenge dataset. Fig. \ref{fig-gao2024enrich} illustrates the architecture of the FoMA framework, which effectively integrates a vision-language foundation model's general knowledge into the GFSS task for RS images.

\subsection{Specific Applications}

\subsubsection{Land Cover Classification}
\begin{figure*}[ht!]
\centering
\includegraphics[width=1\textwidth]{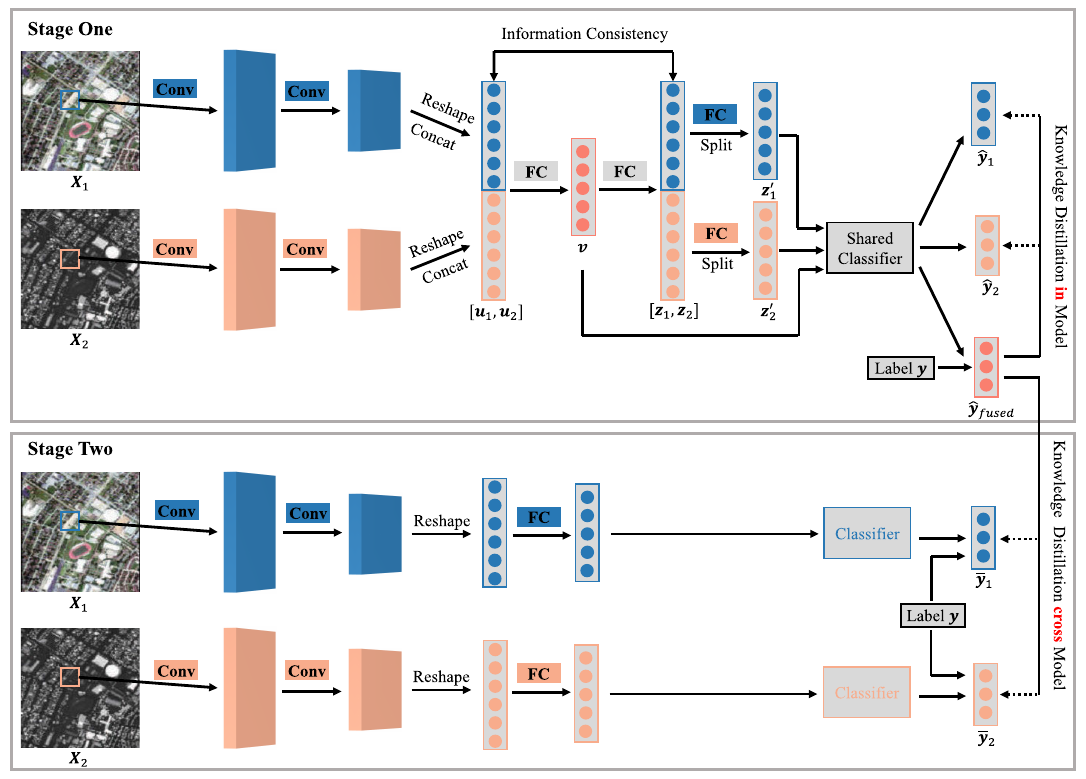}
\caption{Illustration of the framework proposed in \cite{xu2023robust}. The "Conv" block comprises a 3 × 3 convolutional layer, followed by batch normalization, a 2 × 2 max-pooling layer, and a ReLU activation function. The "FC" block includes a fully connected layer, batch normalization, and a ReLU activation function. Both the "Shared Classifier" and "Classifier" share the same structure, composed of "FC" blocks and a softmax layer for final classification.}
\label{fig:xu2023robust}
\end{figure*}

KD improves the classification of land cover types by refining the feature extraction capabilities of student models. This leads to better segmentation and classification of different land cover types, essential for environmental monitoring and urban planning. Several studies have proposed innovative methods to improve land cover classification and other RS tasks using KD and multimodal data fusion. For example, Xu et al. \cite{xu2023robust} developed a two-branch patch-based CNN with an encoder-decoder (ED) module to fuse multimodal RS (RS) data. They introduced a KD in model (DIM) module for better multimodal data fusion and a cross-model (DCM) module to enhance single-modal classification using multimodal knowledge. Their approach demonstrated superior performance on hyperspectral (HS) and light detection and ranging (LiDAR) data as well as HS and synthetic aperture radar (SAR) data. 
\textcolor{black}{Fig. \ref{fig:xu2023robust} depicts the approach proposed in \cite{xu2023robust}.}
Wang et al. \cite{wang2023cross} proposed the cross-modal graph knowledge representation and distillation learning (CGKR-DL) framework, which combines CNN and graph convolutional network (GCN) to enhance land cover classification. Their method addresses the limitations of traditional CNN-based cross-modal distillation methods and significantly improves performance on various multimodal RS datasets.

The Generalized KD (GKD) framework has been introduced by Ienco et al. \cite{ienco2020generalized} to handle data misalignment between training and test phases. Their method, applied to radar and optical satellite image time series data, improved land use land cover mapping, especially for agricultural classes.
A multimodal online KD (MMOKD) framework that supports both multimodal and cross-modal learning, showing superior performance in both scenarios has been proposed by Liu et al. \cite{liu2024multimodal} for land use/cover classification using optical and SAR images.
Finally, Li et al. \cite{li2021dynamic} introduced the dynamic-hierarchical attention distillation network (DH-ADNet) with multimodal synergetic instance selection (MSIS) for land cover classification using missing data modalities. Their method emphasizes selective instance enhancement and hierarchical attention distillation, achieving state-of-the-art results.

Several other studies focused on specific RS challenges. For instance, Lu et al. \cite{lu2024weakly} developed a weakly supervised change detection technique via KD and Multiscale Sigmoid Inference (KD-MSI), significantly improving change detection performance on multiple datasets. Similarly, Zhang et al. \cite{zhang2023deep} proposed a transfer learning framework using teacher-student structure for better generalizability and performance in land cover classification.
Kanagavelu et al. \cite{kanagavelu2023fedukd} and Gbodjo et al. \cite{gbodjo2021multisensor} explored federated learning and multisensor data integration, respectively, to enhance land cover mapping and monitoring. The former work federated UNet model integrated KD to reduce communication costs while maintaining high accuracy whereas the later developed a self-distillation strategy within a CNN framework to combine multitemporal SAR and optical data for improved land cover classification.
\textcolor{black}{The works that use KD for the classification of land cover and their main characteristics are summarized in Table \ref{tab:comparison_land_cover}.}

\begin{table*}[t]
\caption{Comparison of Studies on KD-based Land Cover Classification}
\centering
\scriptsize
\begin{tabular}{|p{0.4cm}|p{2cm}|p{2.4cm}|p{4.8cm}|p{3cm}|p{3cm}|}
\hline
\textbf{Ref.} & \textbf{Model(s) Used} & \textbf{Dataset/Data Type} & \textbf{Main Contribution} & \textbf{Best Performance Value Achieved} & \textbf{Limitation} \\ \hline
\cite{xu2023robust} & Two-branch patch-based CNN with ED and DIM modules & Hyperspectral (HS) and LiDAR data (Houston2013) & Developed a KD in model (DIM) and cross-model (DCM) module for better LC classification & Improved LC classification performance on two multimodal RS datasets & The study mainly focuses on LC classification; does not cover other RS applications \\ \hline
\cite{wang2023cross} & CGKR-DL framework with CNN and GCN & HS-LiDAR, HS-SAR, HS-SAR-DSM datasets & Proposed cross-modal graph knowledge representation and distillation learning & Significant improvement in land cover classification accuracy & Focuses on classification; not on other types of RS tasks \\ \hline
\cite{ienco2020generalized} & Generalized KD (GKD) framework & Radar (Sentinel-1) and optical (Sentinel-2) SITS & Managed information misalignment between training and test data & Accuracy: 65.01\%, F-Measure: 64.27\%, Kappa: 0.5775 & Limited to cases where radar data is always available \\ \hline
\cite{liu2024multimodal} & MMOKD framework & Optical and SAR images & Developed multimodal online KD framework for land use/cover classification & Outperformed other networks in both full- and missing-modality scenarios & Large semantic gap between modalities poses a challenge \\ \hline
\cite{kanagavelu2023fedukd} & Federated UNet model with KD & Satellite and street view images & Improved efficiency and privacy of real-time climate tracking & Accuracy above 95\% & Focus on semantic segmentation, not other RS tasks \\ \hline
\cite{li2021dynamic} & DH-ADNet with MSIS & Coregistered optical and SAR datasets & Introduced dynamic-hierarchical attention distillation for land cover classification & State-of-the-art results in the privileged information scenario & Limited to privileged information scenarios \\ \hline
\cite{lu2024weakly} & KD-MSI with CAMs & WHU-CD, DSIFN-CD, LEVIR-CD datasets & Weakly supervised change detection using KD & F1-score: 0.854 on WHU-CD & Focuses on change detection; not applicable to other RS tasks \\ \hline
\cite{zhang2023deep} & Transfer learning framework with CMD and high-temperature softmax & Various RS datasets & Improved land cover classification using teacher-student structure & Average increase in mIoU: 9.9\%, 2.1\%, 4.3\% & Requires large datasets for teacher model training \\ \hline
\cite{li2022dense} & DAGDNet with IG-FGM and MS-ADL & Coregistered optical and SAR datasets & Efficient dense adaptive grouping distillation network for MLCC & Superior performances on representative datasets & Limited to scenarios with privileged modality \\ \hline
\cite{gbodjo2021multisensor} & Patch-based multibranch CNN & Multitemporal SAR/optical data & Integrated multisensor RS data using self-distillation strategy & Accuracy: 94\% (Reunion island), 88\% (Dordogne) & Requires sparsely annotated ground-truth data \\ \hline
\cite{kumar2021improved} & Hallucination network with KD & PAN-MS image pairs, hyperspectral dataset & Provided robust solution for missing modalities using hallucination module & Overall accuracy: 97.01\% & Focused on scene recognition and image classification \\ \hline
\cite{xu2024cloudseg} & CloudSeg framework with multi-task learning & M3M-CR, WHU-OPT-SAR datasets & Addressed semantic segmentation under cloud cover using KD & mIoU improvement: 3.16\% (M3M-CR), 5.56\% (WHU-OPT-SAR) & Focuses on cloudy conditions; not applicable to cloud-free scenarios \\ \hline
\cite{julka2023knowledge} & Segment Anything (SAM) model & Planetary images & Rapid annotation for geological mapping using KD & Comparable to state-of-the-art on mapping planetary skylights & Limited to geological mapping tasks \\ \hline
\cite{bazzi2020distilling} & Distill and refine strategy with CNN & Sentinel-1 data & Addressed spatial transfer challenge for mapping irrigated areas & Best performance in spatial transferability & Focused on spatial transfer; not on other RS tasks \\ \hline
\cite{quan2021lightweight} & Lightweight model with KD & UC Merced Land Use dataset & High accuracy and efficiency for RS image retrieval & mAP: 0.9680 with 3.8M parameters & Limited to image retrieval tasks \\ \hline
\cite{broni2024unsupervised} & MRF-NAS with self-training UDA & OpenEarthMap, FLAIR \#1 datasets & Lightweight neural networks for UDA in RS & mIoU: 59.38\% (OpenEarthMap), 51.19\% (FLAIR \#1) & Focus on UDA; not on other RS tasks \\ \hline
\cite{garg2023cross} & Cross-modal distillation framework & Sen1Floods11 dataset & Improved flood detection with cross-modal distillation & IoU improvement: 6.53\% on test split & Limited to flood detection; not other RS tasks \\ \hline
\cite{yan2022pansharpening} & GCPNet with GCN and ASPM & Various satellite datasets & Enhanced pansharpening using GCN and KD & Outperformed state-of-the-art visually and quantitatively & Limited to pansharpening tasks \\ \hline
\cite{yan2023domain} & Domain knowledge-guided self-supervised learning & Onera Satellite Change Detection dataset & Improved unsupervised change detection using domain knowledge & Kap: 53.34\%, F1: 55.69\% & Focused on change detection; not other RS tasks \\ \hline
\cite{matin2023discern} & VGG13 (teacher), ResNet8 (student) & SMAP satellite data & Improved soil moisture prediction using KD & High prediction accuracy with efficient student model & Focused on soil moisture prediction; not other RS tasks \\ \hline
\cite{ren2023incremental} & LSAW with adaptive weights & CCF, Potsdam, Vaihingen datasets & Addressed catastrophic forgetting in incremental learning & Best results on three datasets & Focus on incremental learning; not other RS tasks \\ \hline

\end{tabular}
\label{tab:comparison_land_cover}
\end{table*}

Besides, significant research has been devoted to enhancing land cover classification using multimodal RS data, which significantly outperforms single-modal methods due to its richer information content. To advance this field, a two-branch, patch-based CNN with an encoder-decoder (ED) module for effective multimodal data fusion is proposed in \cite{xu2023robust}. Typically, a KD in model (DIM) module to guide per-modality encoder learning is introduced, ensuring more efficient fusion. Additionally, we explored guiding single-modal learning with multimodal information through the KD cross-model (DCM) module. This approach treats the multimodal method as a teacher, transferring its knowledge to single-modal methods. Extensive experiments on the Houston2013 and Berlin datasets, combining hyperspectral (HS) with LiDAR and synthetic aperture radar (SAR) data, respectively, demonstrated the superiority of our multimodal fusion strategy over state-of-the-art methods. The DCM module also significantly enhances LC classification performance for single-modal methods.

\subsubsection{Precision Agriculture}
KD is crucial in smart agriculture as it allows for the development of lightweight models that maintain high accuracy while being deployable on resource-constrained edge devices, such as drones or sensors. This is particularly important for precision agriculture tasks like early weed detection and crop monitoring, where efficient and accurate models are needed for real-time decision-making. By transferring knowledge from larger, more complex models to smaller ones, KD helps optimize these tasks, enhancing agricultural productivity and sustainability. Numerous studies have investigated the application of KD in precision agriculture. These efforts focus on enhancing model efficiency and accuracy in tasks such as crop monitoring, weed detection, and resource management. 
For instance, Liangde et al. \cite{liangde2021agricultural} develop a model distillation approach to enhance agricultural named entity recognition, leveraging a BERT-based model enhanced by BiLSTM and CRF for precise entity detection from a constructed agriculture knowledge graph.
Ghofrani and Mahdian Toroghi \cite{ghofrani2022knowledge} focus on plant disease detection, using a KD approach to enable smaller CNN architectures, like MobileNet, to achieve near high-end model accuracy on the Plantvillage dataset.
On the same line, Hu et al. \cite{hu2023lightweight} and Dong et al. \cite{dong2024apple} address crop disease detection. The former approach optimizes YOLOv5s for maize disease detection, while the latter uses ECA-KDNet for efficient apple disease diagnosis on mobile devices.
Finally, Huang et al. \cite{huang2023knowledge} develop multistage KD to create lightweight models for diagnosing multiple crop diseases effectively.

In the task of image segmentation, Angarano et al. \cite{angarano2024domain} introduce a method for robust crop segmentation using KD, aimed at improving the generalization across different environmental conditions for robotic field management.
Similarly, Li et al. \cite{li2023knowledge} employ KD for efficient panoptic segmentation, creating lightweight networks capable of detailed scene understanding at high speeds, whereas Jung et al.
\cite{jung2022plant} improve plant leaf segmentation using KD to maintain high-quality instance segmentation.
The work of Pag{\'e}-Fortin \cite{page2023class} investigates class-incremental learning methods to address the challenge of learning new plant species and diseases incrementally, focusing on mitigating catastrophic forgetting.

In the domain of precise image analysis and preventive detection in vineyards, Wang et al. \cite{wang2024cognition} propose a lightweight semantic segmentation model for identifying grape picking points, enhancing the picking efficiency in vineyard environments, whereas Hollard and Mohimont \cite{hollard2023applying} apply KD to enhance grapevine detection for early yield prediction, focusing on lightweight model deployment for embedded devices. As far as it concern disease detection, Musa et al. \cite{musa2022low} propose a low-power DL model for detecting plant diseases in hydroponic systems, aiming at efficiency and reduced resource consumption and Zhang and Wang \cite{zhang2022mixkd} improve plant leaf disease recognition using a novel data augmentation-based KD framework, enhancing recognition accuracy in natural environments.

\textcolor{black}{More advanced and complex tasks have also been addressed using KD.} 
In the aquaculture domain, Yin et al. \cite{yin2024novel} propose a novel fish individual recognition method using KD within a vision transformer framework, improving accuracy. In the same application domain, Li et al. \cite{li2024t} focus on underwater fish species classification using a novel two-tier KD method to enhance model accuracy and reduce computational demands.
Back to the plants and trees images, Yang et al. \cite{yang2023fast} developed a fast pest detection algorithm using lightweight feature extraction and KD to enhance performance on edge devices and Wu et al.
\cite{wu2021deep} presented Deep BarkID, a lightweight CNN for tree species identification from bark images, tailored for use in forest environments with limited computing resources. Finally, Yamamoto \cite{yamamoto2019distillation} utilized CNNs to distill crop models to accelerate understanding of plant physiology, applying DL to evaluate environmental impact on grain yield.

\textcolor{black}{Researchers have also contributed to the development and deployment of lightweight student models on low capacity devices on the edge.}
Wenjie et al. \cite{wenjie2021distilled} discuss structured model compression via KD, transferring knowledge from a complex VGG16 model to a lightweight MobileNet. This approach significantly reduces model size and improves performance, making it suitable for deployment on devices with limited resources.
Wang et al. \cite{wang2023identification} explore lightweight model development for leaf image analysis, particularly for coffee leaf pest and disease identification. Using VGG as a teacher network, a student network is trained with KD, achieving high recognition accuracy and speed, which is crucial for real-time analysis.
In the same line, Li et al. \cite{li2024knowledge} delve into KD for instance-based semantic segmentation, particularly applying it to transform complex transformer models into more efficient DCNN architectures, which shows effectiveness in agricultural applications on datasets like BUP20 and SB20. Arablouei et al. \cite{arablouei2023situ} use KD to create compact models suitable for classifying animal behavior from accelerometry data on wearable devices. The models are optimized for real-time, in-situ performance on devices with limited computational resources.

Finally, there are works that take advantage of lightweight student models and combine them with heterogeneous remote sensor data to improve prediction accuracy. Castellano et al. \cite{castellano2023applying} study the application of KD for mapping weeds using UAVs in precision agriculture, developing a lightweight Vision Transformer-based model that provides effective weed mapping with minimal computational resources, and Bansal et al. \cite{bansal2024pa} develop a transformer-based network for plant growth monitoring, utilizing KD to enhance the model's performance by fusing RGB and depth image data for more accurate growth predictions.
Table \ref{tab:comparison_agriculture} provides a summary of the main works that employ KD in the agriculture domain.

\begin{table*}[t]
\centering
\caption{Comparison of Studies on KD in Agriculture}
\scriptsize
\begin{tabular}{|p{0.4cm}|p{2.5cm}|p{2.4cm}|p{4.8cm}|p{2.5cm}|p{3cm}|}
\hline
\textbf{Ref.} & \textbf{Model(s) Used} & \textbf{Dataset/Data Type} & \textbf{Main Contribution} & \textbf{Best Performance} & \textbf{Limitation} \\ \hline
\cite{liangde2021agricultural} & BERT-ALA + BiLSTM + CRF & Agriculture named entity data & Enhanced agricultural entity recognition using model distillation & Macro-F1 increased by 3.3\% & High time and space complexity \\ \hline
\cite{ghofrani2022knowledge} & MobileNet, Xception & PlantVillage dataset & Plant disease recognition with KD & Accuracy of 97.58\% & Limited to small architectures \\ \hline
\cite{hu2023lightweight} & Improved YOLOv5s & Maize leaf disease dataset & Lightweight model for maize leaf disease detection & mAP(0.5): Increased by 3.8\% & Only focuses on maize; may not generalize to other crops \\ \hline
\cite{li2023knowledge} & ResNet-34 & Various datasets for panoptic segmentation & KD for panoptic segmentation & Improved panoptic quality by up to 4.1 points & Requires extensive fine-tuning of balancing weights \\ \hline
\cite{jung2022plant} & Identical architecture for teacher and student & Large dataset for plant leaf segmentation & Improved instance segmentation using spatial embedding and KD & Enhanced segmentation accuracy & High dependency on the quality and size of the dataset \\ \hline
\cite{dong2024apple} & ECA-KDNet & Apple leaf dataset & Lightweight model for apple leaf disease diagnosis & Accuracy of 98.28\% & Focused only on apple leaves, might not generalize \\ \hline
\cite{huang2023knowledge} & YOLOR model variants & PlantDoc dataset & Multistage KD for plant disease detection & 60.4\% mAP@.5 & Model complexity and distillation stages may be challenging to manage \\ \hline
\cite{shen2021multi} & Multilevel distillation framework & CIFAR100 and CIFAR10 & Addressing low resolution identification problems & Improved low-resolution recognition accuracy & Specific to low-resolution datasets \\ \hline
\cite{wang2024cognition} & Lightweight semantic segmentation model & Custom dataset for grape picking point localization & Efficient grape picking point localization in complex environments & 91.08\% accuracy in picking point localization & Limited to grape picking, may not extend to other fruits \\ \hline
\cite{musa2022low} & Low-power DL model & Hydroponic systems & Plant disease detection in low-power IoT devices & Accuracy of 99.4\% & Focus on hydroponics; broader application unknown \\ \hline
\cite{zhang2022mixkd} & Data augmentation-based KD framework & PlantDoc dataset & Enhanced recognition accuracy for plant leaf diseases & Improved accuracy by up to 3.06\% & Performance heavily dependent on data augmentation quality \\ \hline
\cite{hollard2023applying} & Knowledge-distilled models & Datasets for grapevine detection & Early grape detection and yield prediction with KD & Improvement in various metrics, e.g., 13.63\% in mAP50-95 & Predominantly focused on early detection stages \\ \hline
\cite{wang2023identification} & Lightweight model using VGG for KD & Coffee leaf dataset & High accuracy in coffee leaf disease identification with a lightweight model & Accuracy of 96.73\% & Generalization to other crop diseases not demonstrated \\ \hline
\cite{arablouei2023situ} & GRU-MLP models, ResNet & Animal behavior datasets & In-situ animal behavior classification on wearable devices & MCC of 0.882 (ResNet) & Mainly applicable to animal behavior, not crops \\ \hline
\cite{wenjie2021distilled} & Distilled-MobileNet & Common diseases of crops & Lightweight disease recognition model for limited-resource devices & Accuracy of 97.62\% & Limited to specific diseases and crops \\ \hline
\cite{li2024knowledge} & Instance-based semantic segmentation with Mask2Former & Agricultural datasets & KD for instance semantic segmentation & AP improvement of 1.8 for ResNet-50 & Focused on specific types of segmentation \\ \hline
\cite{castellano2023applying} & Lightweight Vision Transformer & WeedMap dataset & Mapping weeds with drones using KD & F1 score of 0.863 & Specific to drone-based RS \\ \hline
\cite{bansal2024pa} & PA-RDFKNet & Various datasets for plant growth monitoring & RGB-depth fusion for plant age estimation with KD & MSE reduced from 2 to 0.14 weeks & Focused on plant growth, might not extend to other agricultural tasks \\ \hline

\cite{phan2022efficient} & KD from Multi-head Teacher (KDM) & Bio-HSI & Efficient hyperspectral image segmentation with a compact student network & mIoU of 90.03\% & Over-compression degrades performance without medium-sized teacher assistants \\ \hline
\cite{mane2023efficient} & UNet with various backbones & On-field images of pomegranate fruit & Effective segmentation of pomegranate fruits for agricultural automation & F1 score of 90.35\% for VGG19 backbone & Dependency on the choice of backbone for performance \\ \hline
\cite{yin2024novel} & Vision Transformer with chunking method & DlouFish dataset & Enhanced fish individual recognition using a novel KD strategy & Accuracy of 93.19\% & Specific to underwater environments \\ \hline
\cite{yang2023fast} & C3Faster with KD & CropPest6 dataset & Fast and efficient crop pest detection suitable for edge devices & 97.5\% mAP & Reduced feature extraction capability in lightweight models \\ \hline
\cite{wu2021deep} & Lightweight CNN models & Indiana Bark Dataset & Portable tree species identification system for smartphones & 96.12\% accuracy & Limited to specific tree species in Indiana \\ \hline
\cite{yamamoto2019distillation} & CNN & Crop growth dataset generated by a crop model & Learning plant physiology from crop models to enhance model portability & MSE of 52.9 during training & Limited by synthetic data generation from crop models \\ \hline
\cite{li2024t} & Two-tier KD (T-KD) & Fish37 dataset & Improved accuracy and reduced parameters for underwater fish species classification & Top-1 accuracy of 97.20\% & Requires large model sizes for initial training \\ \hline
\cite{tsagkatakis2021knowledge} & KD from multispectral to RGB models & Mullus Marbatus family dataset & Fish quality estimation using RGB cameras with knowledge from multispectral images & Classification accuracy of 84.3\% & Limited to specific types of fish and conditions \\ \hline
\cite{mengisti2024explainable} & ResNet50 and a lightweight student model & Dataset of Ethiopian medicinal plants & Accurate identification of medicinal plants using a distilled knowledge approach & 96.91\% accuracy & High accuracy dependent on extensive data preprocessing \\ \hline

\end{tabular}
\label{tab:comparison_agriculture}
\end{table*}


\subsubsection{Urban Planning}
KD has emerged as a vital technique in urban planning, particularly in the context of enhancing the efficiency and accuracy of models used for complex tasks such as environmental monitoring, infrastructure management, and resource allocation. For instance, KD has been effectively used to improve the real-time detection of building defects, optimize building extraction from noisy datasets, and enhance the accuracy of traffic flow prediction and travel time estimation. This technique is particularly valuable in scenarios involving large-scale urban data, where it enables the deployment of sophisticated models on resource-constrained devices, such as UAVs and edge computing frameworks, facilitating more efficient management of urban infrastructure and services. By enabling the transfer of knowledge from powerful teacher models to more efficient student models, KD supports the development of robust, scalable solutions that are essential for modern urban planning and the creation of smarter, more responsive cities For instance, Rithanasophon et al. \cite{rithanasophon2023quality} proposed a method that leverages deep CNNs (DCNNs) and KD to evaluate QoL for pedestrians using walkability data collected through virtual reality tools, achieving significant improvements in model performance and computational efficiency. Similarly, Liu et al. \cite{liu2023urbankg} introduced UrbanKG, a knowledge graph system that integrates KD for urban data fusion, showing promising results in boosting performance across various urban computing applications. Xu et al. \cite{xu2023building} also addressed the challenges of limited training samples in building polygon extraction by proposing BPDNet, a KD-based framework that effectively integrates generalization knowledge from large datasets with task-specific characteristics, resulting in superior performance in complex urban environments.

Federated learning frameworks have also benefited from KD, particularly in the context of land use monitoring and environmental impact assessment. Kanagavelu et al. \cite{kanagavelu2023fedukd} demonstrated the potential of integrating KD with federated UNet models for the semantic segmentation of satellite and street view images, achieving high accuracy and significant model compression. In a similar vein, Xu et al. \cite{xu2022improving} developed a KD-based building extraction method that reduces the impact of noise on model performance while maintaining generalization, achieving notable improvements in precision, recall, and IoU metrics. 
In the context of transportation systems, KD has been utilized to enhance travel time estimation (TTE) models and improve traffic flow prediction.  Yang et al. \cite{yang2023knowledge} proposed KDTTE, a deep neural network model that employs KD to reduce computation and memory costs while increasing accuracy, significantly outperforming state-of-the-art baselines in TTE tasks. In a different but related task, Li et al. \cite{li2024deep} applied deep KD to traffic flow prediction in spatio-temporal networks, demonstrating improvements in both local and global feature perception and achieving better accuracy in traffic predictions.

In the autonomous driving domain and the task of off-road environment segmentation, KD has been instrumental in improving model efficiency and accuracy. Pan et al. \cite{pan2023multitask} developed an end-to-end lane detection method using KD to guide polynomial regression under complex road conditions, achieving competitive results in efficiency and accuracy. Similarly, Kim and An \cite{kim2023knowledge} proposed a KD method for segmenting off-road environment range images, resulting in a favorable trade-off between segmentation performance and computational cost, highlighting its effectiveness for autonomous systems.

Lee et al. \cite{lee2021accelerating} proposed a high-speed detection method for multi-class defects on residential building façades using KD. The study demonstrated that applying KD to a lightweight DL model significantly improved mean average precision (mAP) by approximately 20\% and reduced inference time by 2.5 times, making it more suitable for real-time applications. 
Moving on, Chen et al. \cite{chen2023building} introduced a novel approach to building extraction that utilizes KD to enhance the robustness of the distilled student model. The study employed a multi-teacher collaborative distillation strategy to transfer comprehensive feature knowledge from teacher networks to the student model. The approach demonstrated state-of-the-art performance on multiple datasets, including the Massachusetts Roads Dataset, LRSNY Roads Dataset, and WHU Building Dataset, achieving high IoU scores and improving learning capabilities. 
Geng et al. \cite{geng2020topological} developed a lightweight topological space network for road extraction from optical RS images, leveraging KD. The study addressed the challenge of extracting topological features from complex road networks by proposing a topological space loss calculation model. The method resulted in significant improvements in accuracy and computational efficiency, demonstrating a good balance between performance and model size.

Besides, Li et al. \cite{li2022driver} proposed an off-policy imitation learning method for autonomous driving that employs task KD. This approach was designed to clone human driving behavior and transfer driving strategies to new, unseen scenarios. The method showed promising results in transferring knowledge to different illumination and weather conditions, enhancing route-following performance in realistic urban driving scenes.
Hong et al. \cite{hong2024knowledge} introduced a hierarchical edge-decision framework for intelligent transportation systems (ITS) that incorporates KD. The framework enables vehicle-road-cloud cooperation to enhance real-time motion planning by distilling complex spatial-temporal event reasoning into efficient decision-making processes. The method was validated on autonomous driving scenarios, demonstrating improved adaptability to complex environments.
Luo et al. (2022) \cite{luo2022keepedge} presented the KeepEdge framework, which integrates deep neural networks into an edge computing system for UAV-assisted parcel delivery. By employing KD, the study created a lightweight model that maintained high accuracy while reducing the computational load on UAVs. This approach proved effective in complex environments where traditional GPS-based positioning might fail. 
Pelizari et al. (2023) \cite{pelizari2023deep} developed a deep multitask learning (MTL) architecture for building characterization using street-level imagery. The study incorporated KD to encode cross-task interdependencies, which improved the generalization capabilities of the model across multiple natural hazards. The proposed MTL methods outperformed traditional single-task learning (STL) models, achieving higher accuracy and efficiency. The aforementioned studies that demonstrate the versatility of KD in enhancing the efficiency, accuracy, and scalability of models in various prediction and classification tasks in urban planning and intelligent transportation systems, using RS data are summarized in Table \ref{tab:kd_studies_comparison}.


\begin{table*}[h!]
\caption{Comparison of KD Studies for Urban Planning}
\centering
\scriptsize
\begin{tabular}{|p{0.4cm}|p{2.5cm}|p{2.4cm}|p{4.8cm}|p{2.5cm}|p{3cm}|}
\hline
\textbf{Citation} & \textbf{Model(s) Used} & \textbf{Dataset/Data Type} & \textbf{Main Contribution} & \textbf{Best Performance Value Achieved} & \textbf{Limitation} \\ \hline
\cite{rithanasophon2023quality} & DCNNs, LSTM, KD & VR-based questionnaire data & Evaluates walkability using AI and enhances real-time performance through KD & MSE of $7.19 \times 10^{-3}$ (within-city) and $9.73 \times 10^{-3}$ (across-cities) & Limited to VR data, may not generalize to all environments \\ \hline
\cite{liu2023urbankg} & FedUKD, UNet & Satellite and street view images & Integrates knowledge graphs for urban data fusion & 97\% accuracy on Chennai land use dataset & May struggle with dynamic, heterogeneous urban data \\ \hline
\cite{xu2023building} & BPDNet & Building polygons & Distills knowledge for generalization in building extraction tasks & IoU of 66.54\% & Performance may drop in complex urban settings \\ \hline
\cite{kanagavelu2023fedukd} & FedUKD & Satellite and street view images & Reduces communication costs in land use classification via federated learning & Above 95\% accuracy with significant compression & Scalability to other urban data types may be limited \\ \hline
\cite{yang2023knowledge} & KDTTE & Travel time estimation datasets & Improves travel time estimation with KD & 86.8\% accuracy improvement on Porto dataset & Limited generalization to diverse traffic conditions \\ \hline
\cite{liu2022developing} & UrbanKG & Urban spatial-temporal data & Develops an urban knowledge graph for data fusion & Effective in various urban applications & Requires extensive setup and integration \\ \hline
\cite{xu2022improving} & UPerNet, Swin Transformer & Noisy RS images & Enhances building extraction from noisy images with KD & IoU of 81.61\% & Dependent on noisy label quality \\ \hline
\cite{lee2021accelerating} & DCNN & Building façade images & Accelerates defect detection on building façades using KD & 20\% mAP increase, 2.5x faster inference & Limited to façade defects, may not generalize \\ \hline
\cite{chen2023building} & U-Net, DeepLabV3Plus & Road and building datasets & Enhances model robustness via multi-teacher distillation & IoU scores: 48.56\%, 79.51\%, 81.35\% & Teacher weight optimization is still needed \\ \hline
\cite{li2024deep} & Deep KD Model & Traffic flow datasets & Improves spatio-temporal traffic flow prediction using KD & Accuracy improvement of 0.19 and 0.18 on respective datasets & Focused on local data, may miss global patterns \\ \hline
\cite{pan2023multitask} & End-to-End Lane Detection with KD & TuSimple, CULane Datasets & Lane detection method using auxiliary supervision & Competitive accuracy, high efficiency & Post-processing still needed for some tasks \\ \hline
\cite{gupta2024towards} & Interaction-aware Trajectory Planning with KD & Real-world driving scenarios & Combines DL with optimization for trajectory planning & Fivefold improvement in computation time & Integration with control paradigms is complex \\ \hline
\cite{tsanakas2024light} & Lightweight Next Location Prediction Model & Mobility data & Efficient next-location prediction with reduced inference time & 6.57\% error reduction, 99.8\% faster inference & Focuses on reducing computational load \\ \hline
\cite{zhou2024mjpnet} & MJPNet-S* & RGB-T/D data & Trimodal joint-perception network for crowd density estimation & 92\% faster, 83\% fewer parameters & Reduced resource consumption may impact generalization \\ \hline
\cite{geng2020topological} & TSKD-Road & RS images & Topological network for road extraction with KD & Road IoU: 59.16\%, mIoU: 78.49\%, F1: 74.15\% & Limited to road extraction tasks \\ \hline
\cite{kim2023knowledge} & MobileNet\_v2 DeepLabV3+ with SLKD & Off-road environment dataset & Lightweight model for off-road segmentation using KD & mIoU of 57.28\%, low computational cost & Trade-off between accuracy and efficiency \\ \hline
\cite{hong2024knowledge} & GSCNN & Autonomous driving scenarios & Edge-decision framework for motion skill enhancement & Improved adaptation to dynamic environments & Complexity in real-time implementation \\ \hline
\cite{luo2022keepedge} & DNN & UAV delivery environments & Edge intelligence framework for UAV positioning & High accuracy with reduced model complexity & Dependent on visual data quality \\ \hline
\cite{pelizari2023deep} & Deep MTL & Street-level imagery & Cross-task interdependency modeling for building characterization & accuracy = 88.43\% & Complexity in MTL model training \\ \hline
\end{tabular}
\label{tab:kd_studies_comparison}
\end{table*}

\subsubsection{Oceanographic Monitoring }
KD can significantly enhance the efficiency and practicality of AI applications in ocean and sea studies by simplifying complex models for deployment on resource-constrained devices. This includes improving marine wildlife detection, real-time oceanographic monitoring, and underwater object detection by compressing large models into smaller, more efficient versions without compromising accuracy. Additionally, it can aid in climate change prediction and fisheries management, making advanced AI models more accessible and effective for monitoring and analysis in remote or resource-limited environments. 
In this direction, the authors in \cite{wang2023deepblue} explore the application of CNNs in ocean RS, highlighting their effectiveness in tasks such as 3D ocean field reconstruction, image super-resolution, and ocean phenomena forecasting. The study demonstrates significant improvements in classification accuracy for sea ice and open water areas in SAR images and a notable enhancement in image resolution using CNN-based models.

Several studies focus on underwater environments, where detection and analysis face unique challenges due to poor visibility and environmental complexities. Chen et al. \cite{chen2024online_xkd} propose an online KD framework, Online-XKD, to enhance the accuracy and generalizability of underwater object detection models while maintaining their lightweight nature. Similarly, Ben Tamou et al. \cite{ben2022live} present a CNN-based approach for classifying live reef fish species in underwater environments, using incremental learning to maintain high accuracy as new species are added. Another underwater-focused study introduces WaterMono \cite{ding2024watermono}, a framework for depth estimation and image enhancement in underwater scenes, leveraging KD to address challenges such as dynamic scenes and image degradation.

In the domain of geophysical field reconstruction, AdaptDeep \cite{wang2024self}, a self-supervised framework designed, has been proposed to reconstruct fine-grained spatial structures from coarse-scale geophysical data. The proposed method effectively identifies and recovers detailed information in sea surface temperature fields, demonstrating the potential of domain adaptation techniques in enhancing data resolution and accuracy. Moving on, Tropical cyclone (TC) wind radii estimation is the focus of Jin et al. \cite{jin2024towards}, who propose a multimodal fusion network, MT-TCNet, and its distillation variant, MT-TCNet-Distill. These models utilize a combination of satellite infrared images, wind field reanalysis, and maximum sustained wind speed data to estimate TC wind radii, achieving superior performance even in scenarios with incomplete data.

In the domain of water segmentation, the challenge of accurately segmenting water areas for unmanned surface vehicles (USVs) has been presented in \cite{zhang2023efficient}. The study introduced a multimodal fusion method combining 2D camera images and 3D LiDAR point clouds, utilizing transformers and KD to improve segmentation accuracy and processing speed.
Lastly, Yang et al. \cite{yang2023precise} focus on sea ice segmentation, proposing a CNN-based method enhanced with data augmentation, a novel loss function, and multiscale strategies. Their study achieves high segmentation accuracy using the HRNet-W48 backbone, demonstrating the effectiveness of innovative DL techniques in environmental monitoring. Table  \ref{tab:comparison_oceanographic} provides a summary of the studies that employ KD techniques to improve model performance in the oceanographic remote imaging domain.

\begin{table*}[ht]
\centering
\caption{Summary of Oceanographic Studies that employ KD}
\scriptsize
\begin{tabular}{|p{0.4cm}|p{2.5cm}|p{2.4cm}|p{4.8cm}|p{2.5cm}|p{3cm}|}
\hline
\textbf{Ref.} & \textbf{Model(s) Used} & \textbf{Dataset/Data Type} & \textbf{Brief Description of Main Contribution} & \textbf{Best Performance Value Achieved} & \textbf{Limitation} \\ \hline

\cite{wang2023deepblue} & CNNs & Various ocean RS data & Applied CNNs across multiple ocean RS tasks, including 3D ocean field reconstruction and image super-resolution. & ACC=92.36\% for sea ice and open water areas in SAR images & High computational cost and model interpretability challenges. \\ \hline

\cite{chen2024online_xkd} & Online-XKD & URPC2020 dataset & Enhanced feature extraction and generalization in underwater object detection using mutual knowledge transfer in a distillation framework. & 3.6 mAP improvement in student model detection accuracy & Complexity may hinder deployment in low-resource environments. \\ \hline

\cite{ben2022live} & CNN with incremental learning & LifeClef 2015 Fish dataset & Developed an incremental learning strategy for live reef fish species classification, maintaining high performance on previously learned species. & 81.83\% accuracy on LifeClef 2015 Fish benchmark dataset & Scaling to larger datasets or complex environments could be challenging. \\ \hline

\cite{wang2024self} & AdaptDeep & Coarse and fine-scale geophysical field data & Proposed a self-supervised framework for fine-grained reconstruction of geophysical data using domain adaptation and contrastive learning. & Recovered 81.2\% detailed information in sea surface temperature fields & Performance depends on the availability of coarse-scale data and temporal correlations. \\ \hline

\cite{ding2024watermono} & WaterMono & Underwater images & Introduced a self-supervised depth estimation framework with image enhancement for underwater environments using KD. & RMSE: 0.945, RMSE log: 0.152 & Limited generalization to diverse camera angles and extreme conditions. \\ \hline

\cite{jin2024towards} & MT-TCNet, MT-TCNet-Distill & Multimodal data including satellite IR images, reanalysis wind fields, and MSW speed & Developed a multimodal fusion network and distillation method for robust TC wind radii estimation with both complete and missing modalities. & R34 estimation: RMSE 22.458 nmi, MAE 16.577 nmi, R-value 0.855; RMW estimation: RMSE 7.958 nmi, MAE 5.689 nmi, R-value 0.738 & Reliance on reanalysis data limits real-time applicability. \\ \hline

\cite{zhang2023efficient} & Transformer-based multimodal fusion & 2D camera images, 3D LiDAR point clouds & Proposed a water segmentation method using transformers and KD for improved 2D image-based segmentation with faster speed. & Approx. 1.5\% improvement in accuracy and MaxF, speed of 15-110 fps & High computational load during training phase, though reduced with distillation. \\ \hline

\cite{yang2023precise} & CNN with HRNet-W48 backbone & Large sea-ice segmentation dataset & Introduced innovative data augmentation, loss function, and multiscale strategies for accurate sea ice segmentation with KD for real-time application. & FWIoU score of 97.8439 & High computational resource requirement for real-time processing. \\ \hline

\cite{chen2020learning} & Tiny YOLO-Lite & SSDD, HRSID, large-scene SAR images & Developed a lightweight SAR ship detector using network pruning and KD to reduce model size and computation while maintaining high accuracy. & Average Precision (AP) of 89.07\%, 2.8 MB model size, inference speed >200 fps & Performance may decline with further model size reduction. \\ \hline

\end{tabular}
\label{tab:comparison_oceanographic}
\end{table*}

\section{Challenges and Limitations}

\begin{figure*}[t!]
\begin{center}
\includegraphics[width=0.7\textwidth]{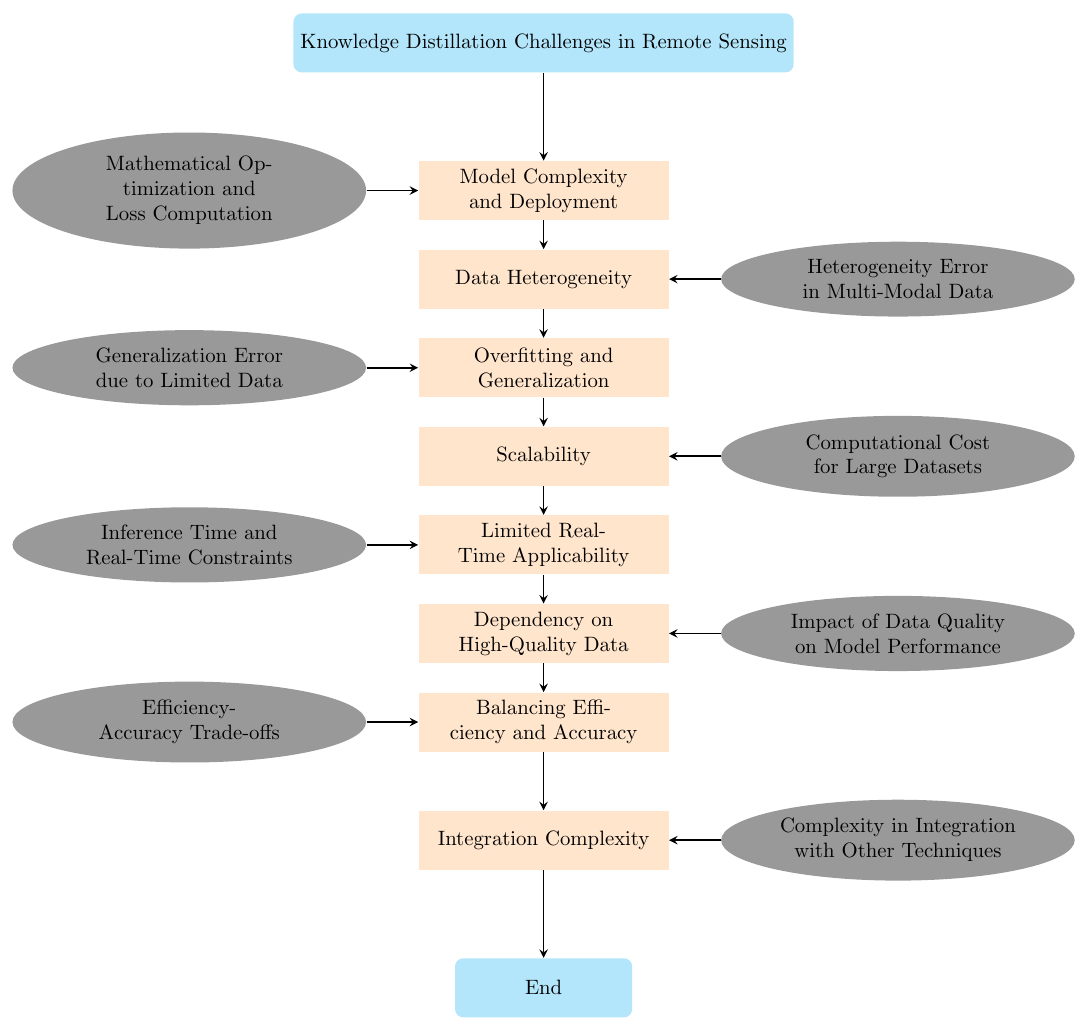}\\
\end{center}
\caption{Challenges and Limitations of KD in RS.}
\label{fig:limitations_KD}
\end{figure*}

\textcolor{black}{Despite the many advantages that KD techniques offer and the wide range of their applications, they still face several limitations as portrayed in Fig. \ref{fig:limitations_KD}. These challenges are mainly related to the deployment of the models to resource-constrained devices and to keeping the performance of these models high when handling heterogeneous data or data from new, unseen distributions. Finding the balance between model efficiency and prediction accuracy is the key challenge as explained in the following.}

\subsection{Model Complexity and Deployment}
In RS applications, KD is often employed to create smaller, more efficient student models by transferring knowledge from a larger, more complex teacher model. The main goal of KD is to retain the high accuracy of the teacher model while reducing the computational load and model size, which is crucial for deployment on resource-constrained devices commonly used in RS.
However, the process of optimizing the distillation loss function to achieve this balance between model size and performance is inherently complex and computationally demanding. The distillation loss \(\mathcal{L}_{\text{KD}}\), which combines the cross-entropy loss \(\mathcal{L}_{\text{CE}}\) and the Kullback-Leibler divergence \(\mathcal{L}_{\text{KL}}\), needs to be carefully minimized to ensure that the student model effectively approximates the teacher model. This optimization process becomes more challenging as the complexity of the teacher model increases, leading to a higher computational burden during training \cite{li2022distilling}.
Furthermore, when deploying the distilled student model on resource-constrained devices, such as those used in RS for on-board data processing, the reduced model complexity must still meet the real-time processing requirements and maintain high accuracy. The complexity of optimizing the KD process for deployment is expressed by the computational cost associated with the gradient of the distillation loss with respect to the student model parameters. As this complexity increases, it can lead to longer training times, higher energy consumption, and potentially suboptimal model performance, particularly when deployed in environments with limited computational resources. In this reagrd,  given a teacher model \( T \) with parameters \( \theta_T \) and a student model \( S \) with parameters \( \theta_S \), the distillation loss is defined as:

\begin{equation}
\mathcal{L}_{\text{KD}}(\theta_S) = \alpha \mathcal{L}_{\text{CE}}(y, S(x; \theta_S)) + (1-\alpha) \mathcal{L}_{\text{KL}}(T(x; \theta_T), S(x; \theta_S))
\end{equation}

The challenge related to the complexity of optimizing this loss function for deployment is defined as follows:

\begin{equation}
\text{Complexity} \propto O\left(\frac{\partial \mathcal{L}_{\text{KD}}}{\partial \theta_S}\right)
\end{equation}

This complexity can affect the feasibility and effectiveness of deploying KD models in real-world RS scenarios, where computational efficiency and model robustness are critical.

\subsection{Data Heterogeneity}
RS data often comes from multiple modalities, such as optical, SAR, and multispectral sensors. Integrating knowledge across these heterogeneous data sources while maintaining accuracy is challenging, as the characteristics of the data can vary significantly. For multi-modal RS data \( x_1, x_2, \dots, x_m \) from different modalities, the aggregate loss is:
\begin{equation}
\mathcal{L}_{\text{multi-modal}}(\theta_S) = \sum_{i=1}^{m} w_i \cdot \mathcal{L}_{\text{KD}}(T_i(x_i; \theta_T), S(x_i; \theta_S))
\end{equation}
Typically, the heterogeneity error is defined as:

\begin{equation}
\begin{split}
\text{Heterogeneity Error} = & \sum_{i=1}^{m} \Big| \mathcal{L}_{\text{KD}}(T_i(x_i; \theta_T), S(x_i; \theta_S)) \\
                             & - \mathcal{L}_{\text{KD}}(T_j(x_j; \theta_T), S(x_j; \theta_S)) \Big|
\end{split}
\end{equation}

The performance of knowledge distillation can be reduced by data heterogeneity due to the following reasons.
RS data from different modalities often exhibit vastly different characteristics, leading to inconsistencies in feature representations. The student model may struggle to generalize well across all modalities, resulting in higher overall prediction errors.
Moving on, the loss landscape associated with each modality differs due to the inherent characteristics of the data. The aggregate loss function, \(\mathcal{L}_{\text{multi-modal}}(\theta_S)\), combines the knowledge from multiple teacher models. The heterogeneity in data causes the loss functions for each modality to diverge, making it difficult to effectively minimize the combined loss.
Additionally, the heterogeneity error quantifies the discrepancy between the losses associated with different modalities. A large heterogeneity error indicates misalignment in the distilled knowledge, which can cause the student model to perform poorly on certain modalities.
Lastly, determining the appropriate weights \(w_i\) for each modality's contribution to the overall loss is challenging. Incorrectly weighting a modality can lead to an imbalance, where the student model prioritizes less important or noisier data, further degrading performance.

\subsection{Overfitting and Generalization}

While KD helps reduce the size of models, it can also lead to overfitting, particularly when the student model is trained on a limited dataset. This results in poor generalization to new, unseen data, which is critical for the success of RS applications \cite{wang2021knowledge,li2021online}. The generalization error is given by:
\begin{equation}
\text{Generalization Error} = \mathcal{L}_{\text{KD}}(\theta_S; D_{\text{test}}) - \mathcal{L}_{\text{KD}}(\theta_S; D_{\text{train}})
\end{equation}
Overfitting occurs when: $\text{Generalization Error} \gg 0$. The impact of overfitting on the performance of knowledge distillation in RS applications includes: Overfitting causes the student model to perform well on the training data but poorly on unseen test data, which can significantly reduce the model's effectiveness in real-world RS applications where data variability is high; Overfitted models are often overly sensitive to noise in the training data. This sensitivity can lead to incorrect predictions when the model encounters noisy or outlier data in RS, where data quality can vary widely across different sensors and conditions; In RS, data is often collected from multiple modalities (e.g., optical, SAR, multispectral). An overfitted student model might fail to generalize well across these different modalities, leading to inconsistent performance and reduced reliability in practical applications; Overfitting can limit the ability of the student model to transfer learned knowledge to new tasks or domains within RS, reducing the versatility and adaptability of the distilled model.

\subsection{Scalability}
As the size of RS datasets increases, the computational complexity of applying KD also increases. This scalability issue can limit the practicality of deploying distilled models on large datasets.

For a dataset of size \( N \), the computational complexity scales as:
\begin{equation}
\text{Scalability} \propto O(N \cdot \mathcal{C}(\mathcal{L}_{\text{KD}}))
\end{equation}
\noindent where \( \mathcal{C}(\mathcal{L}_{\text{KD}}) \) is the computational cost of evaluating the distillation loss.

Scalability challenges in knowledge distillation (KD) for RS applications can significantly affect performance by increasing training times, requiring more computational resources, and decreasing model efficiency. As datasets grow larger, the time needed to train student models becomes prohibitively long, making iterative improvements difficult. Resource constraints, particularly on devices used in RS, limit the ability to handle large datasets effectively. Additionally, scalability issues can reduce the efficiency of distilled models, complicate real-time data processing, and hinder the integration of data from multiple sources, ultimately leading to suboptimal knowledge transfer.

\subsection{Limited Real-Time Applicability}
In RS applications, the need for real-time processing is critical, as delays in data processing can render the information outdated and less useful for immediate decision-making. Knowledge Distillation (KD) aims to create more efficient models, but even with distilled models, achieving the required inference speed can be challenging, especially when dealing with complex student models. The inference time \( t_{\text{inference}} \), which depends on both the size of the student model \( \theta_S \) and the computational complexity of evaluating the distillation loss \( \mathcal{C}(\mathcal{L}_{\text{KD}}) \), must be kept within the real-time processing limit \( t_{\text{real-time}} \). If \( t_{\text{inference}} \) exceeds \( t_{\text{real-time}} \), the performance of the KD model is compromised, as it may not be able to process data quickly enough to be useful in time-sensitive RS applications. This limitation can reduce the effectiveness of KD in scenarios where immediate data analysis and decision-making are required \cite{wang2021knowledge,li2021online}.

\begin{equation}
t_{\text{inference}} \leq t_{\text{real-time}}
\end{equation}
where \( t_{\text{inference}} \) is the inference time and \( t_{\text{real-time}} \) is the allowable time for real-time processing:

\begin{equation}
t_{\text{inference}} \approx O(\theta_S) + O(\mathcal{C}(\mathcal{L}_{\text{KD}}))
\end{equation}

\subsection{Dependency on High-Quality Data}
The effectiveness of Knowledge Distillation (KD) is highly contingent on the quality of the training data, which plays a critical role in the success of the distillation process. In RS applications, the data is often noisy, sparse, or collected under varying conditions, leading to inconsistencies that can adversely impact the KD process. The distillation loss, which combines the cross-entropy loss \(\mathcal{L}_{\text{CE}}\) between the student model's predictions and the true labels, and the Kullback-Leibler divergence \(\mathcal{L}_{\text{KL}}\) between the teacher and student model outputs, assumes that the input data is of high quality. However, when the data is of poor quality, the student model may struggle to learn effectively from the teacher model, leading to increased errors and reduced generalization capability. The formula for the impact of data quality indicates that as the proportion of low-quality data increases, the overall performance of the KD model diminishes. This reduction in performance can result in less robust models, which may fail to accurately process and interpret RS data, ultimately hindering the effectiveness of KD in real-world RS tasks \cite{binici2022preventing,zhang2020reliable,chen2023consistency}.

\begin{equation}
\mathcal{L}_{\text{KD}}(\theta_S) = \sum_{i=1}^{n} \left(\mathcal{L}_{\text{CE}}(y_i, S(x_i; \theta_S)) + \mathcal{L}_{\text{KL}}(T(x_i; \theta_T), S(x_i; \theta_S))\right)
\end{equation}

Typically, the impact of data quality is defined as follows:

\begin{equation}
\text{Data Quality Impact} = \frac{1}{n} \sum_{i=1}^{n} \mathbb{I}(\text{quality}(x_i, y_i) < \epsilon)
\end{equation}

\subsection{Balancing Efficiency and Accuracy}
One of the key challenges in Knowledge Distillation (KD) is striking the right balance between model efficiency and accuracy. In the context of RS applications, where the stakes are often high, such as in disaster monitoring or environmental protection, compressing the student model too much in the pursuit of efficiency can lead to a significant loss in accuracy. This reduction in accuracy could result in the failure to correctly interpret RS data, leading to erroneous decisions \cite{mishra2018apprentice,stanton2021does}. The trade-off between efficiency and accuracy is represented by the following relationships:

\begin{equation}
\text{Efficiency} \propto \frac{1}{\text{Model Size}(\theta_S)}, \quad \text{Accuracy} \propto \frac{1}{\mathcal{L}_{\text{KD}}(\theta_S)}
\end{equation}

\noindent where \( \text{Model Size}(\theta_S) \) refers to the number of parameters in the student model, and \( \mathcal{L}_{\text{KD}}(\theta_S) \) is the distillation loss, which is inversely proportional to the accuracy of the student model.

The optimization problem, therefore, involves maximizing the product of efficiency and accuracy:

\begin{equation}
\max_{\theta_S} \left(\frac{1}{\text{Model Size}(\theta_S)} \cdot \frac{1}{\mathcal{L}_{\text{KD}}(\theta_S)}\right)
\end{equation}

However, this optimization is complex because increasing efficiency (i.e., reducing model size) often leads to a rise in distillation loss \( \mathcal{L}_{\text{KD}} \), which in turn decreases accuracy. Conversely, maintaining high accuracy may require a larger model size, reducing efficiency. In RS, where both computational resources and model performance are critical, failing to achieve an optimal balance can limit the effectiveness of KD. This trade-off must be carefully managed to ensure that the compressed model performs adequately in practical RS scenarios, where both speed and accuracy are essential.

\subsection{Integration Complexity}
Integrating Knowledge Distillation (KD) with other techniques such as multi-modal fusion or domain adaptation introduces significant complexity to the model and its training process, affecting its performance in RS applications. Integrating these techniques requires careful balancing of multiple loss functions, as the overall performance now depends on the combined effectiveness of KD, multi-modal learning, and domain adaptation. This added complexity can make the training process more computationally expensive, harder to optimize, and more prone to issues like overfitting or convergence to suboptimal solutions. For instance, when integrating KD with other techniques, the overall loss function is expressed as a combination of multiple loss components, each weighted by a coefficient (e.g., $\alpha$, $\beta$, $\gamma$). The need to fine-tune these coefficients to achieve the desired balance between KD, multi-modal fusion, and domain adaptation further complicates the training process \cite{yu2024incremental,lin2024component}. Moreover, the integration complexity, represented by the derivative of the total loss function concerning the student model parameters, reflects the increased difficulty in optimizing the student model. As the complexity increases, the risk of inefficient training or suboptimal performance also rises, making it challenging to achieve the desired accuracy and efficiency in real-world RS applications.


\section{Future Directions}

\subsection{Advanced Model Compression Techniques}

\subsubsection{Dynamic Distillation}  
Dynamic Distillation is a technique within the broader category of advanced model compression. It aims to optimize the student model's performance by dynamically adjusting its complexity based on the specific characteristics of the input data or the task at hand. The core idea behind dynamic distillation is to create a flexible and adaptive student model that can efficiently learn from the teacher model without being overly constrained by a fixed architecture or predetermined level of complexity \cite{zhu2024dynamickd,liang2024dynamic}.
In many RS applications, the complexity of the input data can vary significantly. For example, a satellite image of a dense urban area may contain more intricate features than an image of a rural landscape. A static, one-size-fits-all student model may struggle to balance performance across such diverse inputs. Dynamic distillation allows the student model to adapt its architecture or parameters based on the specific task, ensuring that it allocates resources efficiently \cite{yu2023urban,ye2022method}.
Not all inputs require the same level of processing. Dynamic distillation enables the student model to adjust its complexity (e.g., the number of layers, the size of feature maps, or the degree of feature extraction) depending on the input. For instance, simpler inputs might be processed with a reduced version of the student model, while more complex inputs trigger a more detailed processing approach \cite{zhang2021self}.
The term "dynamic" implies that these adjustments occur in real-time or near-real-time, during the inference phase. This is particularly useful in resource-constrained environments, such as edge devices or real-time RS applications, where computational resources are limited. By making on-the-fly adjustments, the model can maintain high performance while conserving resources \cite{zhang2021learning}.
The dynamic distillation process can be expressed as an optimization problem:

\begin{equation}
\min_{\theta_S} \mathbb{E}_{x \sim \mathcal{D}} \left[ \lambda(x) \cdot \mathcal{L}_{\text{KD}}(T(x; \theta_T), S(x; \theta_S)) \right]
\end{equation}
Where \( \theta_S \) represents the parameters of the student model that need to be optimized. The input sample, denoted as \( x \), is drawn from the data distribution \( \mathcal{D} \). The teacher model, parameterized by \( \theta_T \), produces an output \( T(x; \theta_T) \) for the input \( x \), while the student model, with parameters \( \theta_S \), generates an output \( S(x; \theta_S) \) for the same input. The KD loss function, \( \mathcal{L}_{\text{KD}} \), typically quantifies the difference between the teacher's and student's outputs. Additionally, \( \lambda(x) \) is a dynamic weighting function that adjusts the contribution of each input \( x \) to the overall loss, depending on its complexity or the specific requirements of the task.

Specifically, in dynamic distillation, the factor \( \lambda(x) \) acts as a critical gatekeeper, adjusting the influence of each input on the student model's training. For complex or critical inputs, \( \lambda(x) \) increases, prompting the student model to allocate more resources, such as deeper layers or enhanced feature extraction. Conversely, simpler inputs lead to a lower \( \lambda(x) \), allowing the student model to process the data more efficiently with reduced resources. The KD loss function \( \mathcal{L}_{\text{KD}} \) is central to this process, focusing on minimizing the difference between the teacher and student models' outputs to ensure the student effectively mimics the teacher. This approach is generalized across the entire dataset, as captured by the expectation \( \mathbb{E}_{x \sim \mathcal{D}} \), optimizing the student model's performance across diverse inputs.

Dynamic distillation enhances resource efficiency by dynamically adjusting the student model's complexity based on the input, ensuring that computational resources are used optimally, especially in resource-constrained environments. This adaptability allows the student model to maintain or even surpass the performance of static models, particularly when dealing with heterogeneous datasets like those in RS. Additionally, the scalability of dynamic distillation makes it a versatile solution that is suitable for deploying ML models across various environments, from cloud-based systems to edge devices.
\textcolor{black}{Typically, in RS, dynamic distillation could be applied to urban monitoring where satellite images vary significantly between dense urban areas and sparse rural regions. For instance, when analyzing satellite imagery for urban heat island detection, the student model could dynamically adjust its complexity, using more layers and features for complex urban environments with varied structures, while simplifying its approach for less complex rural landscapes, thus optimizing processing efficiency and accuracy.}


\subsubsection{Layer-Wise Distillation:}  
Layer-wise distillation is an advanced technique in model compression that focuses on transferring knowledge from a teacher model to a student model at a more granular level \cite{du2024learning}. Unlike traditional KD, which typically focuses on aligning the final outputs of the teacher and student models, layer-wise distillation involves aligning the outputs of corresponding layers in both models \cite{kokane2024improving}. This approach ensures that the student model learns not only the final output distribution but also the intermediate representations that the teacher model uses to arrive at that output \cite{kim2024ladimo}.
Typically, layer-wise distillation enables a more effective transfer of knowledge by focusing on the outputs of different layers within a complex model, where each layer captures varying levels of abstraction—from basic features like edges to more complex patterns \cite{zhong2024self}. This approach allows for tailored compression by assigning different importance to each layer, ensuring that critical features are preserved while less important layers are compressed more aggressively. As a result, the student model becomes more compact and maintains or improves performance, particularly in tasks requiring detailed understanding, such as detecting and classifying intricate patterns in RS applications \cite{liang2024module}.
The layer-wise distillation process can be expressed as follows:

\begin{equation}
\mathcal{L}_{\text{KD-layer}} = \sum_{l=1}^{L} w_l \cdot \mathcal{L}_{\text{KD}}(T^l(x), S^l(x))
\end{equation}
where \( T^l(x) \) and \( S^l(x) \) denote the outputs of the \( l \)-th layer in the teacher and student models, respectively, while \( \mathcal{L}_{\text{KD}} \) represents the knowledge distillation loss function applied to these corresponding layer outputs. The term \( w_l \) is a weight assigned to each \( l \)-th layer, indicating its significance in the distillation process, and \( L \) denotes the total number of layers in the model.

In layer-wise distillation, \( T^l(x) \) and \( S^l(x) \) represent the outputs of the \( l \)-th layer for a given input \( x \) in the teacher and student models, respectively, ensuring the student model learns the same feature representations as the teacher model at each stage. The KD loss function, \( \mathcal{L}_{\text{KD}} \), measures the difference between these outputs, and when applied layer-wise, it ensures close alignment between the corresponding layers of both models. The layer-specific weight \( w_l \) allows for fine-tuning the importance of each layer, with critical layers in the teacher model being given higher weights to ensure their knowledge is effectively transferred. Finally, the summation across all layers \( L \) ensures that the distillation process comprehensively covers the entire model, enabling the student model to replicate the full range of the teacher model's capabilities.

Layer-wise distillation offers several practical benefits, including enhanced feature preservation, where focusing on each layer ensures that the student model retains the critical features learned by the teacher, leading to greater accuracy and capability. The flexibility in compression, enabled by layer-specific weights, allows for optimizing the trade-off between model size and performance, depending on the application's needs. Additionally, this approach fosters better generalization, as the student model is trained to replicate the hierarchical representations of the teacher model, making it more adept at handling new data, particularly in tasks that require detailed feature extraction and classification.
\textcolor{black}{
In RS, layer-wise distillation can be applied to multispectral image classification, where different spectral bands capture varying levels of detail. By aligning the outputs of corresponding layers in both the teacher and student models, this technique ensures that the student model effectively learns the intermediate features critical for distinguishing complex land cover types, such as differentiating between various crop types or identifying subtle changes in vegetation health.}

\subsection{Efficient Training and Inference}

\subsubsection{Low-Cost Training Algorithms}  
Low-cost training algorithms aim to reduce the computational burden associated with training both teacher and student models, which is particularly important in the context of KD where the goal is to make the student model as efficient as possible. These algorithms focus on optimizing various aspects of the training process to minimize costs while maintaining or even enhancing the performance of the distilled models \cite{park2024cosine}.
Developing more efficient training algorithms that reduce the computational burden of training both teacher and student models is crucial. This can involve leveraging techniques such as federated learning, transfer learning, or smaller proxy datasets for initial training to minimize the overall training cost \cite{lu2024data}.
Federated learning is a distributed approach that allows training to occur across multiple devices or servers without the need to centralize the data. This can significantly reduce the computational cost associated with data processing and model training by distributing the workload \cite{shao2024selective}. Each device trains a local model using its data and periodically shares updates with a central server, which aggregates these updates to improve the global model. This approach not only reduces the computational burden on individual devices but also enhances privacy since raw data is not shared \cite{qiao2024knowledge,yang2024unideal}.
Transfer learning involves taking a pre-trained model (often trained on a large dataset) and fine-tuning it on a smaller, task-specific dataset. This approach can drastically reduce the training cost because the model has already learned general features from the larger dataset, and only minimal additional training is required to adapt it to the new task. In the context of KD, transfer learning can be used to initialize the teacher model, which then distills its knowledge to a student model with minimal additional training \cite{zhong2024panda,gou2024collaborative}.
Using smaller proxy datasets for initial training can also reduce costs. Proxy datasets are subsets of the original data or synthetic datasets that approximate the characteristics of the full dataset but are much smaller in size. Training on these datasets requires fewer resources and can provide a good initial model that can be further refined with the full dataset. This approach is particularly useful in scenarios where obtaining labeled data is expensive or time-consuming \cite{wu2024exploring,le2024cdkt}. 
The cost of training both the teacher and student models can be expressed as:
\begin{equation}
C_{\text{train}} = \sum_{i=1}^{N} \left( C_{\text{data}}(x_i) + C_{\text{model}}(T, S) \right)
\end{equation}
where \( C_{\text{data}}(x_i) \) denotes the cost associated with processing each data sample \( x_i \), which encompasses activities such as data loading, augmentation, and preprocessing, while \( C_{\text{model}}(T, S) \) refers to the cost incurred during the training process for updating both the teacher model \( T \) and the student model \( S \).
This formulation highlights that the total training cost \( C_{\text{train}} \) is the sum of the costs associated with processing all data samples and updating the models. By optimizing these components—such as by reducing the size of the data samples with proxy datasets, distributing the training workload with federated learning, or leveraging pre-trained models with transfer learning—the overall training cost can be significantly reduced \cite{lu2024data}.
Low-cost training algorithms enhance resource efficiency, allowing organizations to train effective models even in environments with limited computational resources, which is particularly valuable in fields like RS that involve large datasets and complex models. These algorithms also support scalability, enabling the handling of extensive datasets and sophisticated models without a proportional increase in resource demands, making them adaptable to various environments from cloud servers to edge devices \cite{xu2024self}. Additionally, by reducing training costs and time, these algorithms facilitate faster development cycles, allowing for quicker iteration and deployment of ML models, which is essential in rapidly evolving fields like AI \cite{zhao2024data}.
\textcolor{black}{Low-cost training algorithms can be applied to disaster response scenarios, where real-time analysis of satellite imagery is crucial. For example, federated learning can be used to train models across multiple local servers situated near disaster zones, enabling rapid analysis of satellite images for damage assessment without the need for extensive centralized computing resources. Transfer learning can further enhance this process by fine-tuning pre-trained models on smaller, region-specific datasets, ensuring swift deployment and effective monitoring during critical events.}

\subsubsection{Hardware-Aware Distillation}  
Integrating KD with hardware-aware design principles can optimize models specifically for the hardware on which they will be deployed, such as edge devices or GPUs. This approach aims to balance the distillation process with the computational capabilities of the target hardware \cite{balaskas2024hardware}. Hardware-aware distillation integrates KD with hardware-specific optimization strategies to create models that are not only efficient in terms of performance but are also tailored to the computational constraints of the hardware on which they will be deployed. This approach is particularly useful for scenarios where the model needs to be run on edge devices, GPUs, or other specialized hardware, ensuring that the distilled model operates within the physical and computational limits of the target platform \cite{wang2024all}.

Hardware-aware distillation seeks to balance the effectiveness of the knowledge transfer process with the computational efficiency required by the target hardware. The goal is to ensure that the student model retains as much of the teacher model's performance as possible while also fitting within the hardware's resource constraints \cite{li2024quasar}. This involves careful consideration of factors such as memory usage, processing speed, and power consumption, which are critical in environments like mobile devices, embedded systems, or cloud-based GPUs \cite{ghebriout2024harmonic}. Different hardware platforms have varying capabilities and limitations. For example, GPUs excel at parallel processing but may have limited memory bandwidth, while edge devices often have strict power and computational limits. Hardware-aware distillation tailors the student model to leverage the strengths of the target hardware while minimizing its weaknesses. This could involve optimizing the model's architecture to reduce the number of parameters, simplify computations, or increase parallelism, depending on the hardware's characteristics \cite{baek2024bit}. The regularization parameter \( \lambda \) in the hardware-aware distillation framework controls the trade-off between the accuracy of the distilled model and its hardware efficiency. A higher \( \lambda \) places more emphasis on minimizing computational costs, potentially sacrificing some accuracy for greater efficiency. Conversely, a lower \( \lambda \) prioritizes accuracy, allowing for more complex models that may require more computational resources. The choice of \( \lambda \) depends on the specific requirements of the application and the hardware. The optimization objective for hardware-aware distillation can be expressed as:

\begin{equation}
\min_{\theta_S} \mathcal{L}_{\text{KD}}(T, S) + \lambda \cdot C_{\text{hardware}}(\theta_S)
\end{equation}
where \( \mathcal{L}_{\text{KD}}(T, S) \) refers to the knowledge distillation loss function, which quantifies the discrepancy between the outputs of the teacher model \( T \) and the student model \( S \). The term \( C_{\text{hardware}}(\theta_S) \) represents the computational cost associated with running the student model \( S \) on specific hardware, encompassing factors such as inference time, memory usage, and power consumption. The regularization parameter \( \lambda \) is introduced to balance the trade-off between reducing the distillation loss and optimizing hardware efficiency. 
This formulation ensures that the student model is not only accurate but also optimized for the computational environment in which it will be deployed. Typically, hardware-aware distillation enhances the feasibility of deploying advanced ML models in resource-constrained environments by tailoring the student model to the specific hardware, making it particularly valuable for edge computing scenarios with strict power, memory, and processing limits. This approach leads to improved performance on the target hardware, offering faster inference times, lower power consumption, and more efficient memory usage, thereby optimizing model deployment in real-world applications. Additionally, hardware-aware distillation allows for customization to meet the unique requirements of various deployment environments, ensuring that models are optimized whether deployed on high-performance GPUs in data centers or low-power microcontrollers in IoT devices \cite{bouzidi2024efficient}.
\textcolor{black}{Hardware-aware distillation can be applied to real-time monitoring on edge devices, such as drones used for precision agriculture. By optimizing the student model to operate efficiently within the power and computational constraints of these drones, the model can quickly process high-resolution imagery to detect crop health issues or identify weeds, enabling swift, in-field decision-making without relying on cloud-based resources. This approach ensures that advanced analysis can be performed directly on the edge, even in remote or resource-limited environments.}

\subsection{Improving Data Quality and Robustness}
\subsubsection{Robust Distillation Against Noisy Data}  
The effectiveness of KD heavily relies on the quality and quantity of training data. Robust distillation techniques need to be developed to handle noisy, sparse, or imbalanced datasets, which are common in RS. Robust Distillation Against Noisy Data focuses on enhancing the resilience of the KD process when dealing with imperfect data \cite{wang2023self}. In real-world applications, particularly in RS, datasets often contain noise, inconsistencies, or imbalances that can degrade the performance of ML models. This approach aims to mitigate the impact of such issues by incorporating robustness into the distillation process, ensuring that the student model can still learn effectively even when the data is not ideal \cite{fang2023reliable}.
In RS, noisy data is common due to various factors like sensor errors, atmospheric interference, or mislabeling during data collection. Standard KD techniques may struggle with such data, leading to poor model performance \cite{tian2024adaptive,shao2024jointnet}. Robust distillation techniques address this by explicitly modeling and compensating for the noise during the training process. This can involve using techniques that identify and either correct or down-weight the influence of noisy samples on the student model \cite{tran2024nayer,wang2024continuous}.
The key to robust distillation is modifying the loss function to account for the presence of noise. The traditional KD loss function, which measures the difference between the teacher and student models, is augmented with a term that penalizes the model based on the amount of noise in the data. This term is controlled by a weighting factor \( \eta \), which determines how much influence the noise has on the overall learning process. By doing so, the student model becomes more robust to the effects of noise, learning to focus on cleaner, more reliable data \cite{park2024leveraging}.
Robust distillation techniques must strike a balance between learning from noisy and clean data. While it is important to minimize the negative impact of noise, completely ignoring noisy data could result in a loss of valuable information \cite{liu2024learning}. Therefore, these techniques aim to optimize the learning process by allowing the student model to still extract useful knowledge from noisy data while minimizing the distortion it causes \cite{li2024extracting}.
The formulation for robust distillation against noisy data can be expressed as:

\begin{equation}
\mathcal{L}_{\text{KD-robust}} = \mathbb{E}_{x, \tilde{x} \sim \mathcal{D}} \left[ \mathcal{L}_{\text{KD}}(T(\tilde{x}), S(x)) + \eta \cdot \text{Noise}(x, \tilde{x}) \right]
\end{equation}
where \( \mathcal{L}_{\text{KD}}(T(\tilde{x}), S(x)) \) denotes the standard knowledge distillation loss, capturing the discrepancy between the teacher model's output on noisy data \( \tilde{x} \) and the student model's output on corresponding clean data \( x \). The weighting factor \( \eta \) regulates the influence of the noise term in the overall loss function, with a higher \( \eta \) placing greater emphasis on noise correction, thereby enhancing the model's robustness to noise. The term \( \text{Noise}(x, \tilde{x}) \) quantifies the noise level between the clean and noisy data, using metrics such as mean squared error (MSE) or other relevant measures of data corruption.

This aforementioned formulation ensures that the distillation process remains effective even in the presence of noisy data by integrating a mechanism to handle noise directly within the training objective.
Robust distillation techniques enhance model robustness by enabling student models to handle real-world data imperfections, resulting in more reliable performance in practical applications like RS where data quality can vary \cite{tang2024learning}. These techniques also improve generalization across diverse data conditions by incorporating noise handling into the training process, ensuring that models perform well in both clean and noisy environments. Additionally, robust distillation is particularly beneficial for sparse or imbalanced datasets, allowing student models to learn effectively from limited or unevenly distributed data while minimizing the risk of overfitting to noisy or rare examples \cite{liu2024small}.
\textcolor{black}{In RS, robust distillation against noisy data can be applied to cloud detection in satellite imagery, where cloud cover often introduces noise that obscures ground features. By employing robust distillation techniques, a student model can be trained to accurately detect clouds even in images with varying levels of noise caused by atmospheric conditions, ensuring more reliable and consistent results for subsequent analyses, such as land use classification or vegetation monitoring.}

\subsubsection{Semi-Supervised and Unsupervised Distillation}  
Semi-Supervised and Unsupervised Distillation represents an emerging area of research in KD, particularly relevant for fields like RS where labeled data is often scarce or expensive to obtain. The traditional KD process relies heavily on labeled datasets to transfer knowledge from a teacher model to a student model \cite{zhang2024semantic}. However, in many practical scenarios, especially in RS, obtaining a large volume of labeled data is challenging. To address this, semi-supervised and unsupervised distillation techniques aim to leverage the abundant unlabeled data available, reducing the dependency on labeled datasets and making the distillation process more robust and scalable \cite{lee2024semi}.

Semi-supervised and unsupervised distillation techniques leverage both labeled and unlabeled data during training to enhance model generalization and scalability. Labeled data provides accurate supervision, while unlabeled data exposes the model to a broader range of scenarios, improving its ability to generalize to new situations \cite{heidler2024pixeldino}. In semi-supervised distillation, a parameter \( \alpha \) balances the influence of labeled and unlabeled data, allowing the model to rely more on labeled data initially and gradually incorporate more from the unlabeled data. In purely unsupervised distillation, the student model learns directly from the teacher model's predictions, using them as pseudo-labels \cite{yang2024knowledge}. These approaches are particularly valuable in RS, where large quantities of data are available but often lack comprehensive labeling, enabling the development of robust models that efficiently handle vast amounts of unlabeled data \cite{pan2024semi}. The loss function for semi-supervised KD can be expressed as:

\begin{equation}
\mathcal{L}_{\text{KD-semi}} = \alpha \cdot \mathcal{L}_{\text{KD}}(T, S; \mathcal{D}_{\text{labeled}}) + (1-\alpha) \cdot \mathcal{L}_{\text{KD}}(T, S; \mathcal{D}_{\text{unlabeled}})
\end{equation}
The overall loss function includes \( \mathcal{L}_{\text{KD}}(T, S; \mathcal{D}_{\text{labeled}}) \), which represents the KD loss computed using the labeled dataset \( \mathcal{D}_{\text{labeled}} \), and \( \mathcal{L}_{\text{KD}}(T, S; \mathcal{D}_{\text{unlabeled}}) \), which is the distillation loss calculated from the unlabeled dataset \( \mathcal{D}_{\text{unlabeled}} \) where the teacher model's predictions serve as pseudo-labels. The weighting factor \( \alpha \) adjusts the influence of the labeled and unlabeled data on the overall loss function.

The above formulation allows the student model to learn from both labeled and unlabeled data, making the training process more flexible and less dependent on extensive labeled datasets. Semi-supervised and unsupervised distillation techniques offer significant practical advantages by reducing the reliance on large, high-quality labeled datasets, making model training more feasible in resource-constrained environments. These methods enhance generalization by exposing models to a wider variety of data patterns, allowing them to perform well on new, unseen data--an essential capability in fields like RS, where data diversity is high. Additionally, these techniques provide a scalable approach to model training, enabling organizations to leverage vast amounts of unlabeled data to build robust models without the need for extensive manual labeling efforts.
\textcolor{black}{Semi-supervised and unsupervised distillation can be applied to satellite imagery for land cover classification, where acquiring labeled data for every land type is impractical. By using semi-supervised distillation, a model can initially learn from a small set of labeled images and then leverage the large pool of unlabeled satellite images, where the teacher model provides pseudo-labels to refine the student model's performance.}

\subsection{Scalability Solutions}
\subsubsection{Distributed Distillation}  
Distributed distillation is an advanced approach designed to scale the KD process across larger datasets by leveraging distributed learning frameworks. This method spreads the computational load of both the teacher and student models across multiple nodes or devices, making it more feasible to handle large-scale data. The key idea is to perform distillation in a parallel or distributed manner, where each node or device processes a subset of the data, thus reducing the individual computational burden and allowing for more efficient training \cite{bistritz2020distributed}.
In distributed distillation, the overall training task is divided among multiple nodes in a distributed learning framework. Each node independently performs a portion of the distillation process, working with its local subset of data. This division of labor helps in managing the computational load effectively, enabling the distillation process to scale with the size of the dataset. The use of multiple nodes allows for parallel processing, which significantly speeds up the training process \cite{malinin2019ensemble}.
Each node in the distributed system hosts a teacher model and a student model, denoted as \( T_k \) and \( S_k \), respectively, where \( k \) represents the node index. These models operate on the local data available to that node. The student model on each node learns from its corresponding teacher model, capturing knowledge specific to that subset of the data. This localized learning allows the student models to collectively capture diverse knowledge from the entire dataset when combined \cite{anil2018large}.
After the distributed distillation process is complete, the student models from all nodes are aggregated to form a comprehensive model that incorporates the knowledge distilled from all parts of the dataset. The aggregation can be done by averaging the model weights, or by combining the outputs of the student models. The objective function \( \mathcal{L}_{\text{KD-distributed}} \) is averaged over all nodes, ensuring that the final student model reflects a balanced learning from the entire distributed dataset \cite{ryabinin2021scaling}.
The primary advantage of distributed distillation is its scalability. By distributing the workload, it becomes feasible to train on extremely large datasets that would be otherwise impractical to process on a single machine. This approach is particularly useful in scenarios like RS, where data is often collected in large quantities from multiple sources and needs to be processed efficiently. Distributed distillation allows for more rapid training and deployment of models, making it an effective solution for handling big data in ML \cite{shen2021real,fathullah2022self}.
The loss function for distributed distillation is given by:

\begin{equation}
\mathcal{L}_{\text{KD-distributed}} = \frac{1}{K} \sum_{k=1}^{K} \mathcal{L}_{\text{KD}}(T_k, S_k)
\end{equation}
where \( \mathcal{L}_{\text{KD}}(T_k, S_k) \) represents the KD loss computed on the \( k \)-th node, with \( T_k \) and \( S_k \) being the teacher and student models on that node. The term \( K \) denotes the total number of nodes involved in the distributed learning process. By averaging the distillation losses across all nodes, the overall objective function ensures that the student model benefits from the collective knowledge distributed across the dataset \cite{ryabinin2021scaling}. 
\textcolor{black}{Distributed distillation can be utilized in the analysis of global satellite imagery, where the vast amount of data is processed across multiple computational nodes. For instance, in mapping land use changes over large geographic regions, distributed distillation allows each node to handle different segments of the satellite images, collectively training a student model that integrates insights from all segments, leading to a comprehensive and scalable approach for detecting and classifying land cover changes \cite{taya2022decentralized}.}

\subsubsection{Incremental Distillation}  
Incremental distillation is a technique designed to efficiently handle the continuous influx of new data without the need to retrain the student model from scratch each time new information becomes available. This method is particularly beneficial in scenarios where datasets grow over time, such as in real-time applications or when new data is periodically added, as it allows for the model to be updated incrementally \cite{ruan2022class}.
The key idea behind incremental distillation is to enable the student model to learn from new data as it arrives, while also retaining the knowledge gained from previous training sessions. At each time step \( t \), the student model \( S_t \) is trained to mimic the behavior of the teacher model \( T_t \), which is typically derived from the latest data \cite{shen2023class}.
A critical aspect of incremental distillation is the preservation of previously learned knowledge. As the student model is updated with new data, it is important to ensure that it does not forget what it learned in earlier stages. This is achieved by incorporating a historical distillation term \( \mathcal{L}_{\text{history}} \) in the loss function, which measures the difference between the current student model \( S_t \) and its previous version \( S_{t-1} \). This regularization term helps in maintaining continuity and stability in the learning process, preventing catastrophic forgetting \cite{guan2023class}.
The overall loss function for incremental distillation, \( \mathcal{L}_{\text{KD-incremental}} \), includes two components:
        \begin{itemize}
            \item \textbf{KD Loss:} \( \mathcal{L}_{\text{KD}}(T_t, S_t) \) ensures that the student model learns from the current teacher model at time \( t \).
            \item \textbf{Historical Distillation Loss:} \( \mathcal{L}_{\text{history}}(S_t, S_{t-1}) \) penalizes deviations from the knowledge previously acquired by the student model, helping it retain important information from past iterations.
        \end{itemize}
The parameter \( \beta \) controls the balance between learning new information and preserving old knowledge. A higher \( \beta \) places more emphasis on retaining historical knowledge, while a lower \( \beta \) allows the model to adapt more quickly to new data \cite{lu2021lil}.

Incremental distillation provides significant practical benefits by allowing the student model to be updated incrementally, thereby avoiding the high computational costs associated with full model retraining. This approach is especially advantageous for large-scale applications where datasets are continuously evolving, such as in RS or streaming analytics \cite{ye2024multiscale}. It ensures that the model remains adaptive to new data trends, maintaining its relevance and accuracy over time. Additionally, the use of a historical distillation term enhances the stability of the learning process, minimizing the risk of performance degradation as new data is incorporated \cite{xie2024missnet}. \textcolor{black}{
Incremental distillation can be used for real-time monitoring of deforestation in satellite imagery. As new satellite images become available, the student model is updated incrementally to learn from the latest data while preserving its ability to recognize previously identified deforestation patterns, thus allowing continuous and efficient tracking of forest loss over time without needing to retrain the model from scratch \cite{arnaudo2022contrastive}.}

\subsection{Real-Time Processing Enhancements}

\subsubsection{Real-Time Distillation Algorithms}  
Real-time distillation algorithms are critical in scenarios where timely decision-making is essential, such as in RS applications that involve disaster monitoring or autonomous systems. These algorithms are designed to optimize both the accuracy and speed of the inference process, ensuring that the student model can deliver predictions within the stringent time constraints required for real-time operations \cite{shen2021real}.
In real-time applications, there's a trade-off between model accuracy and inference speed. A highly accurate model may be too slow for real-time processing, while a faster model might sacrifice accuracy. Real-time distillation algorithms aim to strike a balance by training the student model to achieve an acceptable level of accuracy while ensuring that it can make predictions quickly enough to meet real-time requirements \cite{wu2021real}.
The inference time, denoted as \( t_{\text{inference}} \), is the time it takes for the model to process an input and produce an output. For real-time applications, this must be less than or equal to a predetermined threshold \( t_{\text{real-time}} \). The distillation process incorporates this requirement into the training by adding a penalty term in the loss function if the student model's inference time exceeds the threshold. This ensures that the final model is optimized not only for accuracy but also for speed \cite{zhuo2024fast}.
The penalty term \( \gamma \cdot \mathbb{I}(t_{\text{inference}} \leq t_{\text{real-time}}) \) in the loss function introduces a strong disincentive for any delay in inference time. Here, \( \mathbb{I}(\cdot) \) is an indicator function that activates the penalty whenever the inference time \( t_{\text{inference}} \) is greater than the real-time threshold \( t_{\text{real-time}} \). This encourages the model to adhere strictly to time constraints, making it suitable for deployment in environments where every millisecond counts, such as in disaster response or real-time traffic management \cite{li2020training}.
In RS, where data often needs to be processed and acted upon quickly (e.g., detecting natural disasters, monitoring climate changes, or guiding autonomous vehicles), real-time distillation ensures that models are both accurate and fast enough to be practical. This is particularly important in disaster monitoring, where delays in data processing could result in missed opportunities to mitigate damage or save lives \cite{grunenfelder2023fast}. 
The loss function for real-time distillation, \( \mathcal{L}_{\text{KD-real-time}} \), combines the traditional KD loss with a penalty for exceeding the inference time threshold:

\begin{equation}
\mathcal{L}_{\text{KD-real-time}} = \mathcal{L}_{\text{KD}}(T, S) + \gamma \cdot \mathbb{I}(t_{\text{inference}} \leq t_{\text{real-time}})
\end{equation}
where \( \mathcal{L}_{\text{KD}}(T, S) \) represents the standard Knowledge Distillation (KD) loss, which measures how effectively the student model replicates the teacher model's behavior. The hyperparameter \( \gamma \) controls the strength of the penalty applied when the inference time exceeds the real-time threshold. Additionally, the indicator function \( \mathbb{I}(\cdot) \) activates this penalty if the inference time \( t_{\text{inference}} \) surpasses the allowed real-time limit \( t_{\text{real-time}} \).
This formulation ensures that the student model not only learns effectively from the teacher model but also adheres to the necessary timing constraints for real-time deployment \cite{thakker2022fast}.
The development of real-time distillation algorithms provides several practical benefits, including the ability to make timely decisions, which is critical in applications such as disaster monitoring where delays can have severe consequences. These algorithms efficiently balance the trade-offs between speed and accuracy, ensuring models are both effective and practical for real-time use. Additionally, their versatility makes them suitable for a wide range of fields that require real-time processing, from RS to autonomous driving and other time-sensitive applications \cite{islam2024spatial}. \textcolor{black}{Real-time distillation algorithms can be utilized in monitoring wildfires through satellite imagery, where immediate detection and response are crucial. The student model is trained to rapidly process satellite images and detect fire outbreaks in real-time while adhering to stringent time constraints, ensuring timely alerts for emergency services and minimizing potential damage \cite{dave2003online}.}

\subsubsection{Edge-AI Distillation}  
Edge-AI distillation focuses on optimizing ML models specifically for deployment on edge devices, such as sensors, drones, or other low-power, resource-constrained hardware commonly used in RS applications \cite{angarano2023generative}. These devices often have limited computational power, memory, and battery life, which necessitates the development of highly efficient models that can perform complex tasks with minimal resources. The goal of edge-AI distillation is to ensure that the distilled student model is not only accurate but also energy-efficient and capable of running with low latency on edge devices \cite{sepahvand2023adaptive}.

The key to achieving this lies in balancing the KD process with the energy consumption constraints of the target hardware. The distillation loss function denoted as \( \mathcal{L}_{\text{KD}}(T, S) \), is typically used to align the outputs of the student model \( S \) with those of the teacher model \( T \). However, in the context of edge-AI, an additional term, \( \text{Energy}(\theta_S) \), is introduced into the objective function to account for the energy consumption of the student model on the edge device \cite{dey2024towards}.

This approach involves minimizing not just the distillation loss but also the energy consumption associated with running the student model. The regularization parameter \( \delta \) is used to balance the importance of energy efficiency against the need to maintain high model performance. By carefully tuning this parameter, it is possible to develop models that are both effective in their predictive capabilities and efficient in terms of energy usage \cite{wang2020industrial}.


\textcolor{black}{Edge-AI distillation can be employed in drones for real-time wildlife monitoring, where models need to process and analyze video feeds on the device itself. By optimizing the model for low power consumption and fast inference, drones can efficiently identify and track animal movements without draining their batteries or relying on constant data transmission to central servers.}

\subsection{Cross-Modal and Multi-Modal Distillation}

\subsubsection{Cross-Modal Knowledge Transfer}  
Cross-modal distillation involves transferring knowledge from one modality (e.g., optical images) to another (e.g., SAR or multispectral images). This approach can improve model generalization across different data sources \cite{huo2024c2kd,zhu2024cross}.

\begin{equation}
\mathcal{L}_{\text{KD-cross}} = \sum_{m=1}^{M} w_m \cdot \mathcal{L}_{\text{KD}}(T_m(x_m), S(x_m))
\end{equation}

where \( M \) is the number of modalities, and \( w_m \) is the weight associated with each modality \( m \).

Cross-modal knowledge transfer is a specialized technique within the broader context of KD, focusing on transferring knowledge from one data modality to another. In RS, data is often collected from various sources, each providing unique and complementary information about the environment. For instance, optical images capture visible light, SAR (Synthetic Aperture Radar) images provide microwave data, and multispectral images cover a range of wavelengths beyond the visible spectrum. Each modality offers distinct advantages and limitations, making it valuable to develop models that can generalize across these diverse data types \cite{li2024rsmodm}.
The essence of cross-modal distillation lies in the ability to leverage a teacher model trained on one modality (e.g., optical images) to improve the performance of a student model operating on a different modality (e.g., SAR or multispectral images). This process enhances the student model's ability to generalize and perform well across different data sources, which is crucial in RS tasks that require robust performance across varying environmental conditions and sensor types \cite{zhu2024cross}.

The formulation of cross-modal distillation is encapsulated in the loss function \( \mathcal{L}_{\text{KD-cross}} \). This loss function aggregates the KD process across multiple modalities, denoted by \( M \). For each modality \( m \), a specific weight \( w_m \) is assigned, reflecting the importance or relevance of that modality in the distillation process. The objective is to minimize the weighted sum of the KD losses \( \mathcal{L}_{\text{KD}} \) for each modality, where \( T_m(x_m) \) represents the output of the teacher model for modality \( m \), and \( S(x_m) \) is the corresponding output of the student model \cite{ienco2024discom}.
By carefully selecting and tuning the weights \( w_m \), the cross-modal distillation process can be tailored to emphasize certain modalities over others, depending on the specific requirements of the application. For example, in scenarios where optical images are more informative, the weight \( w_m \) for optical data can be increased, ensuring that the student model learns more effectively from that modality \cite{chen2024scale}. Cross-modal knowledge transfer is particularly beneficial in RS, where data from different modalities may be abundant but not always consistently available. By enabling models to transfer knowledge across modalities, this approach ensures that the student model remains robust and effective even when some data sources are missing or degraded. This capability is critical in applications such as environmental monitoring, disaster response, and resource management, where reliable and accurate information from diverse data sources is essential for decision-making \cite{zavras2024mind}.
\textcolor{black}{As mentioned above, cross-modal knowledge transfer can be applied in integrating SAR and optical satellite imagery for flood monitoring. By leveraging a teacher model trained on high-resolution optical images, a student model can be optimized to accurately interpret SAR images, enhancing the detection of flood extents and water levels in areas where optical data might be obstructed or unavailable.}

\subsubsection{Multi-Task Distillation}  
Multi-task distillation is an advanced technique in KD where a single student model is trained to perform multiple tasks simultaneously, such as classification, segmentation, and object detection. This approach aims to create a more versatile and efficient model that can handle a variety of tasks without compromising performance in any of them. By distilling knowledge from teacher models specialized in different tasks, the student model learns to balance these tasks effectively, making it highly valuable in applications where multi-functionality is essential, such as in RS or autonomous systems \cite{zhang2023fusion}.
Multi-task distillation involves training a student model to handle multiple tasks concurrently. Each task has its own teacher model, and the student model learns from these teachers to perform all tasks simultaneously. For example, in RS, one teacher model might be specialized in land cover classification, while another is focused on detecting specific objects like vehicles or buildings. The student model is designed to learn from both these tasks, allowing it to perform land cover classification and object detection within the same framework \cite{dong2023multi}.
The loss function for multi-task distillation, denoted as \( \mathcal{L}_{\text{KD-multi-task}} \), is a weighted sum of the distillation losses from each task. The weight \( \lambda_t \) assigned to each task allows for prioritization based on the importance or difficulty of the task. For instance, if segmentation is more critical for a specific application than classification, a higher weight can be assigned to the segmentation task to ensure the student model focuses more on that aspect \cite{hong2023multi}.
\begin{equation}
    \mathcal{L}_{\text{KD-multi-task}} = \sum_{t=1}^{T} \lambda_t \cdot \mathcal{L}_{\text{KD}}(T_t, S_t)
\end{equation}
Where \( T \) represents the total number of tasks, and \( \mathcal{L}_{\text{KD}}(T_t, S_t) \) is the KD loss for task \( t \) between the teacher \( T_t \) and the student model \( S_t \). 
One of the key challenges in multi-task distillation is balancing the learning of different tasks. Since each task might require different levels of focus or complexity, the student model must be carefully trained to avoid underperforming in one task while excelling in another. The weight \( \lambda_t \) helps manage this balance by allowing certain tasks to have more influence on the model's learning process \cite{zhou2022self}.
By combining multiple tasks into a single model, multi-task distillation reduces the need for deploying separate models for each task, saving computational resources and simplifying the deployment process. This makes the distilled model more efficient and versatile, particularly in environments where resources are limited or where real-time processing of multiple tasks is required \cite{liu2024tomato}.

In practical scenarios, such as RS, a multi-task distilled model could simultaneously analyze satellite images for land classification, detect changes over time, and identify specific objects or features. This approach is not only more efficient but also enables the model to leverage shared information across tasks, leading to better overall performance and more coherent results across the different tasks \cite{zhu2024sirs}. While multi-task distillation offers numerous advantages, it also comes with challenges, such as the potential for task interference, where learning one task might negatively impact another. Careful design of the distillation process and appropriate weighting of tasks are essential to ensure that the student model performs well across all tasks \cite{zhang2024dual}. \textcolor{black}{A multi-task distilled model can be used for comprehensive urban monitoring from satellite images, where it simultaneously performs land cover classification, detects infrastructure changes (such as new buildings or roads), and identifies specific objects (like vehicles or trees). This enables efficient and integrated analysis of diverse data, enhancing the ability to track urban development and infrastructure changes in a single model \cite{yuan2024continual}.}

\subsection{Seamless Integration with Existing Workflows}

\subsubsection{Plug-and-Play Distillation Modules}  
Creating plug-and-play distillation modules that can be easily integrated into existing RS workflows offers several benefits \cite{jin2022plug}. These modules are designed to be seamlessly incorporated with minimal adjustments to the current infrastructure, reducing the barriers to adopting knowledge distillation (KD) in RS applications \cite{sun2024logit}. The formulation of this integration can be expressed as:
\[
\mathcal{L}_{\text{KD-plug}} = \mathcal{L}_{\text{KD}}(T, S) + \sum_{m=1}^{M} \mathcal{L}_{\text{mod}}(S, M_m)
\]
where \( \mathcal{L}_{\text{KD}}(T, S) \) represents the standard KD loss between the teacher model \( T \) and the student model \( S \), and \( \mathcal{L}_{\text{mod}}(S, M_m) \) denotes the additional loss terms introduced by integrating existing modules \( M_m \) with the student model \( S \). The modularity of this approach allows different components to be swapped or upgraded independently, facilitating the adoption of KD in diverse scenarios. This scalability ensures that KD remains applicable across a wide range of RS tasks, from small UAV datasets to extensive satellite imagery. By leveraging pre-built modular KD techniques, developers and researchers can save time on implementation, accelerating the development and deployment of RS models, which ultimately enhances their performance \cite{hsiao2024plug,lao2023unikd}.

\subsubsection{Toolkits and Frameworks} 
Developing comprehensive toolkits and frameworks for KD in RS can significantly enhance its performance and adoption. These toolkits provide standardized implementations of KD methods, ensuring consistency and reliability across different RS tasks \cite{yang2020textbrewer}. The complexity of integrating various modules within the KD process can be expressed as:

\[
\text{Toolkit Complexity} \propto \sum_{i=1}^{N} C_{\text{integration}}(M_i, \mathcal{L}_{\text{KD}})
\]
where \( C_{\text{integration}}(M_i, \mathcal{L}_{\text{KD}}) \) represents the complexity of integrating module \( M_i \) with the KD process. This standardization reduces the variability in performance that can arise from ad-hoc implementations, leading to more predictable results and making KD a more reliable option for RS applications. Additionally, these toolkits lower the technical barriers for practitioners by providing user-friendly interfaces and comprehensive documentation. Frameworks often include optimized routines for tasks such as hyperparameter tuning or data preprocessing, which can lead to better model performance and faster training times, especially in computationally intensive RS tasks. The community-driven development of these toolkits and frameworks leads to faster identification of bugs, new feature additions, and overall better support for the technology. This collective improvement makes KD more practical and effective for RS tasks, contributing to enhanced model performance and more efficient applications \cite{matsubara2021torchdistill}.

\subsection{Enhancing Model Interpretability}

\subsubsection{Explainable Distillation}
Explainable distillation is an advanced approach in KD that not only focuses on transferring the predictive performance of the teacher model to the student model but also ensures that the student model's decision-making process is interpretable. The goal is to create models that are not just accurate but also transparent, providing insights into how they arrive at their predictions \cite{batic2023improving}. This is particularly important in critical applications like RS, healthcare, and autonomous systems, where understanding the model's reasoning is crucial for trust and accountability \cite{liu2023explainable}. Explainable distillation mainly relies on the following key concepts.
Traditional KD emphasizes performance, often at the cost of interpretability. However, in many applications, it is essential to know why a model makes a certain decision. Explainable distillation aims to bridge this gap by integrating interpretability into the distillation process. The student model is trained not only to mimic the teacher's outputs but also to generate explanations for its decisions that are comprehensible to humans \cite{taskin2022model}.
The process of explainable distillation involves an additional term in the loss function, denoted as \( \mathcal{L}_{\text{explain}}(S) \), which penalizes the student model if its explanations are not clear or do not align with certain interpretability criteria. This ensures that the student model learns to provide meaningful insights alongside accurate predictions \cite{lee2024unlocking}.
\begin{equation}
    \mathcal{L}_{\text{KD-explain}} = \mathcal{L}_{\text{KD}}(T, S) + \xi \cdot \mathcal{L}_{\text{explain}}(S)
\end{equation}
where, \( \mathcal{L}_{\text{KD}}(T, S) \) is the standard KD loss, and \( \xi \) is a regularization parameter that controls the trade-off between accuracy and interpretability. The term \( \mathcal{L}_{\text{explain}}(S) \) guides the student model toward generating explanations that are either inherently interpretable or match the explanations provided by the teacher model if available \cite{liu2023explainable}.

In domains such as RS, where decisions based on model predictions can have significant real-world impacts, the ability to explain model outputs is critical. For example, when a model identifies a potential disaster area in satellite imagery, it is not enough to simply flag the area; stakeholders need to understand the reasoning behind the decision, such as the specific patterns or features that led to the prediction \cite{taskin2022model}.
Various explainable AI techniques can be incorporated into the distillation process, such as saliency maps, attention mechanisms, or feature attribution methods \cite{lee2021explaining}. These techniques help to visualize and understand which parts of the input data are most influential in the model's decision-making process. By integrating these techniques into the distillation process, the student model becomes not only a distilled version of the teacher but also a more interpretable and transparent model \cite{termritthikun2023explainable}.

All in all, explainable distillation holds significant potential in fields where trust in AI systems is paramount. By producing models that are both accurate and interpretable, it enhances the usability and acceptance of AI systems in sensitive areas. Moreover, it aids in regulatory compliance, as interpretable models can more easily meet legal and ethical standards regarding AI transparency \cite{li2024hybrid}. Besides, while explainable distillation is a promising approach, it comes with challenges, such as defining and quantifying interpretability in a way that aligns with both human understanding and model performance. Future research may focus on developing more sophisticated explainability metrics and integrating them seamlessly into the distillation process, ensuring that models are not only effective but also understandable and trustworthy \cite{mi2022kde}.

\subsubsection{Feature Importance Preservation}  
During the distillation process, there is a risk that the simplified student model might not fully capture or retain the importance of these critical features. If the student model fails to recognize the same features as important, it might lead to reduced performance and decreased interpretability. Moreover, the loss of feature importance can erode trust in the model, especially in high-stakes scenarios where the rationale behind predictions needs to be transparent \cite{xiao2023knowledge}.
To address this challenge, the distillation process can incorporate a feature importance preservation mechanism. This involves adding an additional term to the standard KD loss function that explicitly penalizes differences in feature importance between the teacher and student models \cite{yang2023attention}. The modified loss function can be expressed as:

\begin{equation}
\mathcal{L}_{\text{KD-feature}} = \mathcal{L}_{\text{KD}}(T, S) + \zeta \cdot \sum_{i=1}^{d} \left| f_i^T - f_i^S \right|
\end{equation}
where \( \mathcal{L}_{\text{KD}}(T, S) \) is the standard KD loss, which aligns the outputs of the student model \( S \) with those of the teacher model \( T \). \( f_i^T \) and \( f_i^S \) represent the importance of the \( i \)-th feature in the teacher and student models, respectively. \( \zeta \) is a regularization parameter that controls the weight given to the feature importance preservation term in the overall loss function. The summation term \( \sum_{i=1}^{d} \left| f_i^T - f_i^S \right| \) measures the discrepancy between the feature importance values of the teacher and student models across all features \( i \) in the input space.

By ensuring that the student model not only mimics the predictions of the teacher model but also recognizes the same features as important, the feature importance preservation approach enhances the interpretability of the student model. This is particularly useful in applications where the end-users, such as analysts or domain experts, need to understand the reasoning behind the model's decisions. In RS, for instance, preserving feature importance can ensure that the student model correctly identifies key environmental indicators, such as vegetation health or water quality, that are critical for accurate predictions \cite{chen2020learning}.
The preservation of feature importance can also play a significant role in building trust in AI systems. When users can see that the model is focusing on the right features, they are more likely to trust its predictions. This is particularly crucial in regulated industries or applications involving ethical considerations, where transparency is mandatory \cite{zhou2022effective}.
While feature importance preservation is beneficial, it introduces a trade-off between model simplicity and interpretability. The regularization parameter \( \zeta \) needs to be carefully tuned to balance the preservation of feature importance with the overall performance of the student model. If \( \zeta \) is too high, the student model may overemphasize feature alignment at the cost of predictive accuracy. Conversely, if \( \zeta \) is too low, the student model might neglect important features, compromising its interpretability and trustworthiness \cite{lv2021fusion,xu2022vision}.

\subsection{Hybrid Approaches}

\subsubsection{Combining Distillation with Other Techniques}  
Hybrid approaches in ML involve the integration of multiple learning paradigms to create more powerful and versatile models. When applied to KD, these hybrid approaches can significantly enhance the performance and efficiency of models, especially in complex and data-rich fields like RS \cite{aghli2021combining}. For instance, a hybrid approach relies on combining KD with transfer learning and federated learning. Typically, transfer learning involves leveraging knowledge gained from one task or domain to improve the performance on a related but different task. In the context of hybrid approaches, transfer learning can be integrated into the distillation process to help the student model acquire additional knowledge from pre-trained models on related tasks \cite{malihi2024matching}. The transfer learning loss component, \( \mathcal{L}_{\text{Transfer}}(S) \), represents the cost associated with adapting the student model to the new task or domain. Moving on, reinforcement learning is a paradigm where models learn to make decisions by interacting with an environment and receiving feedback in the form of rewards or penalties. In hybrid distillation approaches, RL can be incorporated to optimize the student model's performance through trial and error, especially in scenarios where decision-making under uncertainty is critical. The RL loss component, \( \mathcal{L}_{\text{Reinforce}}(S) \), measures how well the student model performs in achieving its objectives within the environment \cite{kuldashboy2024efficient}.

Besides, active learning is another technique that can be integrated with KD. It involves selectively querying the most informative data points for training, thereby improving the model's performance with fewer labeled examples. While not explicitly included in the equation, active learning can complement the distillation process by ensuring that the most critical data points are used for training the student model \cite{kwak2022trustal,boreshban2023improving}. 
The hybrid loss function in this approach combines the losses from KD, transfer learning, and reinforcement learning:

    \begin{equation}
    \mathcal{L}_{\text{KD-hybrid}} = \alpha \cdot \mathcal{L}_{\text{KD}}(T, S) + \beta \cdot \mathcal{L}_{\text{Transfer}}(S) + \gamma \cdot \mathcal{L}_{\text{Reinforce}}(S)
    \end{equation}
In this context, \( \alpha \), \( \beta \), and \( \gamma \) are weighting factors that determine the contribution of each component to the overall loss function. The term \( \mathcal{L}_{\text{KD}}(T, S) \) ensures that the student model closely mimics the behavior of the teacher model, while \( \mathcal{L}_{\text{Transfer}}(S) \) aids the student model in adapting to new tasks or domains by utilizing previously acquired knowledge. Additionally, \( \mathcal{L}_{\text{Reinforce}}(S) \) enhances the student model's decision-making abilities through interaction with an environment, thereby optimizing its performance \cite{zhang2020hybrid}.

By integrating multiple learning paradigms, hybrid approaches can create models that are not only smaller and faster (through distillation) but also more knowledgeable and capable (through transfer learning) and more adaptive (through reinforcement learning). Hybrid models are better equipped to handle a variety of tasks and environments, as they combine the strengths of different learning methods. This is particularly useful in RS, where data can vary widely in type, quality, and context. The ability to combine different learning strategies makes hybrid approaches highly versatile. For example, a hybrid model might use transfer learning to understand basic image recognition tasks while using reinforcement learning to make real-time decisions based on that understanding \cite{xie2023hybrid}.

Hybrid distillation approaches are particularly relevant in complex fields such as RS, where models need to process and analyze vast amounts of data from multiple sources. For instance, in disaster monitoring, a hybrid model could use transfer learning to recognize different types of terrain, reinforcement learning to predict the spread of a fire, and KD to ensure the model is efficient enough to run on edge devices deployed in the field \cite{zhang2024soft}.
However, combining multiple learning paradigms can increase the complexity of the model and its training process. Careful tuning of the weighting factors (\( \alpha \), \( \beta \), \( \gamma \)) is necessary to achieve the desired balance between the different learning objectives. Moreover, while the goal of distillation is to create efficient models, the integration of transfer learning and reinforcement learning can demand additional computational resources during training, which may limit the scalability of the approach \cite{zhang2024hybrid}.

\subsubsection{Adaptive Distillation Frameworks}  
Developing adaptive distillation frameworks that can adjust the distillation process based on the complexity of the task or the availability of data could lead to more flexible and robust models. Traditional knowledge distillation methods apply a fixed strategy for transferring knowledge from the teacher model to the student model, which might not be optimal for all scenarios, particularly in RS where data heterogeneity and varying task requirements are common \cite{liang2024module}.
An adaptive distillation framework dynamically adjusts the distillation process by incorporating additional mechanisms that account for task complexity, data quality, or computational constraints. This dynamic adjustment allows the distillation process to be more responsive to the specific needs of the application, improving both the efficiency and effectiveness of the student model \cite{li2024importance}. For instance, in scenarios where the task is relatively simple or where data is abundant and high-quality, the framework could prioritize rapid model convergence by assigning higher weights to the knowledge distillation loss. Conversely, in more complex tasks or when dealing with sparse or noisy data, the framework could allocate more resources to the adaptation process, fine-tuning the student model to handle these challenges more effectively \cite{zhang2024cross}. The adaptive distillation process can be represented as:

\begin{equation}
\mathcal{L}_{\text{KD-adaptive}} = \sum_{t=1}^{T} \alpha_t \cdot \mathcal{L}_{\text{KD}}(T_t, S_t) + \mu_t \cdot \mathcal{L}_{\text{adaptive}}(S_t)
\end{equation}
where \( \alpha_t \) and \( \mu_t \) are time-varying weights for the KD and adaptation loss components, respectively. These weights are not static but rather evolve over time \( t \) based on factors such as the current performance of the student model, the difficulty of the current task, and the quality of the data available at each step \cite{mi2024adaptive}.

The term \( \mathcal{L}_{\text{KD}}(T_t, S_t) \) represents the traditional knowledge distillation loss, which measures the discrepancy between the outputs of the teacher and student models. The term \( \mathcal{L}_{\text{adaptive}}(S_t) \) is an additional adaptation loss that could include penalties for model complexity, regularization terms to prevent overfitting, or other factors that enhance the student model's ability to generalize across different tasks or datasets \cite{yu2024adaptive}.
By allowing the distillation process to adapt to varying conditions, these frameworks can produce student models that are not only smaller and faster but also more versatile and capable of maintaining high performance across a range of tasks and environments. This adaptability is particularly valuable in RS, where the conditions under which models operate can vary significantly, and the ability to generalize effectively is crucial \cite{huang2022extracting}.

\section{Conclusion}
This review provides a comprehensive analysis of knowledge distillation (KD) and its applications in remote sensing (RS), making significant contributions to the understanding and advancement of this technique. The review begins by offering a detailed overview of the fundamentals of KD, including its definition, basic concepts, historical evolution, and underlying mechanisms. The advantages of KD, such as model compression, improved efficiency, and enhanced performance in smaller models, are highlighted, particularly in the context of RS tasks like image classification, object detection, land cover classification, and semantic segmentation. 
A key contribution of this review is the taxonomy of KD models, which categorizes the variations in the models and input data, the type of knowledge transferred, the distillation target, and the structural relationships between network layers. This taxonomy serves as a valuable resource for researchers and practitioners, providing a clear framework for understanding the diverse applications and implementations of KD in RS.

However, the review also addresses the challenges and limitations of applying KD in remote sensing. These challenges include the complexity of model deployment, the heterogeneity of data sources, overfitting and generalization issues, scalability, real-time applicability, dependency on high-quality data, and the difficulty in balancing efficiency and accuracy. These challenges underscore the need for continued innovation and refinement of KD techniques to meet the unique demands of remote sensing applications. Movin forward, the review identifies several important future directions for the field. These include the development of advanced model compression techniques, dynamic and layer-wise distillation, efficient training algorithms, and hardware-aware distillation. The importance of improving data quality and robustness through robust distillation against noisy data and semi-supervised or unsupervised approaches is also emphasized. Additionally, the review suggests scalability solutions like distributed and incremental distillation, real-time processing enhancements, cross-modal and multi-modal distillation, and the seamless integration of KD into existing workflows through plug-and-play modules and standardized toolkits.

Finally, the review calls for efforts to enhance model interpretability through explainable distillation and feature importance preservation, as well as exploring hybrid approaches that combine KD with other techniques. These future directions highlight the ongoing potential for KD to revolutionize remote sensing applications, offering pathways to overcome current limitations and achieve greater accuracy, efficiency, and applicability in this critical field.

\section*{Acknowledgement}
No funding is available for this work.


\section*{Conflict of Interest}
The authors declare no conflicts of interest.


\end{document}